\documentclass[Afour,sageh,times]{sagej_arxiv}
\usepackage{algorithm}
\usepackage[noend]{algpseudocode}
\usepackage{euscript}
\usepackage{color}
\usepackage{booktabs} 

\usepackage{amsfonts}
\usepackage{mathtools}
\usepackage{placeins}
\usepackage{microtype}
\usepackage{soul}
\usepackage{tikz}
\usepackage{wrapfig}

\usepackage[nolist]{acronym}
\usepackage[scr=boondox]{mathalfa}

\usepackage{pdfpages}
\usepackage{xr-hyper}

\usepackage{rotating}

\usepackage{graphicx}
\usepackage{subcaption}
\usepackage[toc=false,symbols,nogroupskip,sort=none]{glossaries-extra}

\usepackage{pdflscape}
\usepackage{afterpage} % used to prevent page break at lscape env

\usepackage{enumitem} %For tighter enumerate and itemize environments. Set globally with \setlist{nosep}. To set the enumerate type, you'll then need to pass the argument as label={(\roman*)}
\setlist{nosep} %Remove all spacing around the environments

\usepackage{array}
\newcolumntype{L}[1]{>{\raggedright\let\newline\\\arraybackslash\hspace{0pt}}m{#1}}
\newcolumntype{C}[1]{>{\centering\let\newline\\\arraybackslash\hspace{0pt}}m{#1}}
\newcolumntype{R}[1]{>{\raggedleft\let\newline\\\arraybackslash\hspace{0pt}}m{#1}}
\usepackage{layouts}

\newcommand{\norm}[1]{\left\lVert#1\right\rVert}
\newcommand{\mean}[1]{\bar{#1}}

\newcommand{\kjgraph}{\mathcal{N}}
\newcommand{\kjweight}{\omega}

\newcommand{\kjlabel}{\ell}
\newcommand{\kjdatasegment}[1]{\texttt{#1}}

\newcommand{\kjse}[1]{$SE\left(#1\right)$}
\newcommand{\kjsealg}[1]{$\mathfrak{se}\left(#1\right)$}
\newcommand{\kjtransform}[2]{\mathbf{T}_{#1}^{#2}}
\newcommand{\kjtransformrigid}[2]{\mathbf{F}_{#1}^{#2}}

\newcommand{\kjstate}{\mathbf{x}}
\newcommand{\kjstatepert}{\delta\mathbf{x}}

\newcommand{\kjoutlier}{\mathscr{O}}
\newcommand{\kjframe}[1]{\protect\underrightarrow{\boldsymbol{\mathcal{F}}}_{#1}}
\newcommand{\kjidentity}{\mathbf{1}}
\newcommand{\kjzero}{\mathbf{0}}
\newcommand{\kjpoint}[1]{\protect\underrightarrow{p}^{#1}}
\newcommand{\kjvecpoint}[2]{\mathbf{p}^{#1}_{#2}}

\newcommand{\kjvecpointmean}[2]{\mean{\mathbf{p}}^{#1}_{#2}}
\newcommand{\kjhopoint}[2]{\boldsymbol{p}^{#1}_{#2}}

\newcommand{\kjrotation}[1]{\mathbf{C}_{#1}}

\newcommand{\kjvelocity}[1]{\boldsymbol{\varpi}_{#1}}
\newcommand{\kjacceleration}[1]{\dot{\boldsymbol{\varpi}}_{#1}}
\newcommand{\kjaccelerationdot}[1]{\ddot{\boldsymbol{\varpi}}_{#1}}
\newcommand{\kjadjoint}[1]{\boldsymbol{\mathcal{T}}_{#1}}
\newcommand{\kjsoel}[1]{\boldsymbol{\phi}_{#1}}
\newcommand{\kjseel}[2]{\boldsymbol{\xi}_{#1#2}}
\newcommand{\kjsetran}[2]{\boldsymbol{\rho}_{#1#2}}
\newcommand{\kjtranset}[1]{\mathscr{T}_{#1}}

\newcommand{\kjpointset}[2]{\mathscr{p}^{#1}_{#2}}
\newcommand{\kjlabelset}[1]{\mathscr{L}_{#1}}
\newcommand{\kjtrackletset}[1]{\mathscr{P}_{#1}}
\newcommand{\kjvarconvergence}{N_{\mathrm{conv}}}

\DeclareMathOperator*{\argmax}{arg\,max}

\makeatletter
\newcommand*{\accite}[2]{%
	\expandafter\ifx\csname AC@#1\endcsname\AC@used
	\acs{#1} \citep{#2}%Easy-peezy.
	\else
	\acl{#1}\acused{#1} \citep[\acs{#1};][]{#2}% Make acronym and reference share ()s
	\fi
}
\newcommand*{\acpcite}[2]{%
	\expandafter\ifx\csname AC@#1\endcsname\AC@used
	\acsp{#1} \citep{#2}%Easy-peezy.
	\else
	\aclp{#1}\acused{#1} \citep[\acsp{#1};][]{#2}% Make acronym and reference share ()s
	\fi
}
\makeatother

\usepackage{moreverb,url}

\usepackage[colorlinks=false,hidelinks,bookmarksopen,bookmarksnumbered,citecolor=red,urlcolor=red,pdfpagelabels]{hyperref}%\def\volumeyear{2021}

\usepackage[capitalise,nameinlink]{cleveref}

\begin{acronym}
	\acro{MEP}{multimotion estimation problem}
	\acro{MOT}{multiobject tracking}
	\acro{VO}{visual odometry}
	\acro{VIO}{visual-inertial odometry}
	\acro{MVO}{Multimotion Visual Odometry}
	\acro{OMD}{Oxford Multimotion Dataset}
	\acro{knn}[$k$nn]{$k$ nearest neighbors}
	\acro{ICP}{iterative closest point}
	\acro{RANSAC}{RANdom SAmple Consensus}
	\acro{MLESAC}{Maximum Likelihood Estimation SAmple Consensus}
	\acro{CORAL}{convex relaxation algorithm}
	\acro{IMU}{inertial measurement unit}
	\acro{SLAM}{simultaneous localization and mapping}
	\acro{JIRCS}{Joint Industry and Robotics CDTs Symposium}
	\acro{ROS}{Robotic Operating System}
	\acro{IROS}{IEEE/RSJ International Conference on Intelligent Robots and Systems}
	\acro{ICRA}{IEEE International Conference on Robotics and Automation}
	\acro{RA-L}{IEEE Robotics and Automation Letters}
	\acro{SLAMMOT}{Simultaneous Localization, Mapping, and Moving Object Tracking}
	\acro{WNOA}{white-noise-on-acceleration}
	\acro{WNOJ}{white-noise-on-jerk}
	\acro{RMS}{root-mean-square}
	\acro{RMSRE}{\ac{RMS} relative error}
	\acro{VDO-SLAM}{Visual Direct Object-aware SLAM}
\end{acronym}

\begin{document}
	\runninghead{Judd and Gammell}
	
	\title{Multimotion Visual Odometry (MVO)}
	
	\author{Kevin M.\ Judd\affilnum{1} and Jonathan D.\ Gammell\affilnum{1}}
	
	\affiliation{\affilnum{1}Estimation, Search, and Planning Group, Oxford Robotics Institute, University of Oxford, UK}
	
	\corrauth{Kevin Judd, Estimation, Search, and Planning Group
		Oxford Robotics Institute,
		University of Oxford,
		Oxford
		OX2~6NN, UK.\\
		\url{https://robotic-esp.com/kjudd}}
	
	\email{kjudd@oxfordrobotics.institute}
	
	\begin{abstract}
		Visual motion estimation is a well-studied challenge in autonomous navigation.
		Recent work has focused on addressing multimotion estimation in highly dynamic environments.
		These environments not only comprise multiple, complex motions but also tend to exhibit significant occlusion.
		
		Estimating third-party motions simultaneously with the sensor egomotion is difficult because an object's observed motion consists of both its true motion and the sensor motion.
		Most previous works in multimotion estimation simplify this problem by relying on appearance-based object detection or application-specific motion constraints.
		These approaches are effective in specific applications and environments but do not generalize well to the full \acf{MEP}.

		This paper presents \acf{MVO}, a multimotion estimation pipeline that estimates the full \kjse{3} trajectory of \emph{every} motion in the scene, including the sensor egomotion, without relying on appearance-based information.
		\acs{MVO} extends the traditional \acf{VO} pipeline with multimotion segmentation and tracking techniques.
		It uses physically founded motion~priors to extrapolate motions through temporary occlusions and identify the reappearance of motions through \emph{motion closure}. 
		Evaluations on real-world data from the \acf{OMD} and the KITTI Vision Benchmark Suite demonstrate that \acs{MVO} achieves good estimation accuracy compared to similar approaches and is applicable to a variety of multimotion estimation challenges.
	\end{abstract}
	
	
	\maketitle

	\section{Introduction}
	\label{sec:introduction}
	\acresetall
	\vspace{-0.5mm}
	Autonomous robotic platforms are being deployed in an increasingly diverse range of applications.
	Many platforms and algorithms are highly specialized to operate in specific environments, but the ability to safely navigate diverse and complex dynamic environments is becoming more important.

	\begin{figure}[t]
		\centering
		\begin{subfigure}[t]{\columnwidth}
			\includegraphics[clip,width=.98\columnwidth]{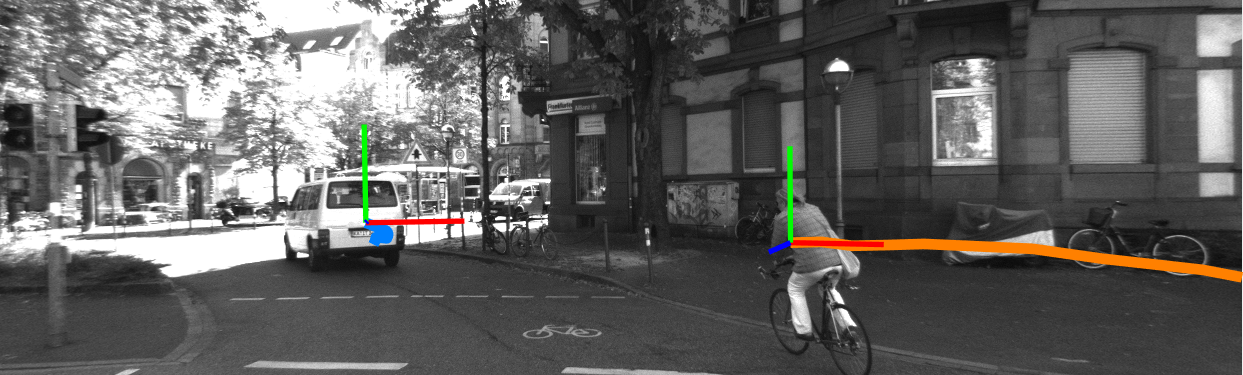}%
		\end{subfigure}\\
		\begin{subfigure}[t]{\columnwidth}
			\includegraphics[clip,width=.98\columnwidth]{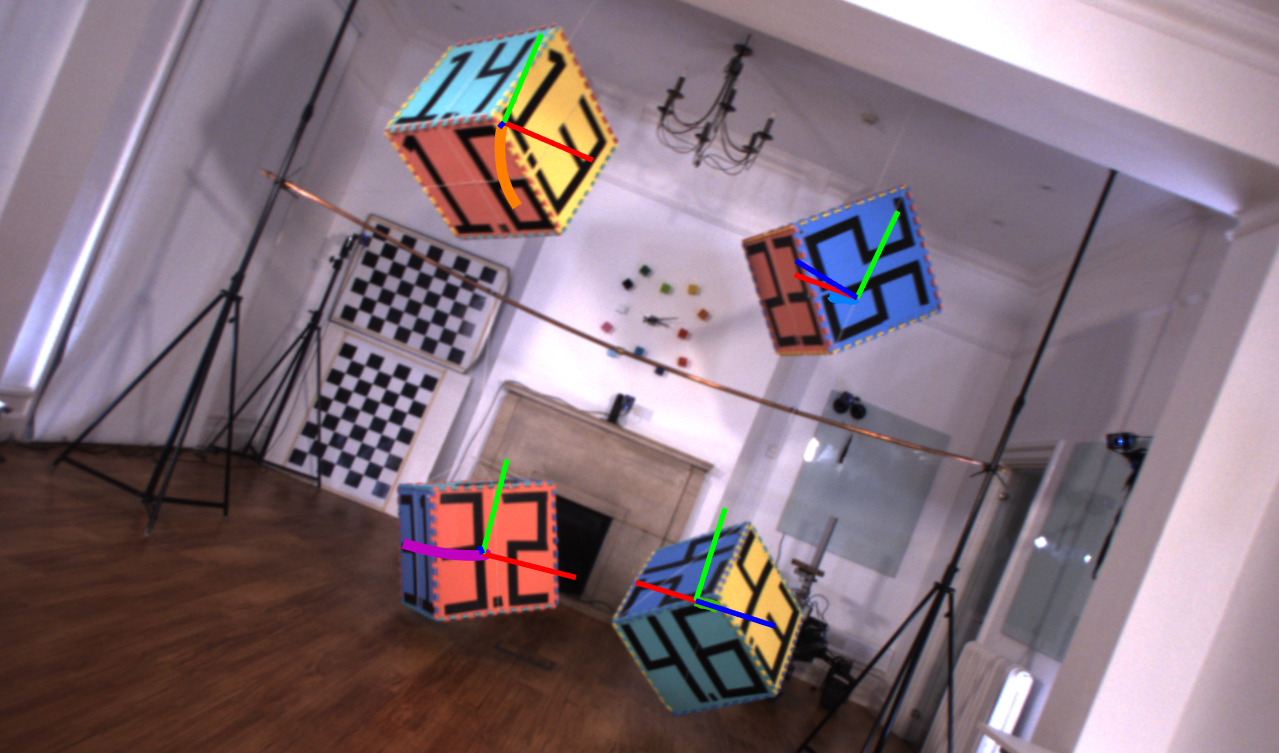}%
		\end{subfigure}
		\caption{
			Motion trajectories estimated by \acs{MVO}	for sequences from the KITTI \citep[top,][]{geiger2012} and \acs{OMD} \citep[bottom,][]{judd2019ral} datasets. 
			The KITTI segment involves an autonomous driving scenario in a residential environment where the car-mounted camera follows a van and a cyclist.
			The \acs{OMD} segment includes four independently swinging blocks observed by a handheld camera.
			In both sequences, the \kjse{3} trajectories of the camera egomotion and third-party motions are estimated simultaneously without prior knowledge of their number, appearance, or nature. 
			\looseness=-1
		}
		\label{fig:marquee}
	\end{figure}
	
	A principle objective of robotic navigation systems is determining the egomotion of a dynamic robot relative to its environment, usually using a mounted sensor.
	\Ac{VO} is a technique for finding the egomotion of a camera by isolating the static parts of a scene and estimating the motion relative to those regions \citep{moravec1980}.
	The segmentation of static portions of the scene is itself an important focus of visual navigation research \citep[e.g.,][]{nister2004,fraundorfer2012}, but much less research has focused on also analyzing the dynamic regions of the scene that the segmentation rejects.
	
	Third-party dynamic motions are more difficult to estimate than the sensor egomotion because their observed motions  comprise both their arbitrary real-world motions and the sensor egomotion.
	This \ac{MEP} is often simplified by constraining motions according to kinematic assumptions \citep[e.g.,][]{sabzevari2016}, or by first isolating and estimating the sensor egomotion and then compensating for it to estimate the remaining third-party motions in the scene \citep[e.g.,][]{jaimez2017}.
	These techniques can be successful in specific applications, but few generalized approaches have been proposed to address the full \ac{MEP}.
	
	Third-party motions are also difficult to track because highly dynamic scenes often include significant occlusions. 
	Occlusions can be defined as any time there are insufficient measurements of part of a scene.
	They are either direct, when an object is partially or fully obscured, or indirect, due to sensor limitations or algorithmic failure.
	\Ac{MOT} techniques can successfully track multiple dynamic objects through full and partial occlusions, but they are often constrained by application-specific object detectors and simplistic motion models \citep[e.g.,][]{milan2016,mitzel2010}.
	Consistently estimating multiple independent motions in the presence of occlusions is necessary for autonomous navigation in complex, dynamic environments.
	
	This paper presents \ac{MVO}, a multimotion estimation pipeline that prioritizes motion over appearance.
	It emphasizes the importance of understanding \emph{how} things are moving in the environment before understanding \emph{what} they are.
	\Ac{MVO} addresses the \ac{MEP} by applying multilabeling techniques to the traditional \ac{VO} pipeline using only a rigid-motion assumption.
	It simultaneously estimates the full \kjse{3} trajectory of every motion in a scene, including the sensor egomotion, without making any \emph{a priori} assumptions about object number, appearance, or motion (\cref{fig:marquee}).
	The pipeline is adaptable to a variety of motion priors and achieves robust motion-based tracking through occlusion via \emph{motion closure}.
	
	The rest of the paper is organized as follows.
	\cref{sec:background} explores the details of the \ac{MEP} and the current approaches for addressing it. 
	\cref{sec:mvo} describes how the \ac{MVO} pipeline extends traditional \ac{VO} techniques to multimotion estimation.
	\cref{sec:estimation} demonstrates how different types of state estimators and motion priors can be used within the \ac{MVO} pipeline.
	\cref{sec:sliding} discusses how to address the challenges involved in implementing \ac{MVO} as a sliding-window estimator, particularly with respect to occlusions.
	\cref{sec:evaluation,sec:discussion} evaluate and discuss the performance of the \ac{MVO} pipeline quantitatively and qualitatively using the \accite{OMD}{judd2019ral} and the KITTI Vision Benchmark Suite \citep{geiger2012}.

	This paper is a continuation of ideas that were first presented at the Joint Industry and Robotics CDTs Symposium \citep{judd2018jircs} and the Long-term Human Motion Prediction Workshop \citep{judd2019lhmp} and were published in \citet{judd2018iros}, \citet{judd2019thesis}, and \citet{judd2020iros}.
	It makes the following specific contributions:
	\begin{itemize}
		\item Presents a unified and updated version of \ac{MVO} that is adaptable to a variety of trajectory representations and estimation techniques. 
		\item Incorporates a continuous \kjse{3} \ac{WNOJ} prior into the estimator, extends it for geocentric third-party estimation, and examines its advantages and limitations in the context of the \ac{MEP}, along with the previously presented discrete and \ac{WNOA} models.
		\item Details the challenges of full-batch and sliding-window implementations of \ac{MVO}, including handling temporary occlusions.
		\item Compares pose-only, pose-velocity, and pose-velocity-acceleration estimators both quantitatively on indoor experiments (\ac{OMD}) and qualitatively in the real world (KITTI).
	\end{itemize}
	
	\section{Background}
	\label{sec:background}
	Dynamic environments consist of the static background, a moving observer (i.e., sensor), and one or more independent, third-party motions. 
	The observed motions include both the objects' real-world motion and the sensor egomotion, so the static scene appears to move with the inverse of the egomotion trajectory.

	Fully addressing this \ac{MEP} requires both \emph{segmentation}, i.e., clustering points according to their movement between observations, and \emph{estimation}, i.e., calculating the motion of a cluster of points.
	The interdependence of these tasks creates a \emph{chicken-and-egg} problem: the motion trajectories must be estimated from groups of points, but the segmentation of those groups depends on the available motion estimates.
	
	This interdependence is addressed in single-motion \ac{VO} systems by using heuristics (e.g., number of features) to select the egomotion and ignore the other motions in the scene (\cref{sec:background:ego}).
	Appearance-based tracking techniques are often used to track multiple dynamic objects, but currently do not accurately estimate the underlying motion of generic objects (\cref{sec:background:mot}).
	These heuristic- and appearance-based techniques are not readily extensible to general \emph{multimotion} estimation problems and analyzing multiple independently moving bodies remains a challenging problem for state-of-the-art vision systems (\cref{sec:background:mep}).
	
	\subsection{Egomotion Estimation}\label{sec:background:ego}
	\ac{VO} estimates the motion of a camera relative to its static environment (i.e., \emph{egomotion}) and has a long history in robotics starting with the motion-estimation pipeline presented by \citet{moravec1980}.
	The camera egomotion is calculated from a stream of images by assuming it corresponds to the dominant motion in the scene.
	
	The egomotion is estimated by first isolating regions of the scene that move according to the dominant motion and assuming they are static.
	Those regions are then used to estimate the egomotion trajectory relative to those static points and the rest of the scene is usually ignored as noise.
	
	Recent developments in \ac{VO} focus on improving the static segmentation \citep{nister2004,sabzevari2016}.
	The estimation can also be improved by using more advanced estimation techniques, such as dense or direct estimation methods \citep{valgaerts2012,engel2017} and continuous \kjse{3} motion priors \citep{anderson2015,tang2019}.
	These techniques make the egomotion estimation more robust and accurate, but few methods extend them to also analyze the dynamic parts of the scene that \ac{VO} rejects.
	\looseness=-1
	
	\subsection{Multiobject Tracking (MOT)}\label{sec:background:mot}
	\ac{MOT} techniques are often used in tandem with \ac{VO} to track multiple dynamic objects over time, usually through a \emph{tracking-by-detection} paradigm \citep{milan2016}.
	They use appearance-based object detectors to find instances of specific object classes in each frame.
	They then seek to accurately associate present and past detections to form simple $\mathbb{R}^3$ trajectories.
	These techniques are limited by the quality of their detectors, and are often constrained to tracking application-specific object classes.
	
	\ac{MOT} algorithms often track objects through occlusions by modeling object interactions \citep{yang2011} or employing simple motion models to extrapolate object positions \citep{mitzel2010}.
	These techniques are limited by the object representations used by the target detectors, and tend to be constrained to simple motion models that do not adequately represent the motion of an object. 
	Fully addressing the \ac{MEP} requires applying more expressive motion estimation techniques, such as those used to estimate egomotion, to the other third-party motions in a scene.
	
	\subsection{Multimotion Estimation}\label{sec:background:mep}
	In contrast to \ac{MOT}, multimotion estimation focuses on the motions in the scene rather than the objects that generate them.
	This involves both segmenting observed points into independent rigid objects and estimating the \kjse{3} trajectories of those objects. 
	Several multimotion estimation approaches have been developed using techniques such as scene flow clustering, \ac{SLAM} frameworks, and energy-minimization, but none has fully addressed the general \ac{MEP}.
	
	\citet{lenz2011} use sparse scene flow to detect and track multiple dynamic objects from a dynamic, vehicle-mounted camera. 
	The approach operates on stereo image pairs and clusters sparse points based on their scene flow.
	These clusters are then used to track objects through $\mathbb{R}^3$ space.
	The approach is not limited to tracking specific classes of objects and is robust to some partial occlusions, but it requires objects to be of limited size and in contact with the ground plane.
	Scene flow also estimates pixel-wise translational motion, so estimating the motion of rotating objects requires further segmentation and estimation.
	
	\citet{menze2015} instead model the world as piecewise-planar \emph{superpixels}.
	The approach simultaneously calculates both scene flow and the homographies defining the motion of the planes in the scene, which are modeled as full \kjse{3} motions.
	Similarly, \citet{quiroga2014} and \citet{jaimez2017} define optimization frameworks that estimate scene flow  by modeling the underlying motions as full \kjse{3} motions.
	This formulation is more robust to rotations but it complicates the \ac{MEP} by introducing additional constraints to estimate the pixel-wise velocity of every point in the scene.
	Each of these techniques also relies on accurate RGB-D sensing and an independent egomotion estimation pipeline, which can lead to errors when large portions of the scene are moving.
	\looseness=-1
	
	\citet{wang2007} extend the traditional \ac{SLAM} formulation to include \ac{MOT} techniques in their \acs{SLAMMOT} framework using lidar sensors and monocular and stereo cameras \citep{lin2010}.
	The traditional \ac{SLAM} state, consisting of the \kjse{3} sensor pose and a static map of the environment, is extended to include the range, bearing, and linear velocity of any tracked objects in the scene.
	Like most 3D \ac{MOT} algorithms, \acs{SLAMMOT} only estimates the $\mathbb{R}^3$ position of third-party objects, rather than their full \kjse{3} pose.
	CubeSLAM \citep{yang2019} similarly extends the \ac{SLAM} framework to track the full \kjse{3} pose of 3D bounding boxes for each detected object, but not the pose of the object itself.
	\looseness=-1

	\begin{figure*}[t]
		\centering
		\def\svgwidth{\textwidth}
		\includegraphics[width=\textwidth]{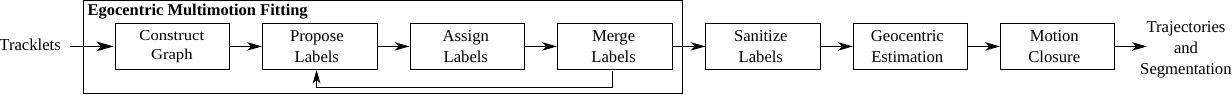}
		\caption{
			An illustration of the stereo \acs{MVO} pipeline, which extends the standard \acs{VO} pipeline by replacing the egomotion estimator with a multimotion estimator.
			\acs{MVO} operates on 3D tracklets and generates the \kjse{3} trajectory for \emph{every} motion in the scene, including the sensor egomotion.
			The pipeline builds a neighborhood graph based on how rigidly pairs of points move over time and iteratively splits and estimates new labels using the graph.
			It assigns labels based on an energy functional, and merges labels that can be considered redundant until convergence. 
			Once the segmentation converges, the labels are sanitized and a batch estimation produces the geocentric \kjse{3} trajectories, employing \emph{motion closure} to determine if newly discovered motions can be explained by the reappearance of an occluded object.
		}
		\label{fig:mvo:pipeline}
	\end{figure*}
	
	Rather than tracking 3D bounding boxes, \citet{huang2019} estimate bulk point motions in ClusterSLAM, a multibody dynamic \ac{SLAM} backend that was extended to a full multimotion estimation pipeline in ClusterVO \citep{huang2020}.
	The pipeline clusters tracked 3D points according to their bulk motion but relies on a semantic object detector to inform its data association and tracking.
	\accite{VDO-SLAM}{henein2020,zhang2020iros,zhang2020arxiv} also extends the \ac{SLAM} formulation to estimate the full \kjse{3} pose of other objects in the scene, but uses an external instance segmentation algorithm to segment the image into its constituent objects first.
	DynaSLAM II \citep{bescos2021} similarly uses front-end instance segmentation to guide feature matching and segmentation before estimating the \kjse{3} trajectory of each tracked object.
	ClusterVO, \ac{VDO-SLAM}, and DynaSLAM II are able to estimate the motion of each object more accurately than techniques that only track bounding boxes, but they are reliant on significant front-end preprocessing, which limits their applicability to the general \ac{MEP}.
	In contrast, DymSLAM \citep{wang2020} is a \ac{SLAM} framework without any significant front-end processing.
	It uses spectral clustering to segment dynamic motions and estimate their \kjse{3} motions but primarily focuses on dense scene reconstruction which is separate from the \ac{MEP}.
	
	Instead of building on the traditional \ac{SLAM} framework, \citet{isack2012} demonstrate an energy-based multimodel-fitting framework that can be used to segment motions in a frame-to-frame manner.
	\citet{roussos2012} use this framework to simultaneously estimate depth maps and object motions from a monocular camera while also performing object tracking.
	The technique minimizes a complex cost function involving photometric consistency, geometric smoothness in depth, spatial smoothness in image space, and a minimum-description-length term that promotes a compact solution.
	The approach proceeds in an offline, full-batch manner and its additional focus on dense reconstruction, combined with its initialization requirements, make it ill-suited for \ac{MEP} applications.
	
	Other techniques use RGB-D sensors to segment and estimate motions while also performing suitable tracking for object reconstruction.
	\citet{runz2017} use an RGB-D camera to segment and track targets while simultaneously fusing 3D object models.
	The technique combines motion segmentation with object instance segmentation, which relies on predefined class-based object detectors.
	\citet{runz2018} extend this work to real-time processing by improving the efficiency of the semantic segmentation, and \citet{xu2019} define a similar system using a volumetric representation, rather than surface normals.
	These techniques represent significant progress in addressing the \ac{MEP}, with the added ability to fuse 3D models of the tracked objects; but they are reliant on high-quality, dense depth sensing and they are limited in the number of active models they can reasonably process.
	\looseness=-1
	
	\citet{qiu2019} and \citet{eckenhoff2019} address the \ac{MEP} using monocular cameras.
	Monocular observations are underconstrained and a separate scale parameter must be estimated using the \ac{IMU}.
	This scale parameter is valid for the egomotion of the camera, but estimating the scale for third-party motions is difficult and leads to several degenerate cases, such as when the object and camera motions are colinear.
	The performance of these techniques is impressive given the limitations of the sensors, but they are not broadly applicable to the general \ac{MEP}.
	
	\vspace{0.65\baselineskip}
	
	In contrast to these approaches, \ac{MVO} fully addresses the \ac{MEP} in an online manner using only a rigid-motion assumption.
	The pipeline estimates the full \kjse{3} trajectory of every motion in a complex, dynamic scene without any \emph{a priori} information about object number, appearance, or motion.
	It also operates directly on sparse 3D points and is therefore applicable to a variety of 3D sensors.
	\ac{MVO} can incorporate physically founded \kjse{3} motion priors to extrapolate motions accurately through occlusions and achieves motion-based tracking through a motion closure procedure.

	\begin{figure*}[t]
		\centering%
		\hfill\begin{subfigure}[]{0.57\textwidth}%
			\includegraphics[clip,width=\textwidth]{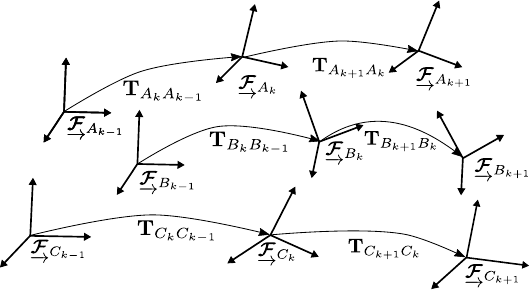}%
		\end{subfigure}\hfill
		\begin{subfigure}[]{0.43\textwidth}%
			\includegraphics[clip,width=\textwidth]{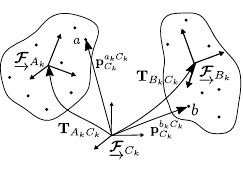}%
		\end{subfigure}\hfill
		\caption[Illustrations of the \acs{MEP} showing the motion of frames through time and the relative point observations]{Illustrations of the \acs{MEP} showing the motion of frames through time (left) and the relative point observations (right). Two independent third-party motions, $\kjframe{A}$ and $\kjframe{B}$, are observed by a moving camera, $\kjframe{C}$, through feature measurements on the objects, $\left\{\kjvecpoint{{a_k}{C_k}}{C_k}\right\}$ and $\left\{\kjvecpoint{{b_k}{C_k}}{C_k}\right\}$. 
			Solving the \acs{MEP} requires simultaneously segmenting and estimating the motions~of~these~measurements.}
		\label{fig:problem}
	\end{figure*}
	
	\section{Simultaneous \texorpdfstring{\kjse{3}}{SE(3)} Estimation and Segmentation}\label{sec:mvo}
	Fully addressing the \ac{MEP} requires both motion estimation and segmentation.
	Motion estimation involves calculating the motion of a set of points.
	Motion segmentation involves clustering points according to their movement. 
	
	The interdependence of estimation and segmentation leads to a \emph{chicken-and-egg} problem: the motion trajectories must be estimated from groups of points, but the segmentation of those groups depends on the available motion estimates.
	In single-motion \ac{VO} systems this is addressed by using heuristics (e.g., number of features) to select the egomotion and ignore the other motions in the scene.
	These heuristics are not readily extensible to \emph{multimotion} estimation problems and analyzing multiple independently moving bodies remains a challenging problem for state-of-the-art vision systems.
	
	\ac{MVO} addresses the \ac{MEP} by casting motion segmentation as a multilabeling problem where a label represents a motion trajectory.
	It iteratively segments and estimates motions using these labels in an alternating fashion until the segmentation converges.
	The set of motion labels adapts to the motions in the scene without making any \emph{a priori} assumptions about object number, appearance, or motion.
	
	\ac{MVO} does not use any assumptions or heuristics to identify the static background until after the segmentation converges, so each motion is estimated as if it corresponds to the true egomotion.
	After the segmentation converges, the true egomotion is selected and used to reestimate the true third-party motions in a geocentric frame (\cref{sec:estimation}).

	The pipeline operates directly on tracked 3D observations, i.e., \emph{tracklets}, from a 3D sensor.
	It iteratively generates labels and segments those observations according to their motion (\cref{fig:mvo:pipeline}).
	The tracklets are embedded in a graph structure that forms the basis of the motion segmentation (\cref{sec:mvo:graph}).  
	New labels are proposed when the motion of tracklets assigned to a given label may be more accurately explained by multiple trajectories (\cref{sec:mvo:proposal}). 
	Motion labels are assigned to each tracklet by minimizing a cost functional that incorporates reprojection residual, graph smoothness, and model complexity (\cref{sec:mvo:assignment}).
	Tracklets whose motions are not well explained by any other label are assigned to an outlier label and redundant and oversegmented labels are then merged (\cref{sec:mvo:assignment:merging}).
	Once the segmentation converges, the  final labels are then sanitized and any remaining outliers are rejected (\cref{sec:mvo:sanitization}).
	Initial versions of these ideas were first presented in \citet{judd2018iros} and \citet{judd2018jircs}, and they are further developed here.
	
	\subsection{Notation}
	A sequence of observations of a point, $j$, relative to a moving sensor frame, $\kjframe{C}$, over multiple time steps forms a \emph{tracklet}
	\vspace{-1.5mm}
	\begin{equation*}
		\kjpointset{j}{} \coloneqq \left(\kjhopoint{j_kC_k}{C_k}\right)_{k=1\dots K} = \left(\begin{bmatrix}\kjvecpoint{j_kC_k}{C_k}\\ 1\end{bmatrix}\right)_{k=1\dots K},
		\vspace{-0.5mm}
	\end{equation*}
	where $C_k$ refers to the observing sensor frame at time $k$ (\cref{fig:problem}, right), and $\kjhopoint{j_kC_k}{C_k} \in \mathbb{R}^{4\times1}$ is the homogeneous representation of the Euclidean vector, $\kjvecpoint{j_kC_k}{C_k} \in \mathbb{R}^{3\times1}$. 
	
	A transform, $\kjtransform{AB}{} \in $ \kjse{3}, relates the current coordinate frame, $\kjframe{A}$, to the origin frame $\kjframe{B}$.
	The transform is defined blockwise,
	\begin{equation}\label{eq:mvo:notation:transform}
		\kjtransform{AB}{} \coloneqq \begin{bmatrix}
			\kjrotation{AB} & \kjvecpoint{BA}{A}\\
			\mathbf{0}^{T} & 1
		\end{bmatrix},
	\end{equation}
	where $\kjrotation{AB}$ is the rotation matrix to $\kjframe{A}$ from $\kjframe{B}$, and $\kjvecpoint{BA}{A}$ is the translation to $\kjframe{B}$ from $\kjframe{A}$ expressed in $\kjframe{A}$.
	This translation can also  be expressed as
	\begin{equation}\label{eq:mvo:notation:rotate}
		\kjvecpoint{BA}{A} = -\kjrotation{AB} \kjvecpoint{AB}{B}.
	\end{equation}
	
	The motion trajectory, $\kjtranset{A}$, consists of a sequence of \kjse{3} transforms,
	\begin{equation*}
		\kjtranset{A} \coloneqq \left(\kjtransform{A_kA_1}{}\right)_{k=2\dots K},
	\end{equation*}
	that relate the pose at time $k$ to a privileged initial pose $\kjframe{A_1}$ (\cref{fig:problem}, left).
	The trajectory can also be modeled continuously as $\kjtransform{A}{}\left(t\right)$ such that $\kjtransform{A_kA_1}{} \coloneqq \kjtransform{A}{}\left(t_k\right)$.
	The frames, $\kjframe{A}$ and $\kjframe{B}$, can be the sensor frame, $\kjframe{C}$, or an arbitrary third-party motion frame, $\kjframe{\kjlabel{}}$.

	The set of tracklets visible in a scene, $\kjtrackletset{}$, comprises the subsets of tracklets assigned to each motion label, $\kjtrackletset{\kjlabel{}}$, i.e., the support for label $\kjlabel{}$	.
	The current set of all motion labels available for assignment is given by $\kjlabelset{}$.
	
	\subsection{Graph Construction}\label{sec:mvo:graph}
	The multilabeling is performed on a graph that represents each tracklet as a vertex.
	The graph, $\kjgraph{}$, embodies the rigid-body assumption by defining the cost, $\kjweight{}_{ij}$, between pairs of vertices as the variance in the $\mathbb{R}^{3}$ distance between their associated tracklets over time,
	\begin{equation}
		\kjweight{}_{ij} \coloneqq \frac{1}{K}\sum_{k=1}^{K} \left(\norm{\kjvecpoint{i_kC_k}{C_k} - \kjvecpoint{j_kC_k}{C_k}} - \bar{d}_{ij}\right)^{2}, 
		\label{eq:mvo:graph:variance}
	\end{equation}
	where $\bar{d}_{ij}$ is the mean distance between $\kjpointset{i}{}$ and $\kjpointset{j}{}$.
	The variance between tracklets moving similarly is small, making this edge cost small for tracklets on the same rigid body.

	The graph consists of vertices for all tracklets and the $k$ least-costly edges from each vertex.
	This $k$-nearest neighbors graph is unlikely to connect independently moving pairs of tracklets so it reasonably and efficiently approximates the rigid-body assumption.
	Its connectivity forms the basis for label generation (\cref{sec:mvo:proposal}) and assignment (\cref{sec:mvo:assignment}), and an over- or under-connected graph can lead to an under- or overfit solution, respectively.
	
	\subsection{Label Proposal}\label{sec:mvo:proposal}
	The motions in the scene are segmented by assigning tracklets to a finite set of motion labels, $\kjlabelset{}$.
	This set must be updated regularly because dynamic environments are constantly changing.
	New labels are proposed by splitting existing labels whenever their motions could be explained by multiple trajectories.
	
	A potential new label, $\kjlabel{}'$, is generated for each fully-disjoint component of the subgraph, $\kjgraph_{\kjlabel{}}\subseteq\kjgraph{}$, defined by the current label support, $\kjtrackletset{\kjlabel{}}$.
	This prioritizes graph smoothness while allowing large label subgraphs consisting of distinct rigid motions to be split into new labels.
	
	\subsubsection{Estimating New Labels}
	Each label is estimated as if it corresponds to the static portion of the scene using standard egomotion techniques.
	The new egomotion hypothesis, ${^{\kjlabel{}'}\kjtranset{C}}$, is calculated from the dominant motion of the corresponding subgraph component using frame-to-frame RANSAC, which segments the inliers and outliers of that motion \citep{fischler1981}.
	\looseness=-1
	
	The frame-to-frame RANSAC procedure estimates the observed motion between pairs of consecutive frames, ${^{\kjlabel{}'}\kjtransform{{{C_{k}}{C_{k-1}}}}{}}$, by sampling three tracklets in both frames and assuming they are static.
	The transform is estimated in a blockwise manner according to \eqref{eq:mvo:notation:transform}.
	
	The translational component is calculated by  
	\begin{equation*}
		{^{\kjlabel{}'}\kjvecpoint{C_{k-1}C_{k}}{C_{k}}} = -{^{\kjlabel{}'}\kjvecpoint{C_{k}C_{k-1}}{C_{k}}}\\
		=  -\left(\kjvecpointmean{}{C_{k}} - {^{\kjlabel{}'}\kjrotation{C_{k}C_{k-1}}} \kjvecpointmean{}{C_{k-1}}\right),
	\end{equation*}
	where $\kjvecpointmean{}{C_{k}}$ is the centroid of the sampled points at time $k$.
	The rotation is found by solving Wahba's problem \citep{wahba1965} using singular value decomposition \citep{markley1988},
	\begin{equation*}
		\begin{aligned}
			\mathbf{U}\boldsymbol{\Sigma}\mathbf{V}^T &\coloneqq \sum_{j=1}^{3}\left(\kjvecpoint{j_{k-1}C_{k-1}}{C_{k-1}} - \kjvecpointmean{}{C_{k-1}}\right)\left(\kjvecpoint{j_{k}C_{k}}{C_{k}} - {\kjvecpointmean{}{C_{k}}}\right)^T,\\
			{^{\kjlabel{}'}\kjrotation{C_{k}C_{k-1}}} &= \mathbf{U}
			\begin{bmatrix}
				1 & 0 & 0\\
				0 & 1 & 0\\
				0 & 0 & |\mathbf{U}
				||\mathbf{V}
				|
			\end{bmatrix}\mathbf{V}^T.
		\end{aligned}
	\end{equation*}
	
	The calculated transform is evaluated by the number of inlier tracklets with reprojection residuals,
	\begin{equation} \label{eq:mvo:proposal:ransac_residual}
		\begin{aligned}
			\hspace{-2.0mm}e_k\hspace{-0.3mm}\left(\kjpointset{j}{}, {^{\kjlabel{}'}\kjtransform{C_{k}C_{k-1}}{}}\right)\hspace{-0.3mm}\coloneqq\hspace{-0.3mm} \Biggl|&\Biggl|\mathbf{s}\hspace{-0.3mm}\left(\kjhopoint{j_{k}C_{k}}{C_{k}}\right)\\
			&-\mathbf{s}\hspace{-0.3mm}\left({^{\kjlabel{}'}\kjtransform{C_{k}C_{k-1}}{}}\kjhopoint{j_{k-1}C_{k-1}}{C_{k-1}}\right)\Biggr|\Biggr|,
		\end{aligned}
	\end{equation}
	less than a chose threshold, $e_{\mathrm{th}}$, where $\mathbf{s}\left(\cdot\right)$ applies the sensor projection (e.g., nonlinear stereo camera; \cref{app:stereo}).
	
	The process of sampling, estimation, and evaluation is iterated a chosen number of times, $N_{\rm{RSAC}}$, and the transform with the most inliers is kept.
	This process is repeated independently for each pair of consecutive frames in the estimation window to generate a new trajectory for the label.
	
	This frame-to-frame procedure is used to propose a new trajectory for each extant label.
	Any tracklets found to be outliers of the newly estimated trajectories are appended to the outlier label, $\kjoutlier{}$.
	After the outliers are removed from all existing label sets, new labels are generated from the outlier label using the same proposal process.
	
	\subsection{Label Assignment}\label{sec:mvo:assignment}
	Casting the \ac{MEP} as a multilabeling problem involves assigning each tracklet, $\kjpointset{}{}\in\kjtrackletset{}$, to a motion label, $\kjlabel{}\in\kjlabelset{}$, such that some energy functional, $E\left(\kjlabelset{}\right)$, is minimized.
	This functional incorporates both the fidelity of a label to the motion of a tracklet and a piecewise-rigid model of the environment.
	A complexity term is included to penalize using additional labels and incentivize a compact solution.
	The energy functional balances these residual error (\cref{sec:mvo:assignment:residual}), label smoothness (\cref{sec:mvo:assignment:smoothness}), and label complexity (\cref{sec:mvo:assignment:complexity}) terms,
	\begin{equation} \label{eq:mvo:assignment_cost}
		\begin{aligned}
			E\left(\kjlabelset{}\right) \coloneqq& \underbrace{\sum_{\kjpointset{}{}\in\kjtrackletset{}} \rho\left(\kjpointset{}{}, \kjlabel{}\left(\kjpointset{}{}\right)\right)}_\text{Residual} \\
			&+ \lambda_{\mathrm{sm}}\underbrace{\sum_{\left(\kjpointset{i}{},\kjpointset{j}{}\right)\in\kjgraph} \exp\left(-\omega_{ij}\right)V\left(\kjpointset{i}{},\kjpointset{j}{}\right)}_\text{Smoothness} \\
			&+ \underbrace{\sum_{\kjlabel{}\in\kjlabelset{}} \mu_{\kjlabel{}}\psi\left(\kjlabel{}\right)}_\text{Complexity},
		\end{aligned}
	\end{equation}
	where $\kjlabel{}\left(\kjpointset{}{}\right)$ gives the label currently assigned to $\kjpointset{}{}$ and $\lambda_{\mathrm{sm}}$ is a user-selected proportionality parameter. 
	Increasing the proportionality parameter can lead to a spatially smoother solution at the cost of ignoring small, but clearly distinct, motions.
	Decreasing the parameter can isolate small motions but can also lead to more spurious motion detections.
	
	The residual and smoothness portions of the energy are minimized using the \accite{CORAL}{amayo2018}, a multimodel-fitting algorithm that relaxes discrete labeling to allow for efficient parallelization on GPGPU devices.
	This ``soft'' labeling assigns each tracklet a score for how applicable each label is to the observed data.
	The score is in the continuous range $\left[0,1\right]$ and the scores for each tracklet across all labels sum to $1$.
	The soft labels are discretized by taking the strongest label for each point.
	Points with ambiguous soft labels (e.g., less than $0.5$) are labeled as outliers to avoid mislabeling points.
	The total energy is finally minimized by merging similar labels to avoid oversegmentation and overfitting (\cref{sec:mvo:assignment:merging}). 
	
	\subsubsection{Residual Cost}\label{sec:mvo:assignment:residual}
	The residual term,
	\begin{equation*} 
		\sum_{\kjpointset{}{}\in\kjtrackletset{}} \rho\left(\kjpointset{}{}, \kjlabel{}\left(\kjpointset{}{}\right)\right),
	\end{equation*}
	penalizes labels that poorly describe their assigned data. 
	It is defined as the sum of the residual errors in applying the label trajectories to tracklets. 
	The residual for each point-label pair is defined as
	\begin{equation*}
		\rho\left(\kjpointset{}{}, \kjlabel{}\right) \coloneqq \max\limits_{k \in t_1\dots t_2} e_k\left(\kjpointset{}{},{^{\kjlabel{}}\kjtransform{C_{k}C_{j}}{}}\right),
	\end{equation*}
	where $e_k$ is defined in \eqref{eq:mvo:proposal:ransac_residual} and $t_1$ and $t_2$ are the first and last time stamps that the tracklet is observed, respectively.
	Maximum error penalizes tracklets that do not fit a motion, even for a single frame.
	This conservative labeling helps preserve the accuracy of the trajectory estimation.
	
	The residual cost of the outlier label, $\kjoutlier{}$, is uniquely defined to have low cost for points not well explained by existing labels.
	The cost decays exponentially with the best-fitting label cost, 
	\begin{equation*}
		\rho\left(\kjpointset{}{}, \kjoutlier{}\right) \coloneqq \alpha\exp\left(-\frac{1}{\beta}\min\limits_{\kjlabel{} \in \kjlabelset{}}\rho\left(\kjpointset{}{}, \kjlabel\right)\right),
	\end{equation*}
	where $\alpha$ and $\beta$ are tuning parameters. 
	The outlier residual cost is high for tracklets whose motions are well-explained by extant labels and low for tracklets with high costs for all labels.

	\subsubsection{Smoothness Cost}\label{sec:mvo:assignment:smoothness}
	The smoothness term,
	\begin{equation*}
		\sum_{\left(\kjpointset{i}{},\kjpointset{j}{}\right)\in\kjgraph} \exp\left(-\omega_{ij}\right)V\left(\kjpointset{i}{},\kjpointset{j}{}\right),
	\end{equation*}
	penalizes neighboring tracklets that do not share the same label by the inverse exponential of their edge cost, $\omega_{ij}$. 
	This encourages a piecewise-rigid solution that is appropriate for most dynamic scenes consisting of contiguous rigid bodies. 
	The cost is a weighted sum of all edges penalized according to
	\looseness=-1
	\begin{equation*}
		V\left(\kjpointset{i}{},\kjpointset{j}{}\right) \coloneqq 
		\begin{cases}
			1 & \mathrm{if} \enskip \kjlabel{}\left(\kjpointset{i}{}\right) \neq \kjlabel{}\left(\kjpointset{j}{}\right) \\
			0 & \mathrm{otherwise}
		\end{cases}.
	\end{equation*}
	This definition of $V$ mirrors the standard Potts model \citep{boykov1999}. 
	The outlier label smoothness cost is the same as all other labels.
	
	\subsubsection{Complexity Cost}\label{sec:mvo:assignment:complexity}
	The complexity term,
	\begin{equation*} 
		\sum_{\kjlabel{}\in\kjlabelset{}} \mu_{\kjlabel{}}\psi\left(\kjlabel{}\right),
	\end{equation*}
	encourages a compact solution by penalizing the use of each label by some cost.
	This prioritizes a \emph{minimum description length}, which favors compactness over complexity and avoids overfitting to the data.
	It is the sum of the per-label cost of each label with nonzero support according to the function,
	\begin{equation*}
		\psi\left(\kjlabel{}\right) \coloneqq
		\begin{cases}
			1 & \mathrm{if} \enskip \vert\kjtrackletset{\kjlabel{}}\vert > 0 \\
			0 & \mathrm{otherwise}
		\end{cases}.
	\end{equation*}
	
	The cost of using a label, $\mu_{\kjlabel{}}$, is an open design decision.
	All motion labels can be assigned the same cost or label costs can be designed to encourage common motions and penalize kinematically complex motions if prior information about the environment is known, such as the sensor platform or third-party kinematics.
	A privileged ``playbook'' of likely motions can also be included in the label set as default options and assigned low costs.
	The label cost, $\mu_{\kjoutlier{}}$, for the outlier label is zero because outliers are assumed to always exist.
	Significantly decreasing the label weights can lead to an overfit solution, while increasing them can make it difficult to segment small motions.
	
	\subsubsection{Label Merging}\label{sec:mvo:assignment:merging}
	Label merging reduces the total energy in \eqref{eq:mvo:assignment_cost} by addressing the model complexity term.
	Two labels, $\kjlabel{}$ and $\kjlabel{}'$, are merged if relabeling all tracklets from one label,  $\kjtrackletset{\kjlabel{}'}$, to the other label, $\kjlabel{}$, would decrease the total energy.
	This occurs when the increase in residual error is less than the cost of using the label and any change in smoothness.
	When more than one merge would reduce the total energy, the merge that results in the greatest decrease in cost is chosen. 
	Merging continues until no more merges would reduce the total energy in \eqref{eq:mvo:assignment_cost}. 
	The outlier label, $\kjoutlier$, is excluded from merging.
	
	\vspace{\baselineskip}
	
	The label splitting, assignment, and merging stages (\cref{sec:mvo:proposal,sec:mvo:assignment}) are iterated until the labels converge or a maximum number of iterations, $\kjvarconvergence{}$, have been reached. 
	Label convergence can be defined relative to various criteria, such as the total energy, the change in energy, or the change in the labeling.
	
	\subsection{Label Sanitization}\label{sec:mvo:sanitization}
	Once the label set has converged, the final segmentation is refined to remove noisy tracklets before the final estimation. 
	Individual tracklets whose residual error is greater than the threshold, $e_{\mathrm{th}}$, are relabeled as outliers, regardless of the potential change in energy.
	Likewise, any label with fewer than a minimum number of support tracklets, $N_{\mathrm{th}}$, or that exists for fewer than a minimum number of frames,  $K_\text{th}$, is removed and its tracklets are relabeled as outliers.
	These thresholds remove spurious labels describing small sets of tracklets or brief bulk motions and must be at least 3 points and  2 frames, respectively.
	This label santization provides a consistent final set of tracklets for the batch estimation of each motion. 
	
	\section{Batch \texorpdfstring{\kjse{3}}{SE(3)} Estimation}\label{sec:estimation}
	The \kjse{3} trajectories of the sensor and each object in the scene are estimated from the segmented tracklets.
	Batch trajectory estimation techniques traditionally calculate the motion of a sensor relative to a set of points assuming they are static.
	This is appropriate for egomotion estimation (\cref{sec:estimation:ego}), but third-party trajectory estimation requires extending these techniques to include the geocentric (i.e., quasi-inertial) frame defined by the egomotion estimation (\cref{sec:estimation:geo}).
	
	Initially, each segmented motion represents an egomotion trajectory hypothesis, ${^{\kjlabel{}}\kjtransform{{{C_{k}}{C_{k-1}}}}{}}$, under the assumption that the associated tracklets, $\kjtrackletset{\kjlabel{}}$, are static.
	Once the true static label is identified, its corresponding egomotion estimate can be used to estimate the remaining third-party motions in a geocentric frame.
	
	These geocentric trajectories can be estimated without making any assumptions about the object or sensor motion other than rigid motion.
	The estimation can be made more robust by using more expressive motion models that constrain the trajectories to the real-world kinematics of the scene.
	Examples of generalized, physically founded \kjse{3} motion models include the \ac{WNOA} prior described by \citet{anderson2015} and the \ac{WNOJ} prior described by \citet{tang2019}.
	The \ac{WNOA} prior extends the pose-only trajectory state to also include velocity and the \ac{WNOJ} prior extends it to include both velocity and acceleration.
	
	The applicability of any motion model depends on how well it matches the motions in an environment.
	The pose-only estimator (\cref{sec:estimation:ego:transform,sec:estimation:geo:transform}) can estimate complex rigid motions but can also generate physically implausible motions.
	The pose-velocity (\cref{sec:estimation:ego:wnoa,sec:estimation:geo:wnoa}) and pose-velocity-acceleration (\cref{sec:estimation:ego:wnoj,sec:estimation:geo:wnoj}) estimators penalize deviation from a constant body-centric velocity or acceleration, respectively.
	They are generally applicable because objects tend to move smoothly through their environment, but they are not robust to motions that deviate significantly from these assumptions.
	Each estimator brings advantages and disadvantages in the context of the \ac{MEP}, and this investigation shows how \ac{MVO} can leverage different estimators in different applications (\cref{sec:evaluation,sec:discussion}).
	
	\subsection{Egomotion Estimation}\label{sec:estimation:ego}
	The egomotion label, $C$, may be selected by a variety of heuristics.
	It can be chosen as the label with the largest support, as in \ac{VO}, or with the most similar motion to the previous egomotion estimate.
	This can be combined with external sensors or application-specific heuristics, such as attention masks that prioritize parts of the field of view that usually contain static background objects, to robustly identify the true egomotion label.
	
	Once identified, the point measurements in the egomotion label are assumed to be static.
	The trajectory is therefore
	estimated relative to a geocentric frame and inertial motion priors can be used in the estimation.
	
	\subsubsection{Pose-Only Estimator}\label{sec:estimation:ego:transform}
	A discrete single-motion bundle adjustment \citep[e.g.,][]{barfoot2017} defines the egomotion state, $\kjstate{}$, to include both the estimated pose transforms, $\kjtranset{C} \coloneqq \left({\kjtransform{C_{k}C_{1}}{}}\right)_{k=2\dots K}$, and the landmark points, $\left\{\kjvecpoint{{j_1}{C_1}}{C_1}\right\}_{j=1\dots\vert\kjtrackletset{C}\vert}$. 
	The state $\kjstate{}_{jk} \coloneqq \left\{{\kjtransform{C_{k}C_{1}}{}}, \kjvecpoint{{j_1}{C_1}}{C_1}\right\}$ is defined for each pair of transforms and points in label $C$, i.e., $\kjtransform{C_{k}C_{1}}{} \in \kjtranset{C}$ and $\kjvecpoint{{j_1}{C_1}}{C_1} \in \kjpointset{C}{}$. 
	The batch size, $K$, is the length of the estimation window, which can either be the entire sequence, or some fixed-length sliding window (\cref{sec:sliding}).
	
	Each observation, $\mathbf{y}_{jk}$, of point, $\kjpoint{j}$, at pose, ${\kjtransform{C_{k}C_{1}}{}}$, is modeled as
	\vspace{-2.5mm}
	\begin{equation}
		\begin{aligned}
			\mathbf{y}_{jk} &\coloneqq \mathbf{g}\left(\kjstate{}_{jk}\right) + \mathbf{n}_{jk} = \mathbf{s}\left(\mathbf{z}\left(\kjstate{}_{jk}\right)\right) + \mathbf{n}_{jk} \\
			&= \mathbf{s}\left({\kjtransform{C_{k}C_{1}}{}}\kjhopoint{{j_1}{C_1}}{C_1}\right) + \mathbf{n}_{jk}.
		\end{aligned}
		\label{eq:estimation:ego:measurement}
		\vspace{-1.75mm}
	\end{equation}
	The measurement model, $\mathbf{g}\left(\cdot\right)$, encompasses both the motion model, $\mathbf{z}\left(\cdot\right)$, which applies the \kjse{3} motion of the camera, and the sensor model, $\mathbf{s}\left(\cdot\right)$. 
	The model assumes additive Gaussian noise, $\mathbf{n}_{jk}$, with zero mean and covariance $\mathbf{R}_{jk}$. 
	The least-squares cost function is defined as
	\vspace{-1.75mm}
	\begin{equation}\label{eq:estimation:ego:transform:cost}
		J\left(\mathbf{x}\right)\coloneqq \frac{1}{2}\sum_{j=1}^{\vert\kjtrackletset{C}\vert}\sum_{k=1}^{K}\mathbf{e}_{y,jk}\left(\kjstate{}\right)^{T}\mathbf{R}_{jk}^{-1}\mathbf{e}_{y,jk}\left(\kjstate{}\right),
		\vspace{-3mm}
	\end{equation}
	where,
	\vspace{-1mm}
	\begin{equation*}
		\mathbf{e}_{y,jk}\left(\kjstate{}\right) \coloneqq \mathbf{y}_{jk} - \mathbf{g}\left(\kjstate_{jk}\right),
		\vspace{-1mm}
	\end{equation*}
	is the residual between the observed and modeled tracklet position.
	
	This cost is linearized about an operating point, $\kjstate{}_{\mathrm{op}}$, and then minimized using Gauss-Newton. 
	The linearized cost function is 
	\vspace{-3mm}
	\begin{equation*}
		\begin{aligned}
			J\left(\kjstate{}\right) &\approx J(\kjstate{}_{\mathrm{op}}) - \mathbf{b}^T\delta\kjstate{}_{jk} + \frac{1}{2}\delta\kjstate{}_{jk}^T\mathbf{A}\delta\kjstate{}_{jk},\\
			\mathbf{b} &= \sum_{j=1}^{\vert\kjtrackletset{C}\vert}\sum_{k=1}^{K}\mathbf{P}_{jk}^{T}\mathbf{G}_{jk}^{T}\mathbf{R}_{jk}^{-1}\mathbf{e}_{y,jk}\left(\kjstate{}_\mathrm{op}\right),\\
			\mathbf{A} &=  \sum_{j=1}^{\vert\kjtrackletset{C}\vert}\sum_{k=1}^{K}\mathbf{P}_{jk}^{T}\mathbf{G}_{jk}^{T}\mathbf{R}_{jk}^{-1}\mathbf{G}_{jk}\mathbf{P}_{jk}.
		\end{aligned}
		\vspace{-2mm}
	\end{equation*}
	
	The Jacobian, $\mathbf{G}_{jk}$, of the measurement model, $\mathbf{g}\left(\cdot\right)$, at $\kjstate{}_{jk}$ is given by
	\vspace{-3.75mm}
	\begin{align}\label{eq:estimation:ego:transform:jacobian}
		\mathbf{G}_{jk} &\coloneqq \mathbf{S}_{jk}\mathbf{Z}_{jk}, = \frac{\partial\mathbf{s}}{\partial\mathbf{z}}\bigg\rvert_{\mathbf{z}(\kjstate{}_{\mathrm{op},jk})}\frac{\partial\mathbf{z}}{\partial\kjstatepert{}}\bigg\rvert_{\kjstate{}_{\mathrm{op},jk}},\\
		\nonumber\frac{\partial\mathbf{z}}{\partial\kjstatepert{}}\bigg\rvert_{\kjstate{}_{\mathrm{op},jk}} &= \begin{bmatrix}
			\left({\kjtransform{\mathrm{op},{C_k}{C_1}}{}}\kjhopoint{j_1C_1}{\mathrm{op},C_1}\right)^{\odot} & {\kjtransform{\mathrm{op},{C_k}{C_1}}{}}\mathbf{D}
		\end{bmatrix},\\
		\nonumber\mathbf{D} &= \begin{bmatrix}
			\kjidentity{}\\
			\kjzero{}^T
		\end{bmatrix},
	\end{align}
	where the operator $\left(\cdot\right)^{\odot} \colon \mathbb{R}^{4\times1} \to \mathbb{R}^{6\times6}$ is defined as
	\vspace{-2mm}
	\begin{equation*}
		\kjhopoint{\odot}{} = \begin{bmatrix}
			k\kjvecpoint{}{}\\
			k
		\end{bmatrix}^{\odot} \coloneqq \begin{bmatrix}
			k\kjidentity{} & -\kjvecpoint{\times}{}\\
			\kjzero{}^T & \kjzero{}^{T}
		\end{bmatrix}
	\end{equation*}
	and $\left(\cdot\right)^{\times} \colon \mathbb{R}^{3\times1} \to \mathbb{R}^{3\times3}$ is the skew-symmetric matrix operator,
	\begin{equation*} 
		\kjvecpoint{\times}{} = \begin{bmatrix}
			x\\
			y\\
			z
		\end{bmatrix}^{\times} \coloneqq \begin{bmatrix}
			0 & -z & y\\
			z & 0 & -x\\
			-y & x & 0
		\end{bmatrix}.
	\end{equation*}
	
	The state perturbation, $\kjstatepert{}$, consists of the transform perturbations, $\{\delta\kjseel{}{k} \in \mathbb{R}^{6}\}$, and landmark perturbations, $\{\delta\boldsymbol{\zeta}_j \in \mathbb{R}^{3}\}$. 
	The indicator matrix $\mathbf{P}_{jk}$ is defined such that $\kjstatepert{}_{jk} = \mathbf{P}_{jk}\kjstatepert{}$ is the perturbation of the $\left\{\kjtransform{\mathrm{op},C_kC_1}{}, \kjvecpoint{j_1C_1}{\mathrm{op},C_1}\right\}$ pair.
	
	The optimal perturbation, $\kjstatepert{}^*$, for minimizing the linearized cost function is the solution to the linear equation
	$\mathbf{A}\kjstatepert{}^* = \mathbf{b}$. 
	Each element of the state is then updated according to
	\begin{equation}\label{eq:estimation:ego:transform:update}
		\begin{aligned}
			{\kjtransform{\mathrm{op},{C_k}{C_1}}{}} &\leftarrow\exp(\delta\kjseel{k}{}^{*^\wedge}){\kjtransform{\mathrm{op},{C_k}{C_1}}{}}\\ 
			\kjvecpoint{j_1C_1}{\mathrm{op},C_1} &\leftarrow \kjvecpoint{j_1C_1}{\mathrm{op},C_1} + \mathbf{D}\delta\boldsymbol{\zeta}^{*}_{j},
		\end{aligned}
	\end{equation}
	where $\left(\cdot\right)^{\wedge} \colon \mathbb{R}^{6\times1} \to \mathbb{R}^{4\times4}$ is the \kjse{3} lift operator,
	\begin{equation}\label{eq:estimation:ego:transform:wedge}
		\kjseel{}{}^{\wedge} = \begin{bmatrix}
			\kjsetran{}{}\\
			\kjsoel{}
		\end{bmatrix}^\wedge \coloneqq \begin{bmatrix}
			\kjsoel{}^\times & \kjsetran{}{}\\
			\mathbf{0}^T & 1
		\end{bmatrix}.
	\end{equation}
	
	The cost function is then relinearized about the updated operating point and the process iterates until convergence.
	Convergence is user defined and can be a threshold on the number of iterations, the total cost, the change in cost, the magnitude of the update, or other criteria.
	
	\subsubsection{Pose-Velocity Estimator}\label{sec:estimation:ego:wnoa}
	The pose-only estimator is applicable to any rigid-body motion, but is not constrained to physically plausible motions.
	This can be mitigated by constraining the trajectory to a locally constant velocity.
	
	The pose-velocity estimator employs an \kjse{3} \ac{WNOA} prior \citep{anderson2015},
	\begin{equation*}
		\kjacceleration{C}\left(t\right) \sim \mathcal{GP}\left(\mathbf{0},\mathbf{Q}_c\delta\left(t-t'\right)\right),
	\end{equation*}
	which defines the trajectory acceleration, $\kjacceleration{C}$, as a continuous zero-mean, white noise Gaussian process with power spectral density matrix, $\mathbf{Q}_c \in \mathbb{R}^{6\times6}$.
	The prior models the velocity as being locally constant, but can vary if there are sufficient observations to support it.

	The pose-velocity estimator state, $\kjstate{}$, comprises the estimated pose transforms, $\kjtranset{C}$, body-centric velocities, $\left(\kjvelocity{C_k}\right)_{k=1\dots K}$, and the labeled landmark points, $\left\{\kjvecpoint{{j_1}{C_1}}{C_1}\right\}_{j=1\dots\vert\kjtrackletset{C}\vert}$.
	The state $\kjstate{}_{k} \coloneqq \left\{\kjtransform{C_{k}C_{1}}{}, \kjvelocity{C_k}\right\}$ is defined as the transform and velocity at each time, $t_k$, and $\kjstate{}_{jk} \coloneqq \left\{\kjstate{}_{k},  \kjvecpoint{{j_1}{C_1}}{C_1}\right\}$ is defined for each pair of time steps and points in the label. 
	
	The prior is characterized by the transition function,
	\begin{equation}\label{eq:estimation:ego:wnoa:transition}
		\mathbf{\Phi}\left(\tau,t_{k}\right) \coloneqq \begin{bmatrix}
			\mathbf{1} & \Delta t_{\tau,k}\mathbf{1} \\
			\mathbf{0} & \mathbf{1}
		\end{bmatrix},
	\end{equation}
	where $\Delta t_{\tau,k} = \tau-t_k$.
	
	The transition function propagates the local prior state, ${\boldsymbol{\gamma}}_{C_{k}}$, from time $t_k$ to time $\tau$ according to
	\begin{align}\label{eq:estimation:ego:wnoa:transition_prior}
		\boldsymbol{\gamma}_{C_{k}}\left(\tau\right) &= \mathbf{\Phi}\left(\tau,t_k\right){\boldsymbol{\gamma}}_{C_{k}}\left(t_k\right),
		\intertext{where}
		\nonumber\boldsymbol{\gamma}_{C_{k}}\left(\tau\right) &\coloneqq \begin{bmatrix}
			\kjtransform{C_k}{}\left(\tau\right) \\
			\kjvelocity{C_k}\left(\tau\right)
		\end{bmatrix}.
	\end{align}

	The least-squares estimate is found by minimizing the combined objective function,
	\begin{equation}\label{eq:estimation:ego:wnoa:combined_cost}
		J\left(\kjstate{}\right) \coloneqq J_{y}\left(\kjstate{}\right) + J_{p}\left(\kjstate{}\right),
	\end{equation}
	consisting of the measurement and prior terms.
	The measurement term, $J_y\left(\kjstate{}\right)$, constrains the estimated state by the observations, as given in \eqref{eq:estimation:ego:transform:cost}.
	The prior term, $J_p\left(\kjstate{}\right)$, constrains the trajectory and velocity estimates by the constant-velocity assumption, 
	\begin{equation}\label{eq:estimation:ego:wnoa:jacobian}
		J_{p}\left(\kjstate{}\right) \coloneqq \frac{1}{2}\sum_{k=1}^{K-1}\mathbf{e}_{p,k}\left(\kjstate{}\right)^{T}\mathbf{Q}_{k}\left(t_{k+1}\right)^{-1}\mathbf{e}_{p,k}\left(\kjstate{}\right),
	\end{equation}
	where the error penalizes deviation from the constant-velocity prior defined in \eqref{eq:estimation:ego:wnoa:transition_prior},
	\begin{equation*}
		\mathbf{e}_{p,k}\left(\kjstate{}\right) \coloneqq \boldsymbol{\gamma}_{C_k}\left(t_{k+1}\right) -  \boldsymbol{\Phi}\left(t_{k+1},t_k\right)\boldsymbol{\gamma}_{C_k}\left(t_k\right),
	\end{equation*}
	and the block covariance matrix,
	\begin{equation}\label{eq:estimation:ego:wnoa:covariance}
		\mathbf{Q}_{k}\left(\tau\right) \coloneqq \begin{bmatrix}
			\frac{1}{3}\Delta t_{\tau,k}^{3}\mathbf{Q}_c & \frac{1}{2}\Delta t_{\tau,k}^{2}\mathbf{Q}_c \\
			\frac{1}{2}\Delta t_{\tau,k}^{2}\mathbf{Q}_c & \Delta t_{\tau,k}\mathbf{Q}_c
		\end{bmatrix},
	\end{equation}
	has the inverse,
	\begin{equation*}
		\mathbf{Q}_{k}^{-1}\left(\tau\right) \coloneqq \begin{bmatrix}
			12\Delta t_{\tau,k}^{-3}\mathbf{Q}_c^{-1} & -6\Delta t_{\tau,k}^{-2}\mathbf{Q}_c^{-1} \\
			-6\Delta t_{\tau,k}^{-2}\mathbf{Q}_c^{-1} & 4\Delta t_{\tau,k}^{-1}\mathbf{Q}_c^{-1}
		\end{bmatrix}.
	\end{equation*}
	
	The total cost is minimized by linearizing the error about an operating point, $\kjstate{}_{\mathrm{op}}$, 
	\begin{equation}\label{eq:estimation:ego:wnoa:linear_cost} 
		J\left(\kjstate\right) \approx J(\kjstate{}_{\mathrm{op}}) - \mathbf{b}^T\kjstatepert{} +\frac{1}{2}\kjstatepert{}^T\mathbf{A}\kjstatepert{},
	\end{equation}
	where,
	\begin{equation*}
		\begin{aligned}
			\mathbf{b} &=\hspace{-1mm} \sum_{j=1}^{\vert\kjtrackletset{C}\vert}\sum_{k=1}^{K}\mathbf{P}_{jk}^{T}\mathbf{G}_{jk}^{T}\mathbf{R}_{jk}^{-1}\mathbf{e}_{y,jk}\left(\kjstate{}_\mathrm{op}\right) \\
			&\qquad\qquad+ \sum_{k=1}^{K-1}\mathbf{P}_{k}^{T}\mathbf{E}_{k}^{T}\mathbf{Q}_{k}^{-1}\mathbf{e}_{p,k}\left(\kjstate{}_\mathrm{op}\right),\\
			\mathbf{A}\hspace{-0.5mm}&= \hspace{-1mm} \sum_{j=1}^{\vert\kjtrackletset{C}\vert}\sum_{k=1}^{K}\mathbf{P}_{jk}^{T}\mathbf{G}_{jk}^{T}\mathbf{R}_{jk}^{-1}\mathbf{G}_{jk}\mathbf{P}_{jk}\hspace{-0.7mm}+\hspace{-1mm} \sum_{k=1}^{K-1}\mathbf{P}_{k}^{T}\mathbf{E}_{k}^{T}\mathbf{Q}_{k}^{-1}\mathbf{E}_{k}\mathbf{P}_{k},
		\end{aligned}
	\end{equation*}
	and $\mathbf{G}_{jk}$ and $\mathbf{E}_{k}$ are the Jacobians of the measurement and prior error functions, respectively.

	The Jacobian of the measurement model, $\mathbf{G}_{jk}$, is defined according to \eqref{eq:estimation:ego:transform:jacobian} with the updated motion model Jacobian,
	\begin{equation*}
		\frac{\partial\mathbf{z}}{\partial\kjstatepert{}}\bigg\rvert_{\kjstate{}_{\mathrm{op},jk}} = \begin{bmatrix}
			\left({\kjtransform{\mathrm{op},{C_k}{C_1}}{}}\kjhopoint{\mathrm{op},C_1}{{j_1}{C_1}}\right)^{\odot} & \mathbf{0} & {\kjtransform{\mathrm{op},{C_k}{C_1}}{}}\mathbf{D}
		\end{bmatrix}.
	\end{equation*}
	
	The Jacobian of the prior error function,  $\mathbf{E}_{jk}$, is
	\begin{equation*}
		\begin{aligned}
			\mathbf{E}_{k}=&
			\begin{bmatrix}
				\mathbf{E}_{11} & \Delta t_{k+1,k}\mathbf{1} & \mathbf{E}_{13} & \mathbf{0}\\
				\mathbf{E}_{21} & \mathbf{1} & \mathbf{E}_{23} & \mathbf{E}_{24}
			\end{bmatrix},\\
			\mathbf{E}_{11}=&\boldsymbol{\mathcal{J}}^{-1}_{C_{k+1}C_{k}}\kjadjoint{C_{k+1}C_{k}},\\
			\mathbf{E}_{13}=&-\boldsymbol{\mathcal{J}}^{-1}_{C_{k+1}C_{k}},\\
			\mathbf{E}_{21}=&\frac{1}{2}\kjvelocity{C_{k+1}}^{\curlywedge}\boldsymbol{\mathcal{J}}^{-1}_{C_{k+1}C_{k}}\kjadjoint{C_{k+1}C_{k}},\\
			\mathbf{E}_{23}=&-\frac{1}{2}\kjvelocity{C_{k+1}}^{\curlywedge}\boldsymbol{\mathcal{J}}^{-1}_{C_{k+1}C_{k}},\\
			\mathbf{E}_{24}=&-\boldsymbol{\mathcal{J}}^{-1}_{C_{k+1}C_{k}}.
		\end{aligned}
	\end{equation*}
	where the operator $\left(\cdot\right)^{\curlywedge} \colon \mathbb{R}^{6\times1} \to \mathbb{R}^{6\times6}$ is defined as
	\begin{equation*}
		{\kjseel{}{}}^{\curlywedge} \coloneqq \begin{bmatrix}
			\kjsetran{}{}\\
			\kjsoel{}
		\end{bmatrix}^{\curlywedge} = \begin{bmatrix}
			\kjsoel{}^{\wedge} & \kjsetran{}{}^{\times}\\
			\kjzero & \kjsoel{}^{\wedge}
		\end{bmatrix},
	\end{equation*}
	using the $SO\left(3\right)$ lift operator, $\left(\cdot\right)^{\wedge} \colon \mathbb{R}^{3\times1} \to \mathbb{R}^{3\times3}$, which is equivalent to the skew-symmetric operator, $\left(\cdot\right)^{\times}$.
	
	The adjoint of $\kjtransform{C_{k+1}C_{k}}{}$ is defined as
	\begin{equation*}
		\begin{aligned}
			\kjadjoint{C_{k+1}C_{k}} &\coloneqq \exp\left({\kjseel{C_{k+1}}{C_{k}}}^{\curlywedge}\right)\\
			&= \begin{bmatrix}
				\kjrotation{C_{k+1}C_{k}} & \left(\kjvecpoint{C_kC_{k-1}}{C_{k-1}}\right)^{\times}\kjrotation{C_{k+1}C_{k}}\\
				\mathbf{0} & \kjrotation{C_{k+1}C_{k}}
			\end{bmatrix},
		\end{aligned}
	\end{equation*}
	where
	\begin{equation*}
		\kjseel{C_{k+1}}{C_{k}} = \ln\left(\kjtransform{C_{k+1}C_{k}}{} \right)^{\vee},
	\end{equation*}
	and $\left(\cdot\right)^{\vee} \colon \mathbb{R}^{4\times4} \to \mathbb{R}^{6\times1}$ is the inverse of the \kjse{3} lift operator in \eqref{eq:estimation:ego:transform:wedge}.
	The inverse of the left Jacobian of \kjse{3}, $\boldsymbol{\mathcal{J}}^{-1}_{C_{k+1}C_{k}} \in \mathbb{R}^{6\times6}$, is defined as
	\begin{equation*}
		\begin{aligned}
			\boldsymbol{\mathcal{J}}^{-1}_{C_{k+1}C_{k}} &\coloneqq \left(\int_{0}^{1}\kjadjoint{C_{k+1}C_{k}}^{\alpha}d\alpha\right)^{-1}\\
			&\approx \mathbf{1} - \frac{1}{2}{\kjseel{C_{k+1}}{C_{k}}}^{\curlywedge}.
		\end{aligned}
	\end{equation*}

	The operating point is perturbed according to the transform perturbations, $\{\delta\kjseel{}{k} \in \mathbb{R}^{6}\}$, velocity perturbations $\{\delta\kjvelocity{k} \in \mathbb{R}^{6}\}$, and landmark perturbations, $\{\delta\boldsymbol{\zeta}_j \in \mathbb{R}^{3}\}$, which are stacked to form the full state perturbation, $\kjstatepert{}$. 
	The indicator matrices $\mathbf{P}_{jk}$ and $\mathbf{P}_k$ are defined such that $\kjstatepert{}_{jk} = \mathbf{P}_{jk}\kjstatepert{}$ is the perturbation of the  $\left\{\kjtransform{\mathrm{op},C_kC_1}{}, \kjvelocity{\mathrm{op},C_k}, \kjvecpoint{j_1C_1}{\mathrm{op},C_1}\right\}$ tuple and $\kjstatepert{}_{k} = \mathbf{P}_{k}\kjstatepert{}$ is the perturbation of the $\left\{\kjtransform{\mathrm{op},C_kC_1}{}, \kjvelocity{\mathrm{op},C_k}\right\}$ pair.
	
	The optimal perturbation, $\kjstatepert{}^*$, to minimize the linearized cost is the solution to the linear equation, 
	$\mathbf{A}\kjstatepert{}^* = \mathbf{b}$.
	Each element of the operating point is then updated using \eqref{eq:estimation:ego:transform:update} and
	\begin{equation}\label{eq:estimation:ego:wnoa:update}
		{\kjvelocity{C_k}} \leftarrow {\kjvelocity{C_k} + \delta\kjvelocity{k}^{*}}.
	\end{equation}
	
	The cost function is then relinearized about the updated operating point and the process iterates until a user-defined convergence criterion.

	\subsubsection{Pose-Velocity-Acceleration Estimator}\label{sec:estimation:ego:wnoj}
	The \ac{WNOA} prior incentivizes locally-constant velocity, i.e., zero acceleration, which yields more physically plausible velocities, but is less accurate for motions with nonconstant velocity.
	This limitation is particularly important for objects that change direction, and a prior that can model nonzero accelerations is more representative of real-world motions.
	
	The \ac{WNOJ} prior models the trajectory \emph{jerk} as a continuous zero-mean, white noise Gaussian process \citep{tang2019},
	\begin{equation*}
		\kjaccelerationdot{C}\left(t\right) \sim \mathcal{GP}\left(\mathbf{0},\mathbf{Q}_c\delta\left(t-t'\right)\right).
	\end{equation*}
	This is analogous to the pose-velocity estimator instead with a locally constant \emph{acceleration}.
	The prior can model nonzero accelerations and is more applicable to real-world motions with objects that change direction.

	The pose-velocity-acceleration estimator state, $\kjstate{}$, comprises the estimated pose transforms, $\kjtranset{C}$, the body-centric velocities, $\left(\kjvelocity{C_k}\right)_{k=1\dots K}$, the body-centric accelerations, $\left(\kjacceleration{C_k}\right)_{k=1\dots K}$. and the labeled landmark points, $\left\{\kjvecpoint{{j_1}{C_1}}{C_1}\right\}_{j=1\dots\vert\kjtrackletset{C}\vert}$.
	The state $\kjstate{}_{k} \coloneqq \left\{\kjtransform{C_{k}C_{1}}{}, \kjvelocity{C_k}, \kjacceleration{C_k}\right\}$ is defined as the transform, velocity, and acceleration at each time, $t_k$, and $\kjstate{}_{jk} \coloneqq \left\{\kjstate{}_{k}, \kjvecpoint{{j_1}{C_1}}{C_1}\right\}$ is defined for each pair of time steps and points in the label. 
	
	The estimator proceeds analogously to the pose-velocity estimator using the same combined objective function in \eqref{eq:estimation:ego:wnoa:combined_cost} with an expanded state transition function,
	\begin{equation}\label{eq:estimation:ego:wnoj:transition}
		\mathbf{\Phi}\left(\tau,t_{k}\right) \coloneqq \begin{bmatrix}
			\mathbf{1} & \left(\Delta t_{\tau,k}\right)\mathbf{1} & \frac{1}{2}\left(\Delta t_{\tau,k}\right)^{2}\mathbf{1} \\
			\mathbf{0} & \mathbf{1} & \left(\Delta t_{\tau,k}\right)\mathbf{1}\\
			\mathbf{0} & \mathbf{0} & \mathbf{1} 
		\end{bmatrix}.
	\end{equation}
	The expanded block covariance matrix,
	\begin{equation}\label{eq:estimation:ego:wnoj:covariance}
		\mathbf{Q}_{k}\left(\tau\right) \coloneqq \begin{bmatrix}
			\frac{1}{20}\Delta t_{\tau,k}^{5}\mathbf{Q}_c & \frac{1}{8}\Delta t_{\tau,k}^{4}\mathbf{Q}_c  & \frac{1}{6}\Delta t_{\tau,k}^{3}\mathbf{Q}_c\\
			\frac{1}{8}\Delta t_{\tau,k}^{4}\mathbf{Q}_c & \frac{1}{3}\Delta t_{\tau,k}^{3}\mathbf{Q}_c  & \frac{1}{2}\Delta t_{\tau,k}^{2}\mathbf{Q}_c\\
			\frac{1}{6}\Delta t_{\tau,k}^{3}\mathbf{Q}_c & \frac{1}{2}\Delta t_{\tau,k}^{2}\mathbf{Q}_c & \Delta t_{\tau,k}\mathbf{Q}_c
		\end{bmatrix},
	\end{equation}
	has the inverse,
	\begin{equation*}
		\begin{aligned}
			&\mathbf{Q}_{k}^{-1}\left(\tau\right) \coloneqq\\ &\quad\;\begin{bmatrix}
				720\Delta t_{\tau,k}^{-5}\mathbf{Q}_c^{-1}  & -360\Delta t_{\tau,k}^{-4}\mathbf{Q}_c^{-1}  & 60\Delta t_{\tau,k}^{-3}\mathbf{Q}_c^{-1}\\
				-360\Delta t_{\tau,k}^{-4}\mathbf{Q}_c^{-1}  &  192\Delta t_{\tau,k}^{-3}\mathbf{Q}_c^{-1}  & -36\Delta t_{\tau,k}^{-2}\mathbf{Q}_c^{-1}\\
				60\Delta t_{\tau,k}^{-3}\mathbf{Q}_c^{-1}  &  -36\Delta t_{\tau,k}^{-2}\mathbf{Q}_c^{-1}  & 9\Delta t_{\tau,k}^{-1}\mathbf{Q}_c^{-1}
			\end{bmatrix}.
		\end{aligned}
	\end{equation*}

	The total cost is again minimized by linearizing the error about an operating point, $\kjstate_{\mathrm{op}}$.
	The linearized form is identical to \eqref{eq:estimation:ego:wnoa:linear_cost} where the Jacobian of measurement model, $\mathbf{G}_{jk}$, in \eqref{eq:estimation:ego:transform:jacobian} is calculated with the updated motion model Jacobian,
	\begin{equation*}
		\frac{\partial\mathbf{z}}{\partial\kjstatepert{}}\bigg\rvert_{\kjstate{}_{\mathrm{op},jk}} = \begin{bmatrix}
			\left({\kjtransform{\mathrm{op},{C_k}{C_1}}{}}\kjhopoint{{j_1}{C_1}}{\mathrm{op},C_1}\right)^{\odot} & \mathbf{0} & \mathbf{0} & {\kjtransform{\mathrm{op},{C_k}{C_1}}{}}\mathbf{D}
		\end{bmatrix},
	\end{equation*}
	and the Jacobian of the prior error function is given as
	\begin{equation*}
		\begin{aligned}
			\mathbf{E}_{k}=&
			\begin{bmatrix}
				\mathbf{E}_{11} & \Delta t_{k+1,k}\mathbf{1} & \frac{1}{2}\Delta t_{k+1,k}^{2}\mathbf{1} & \mathbf{E}_{13} & \mathbf{0}  & \mathbf{0}\\
				\mathbf{E}_{21} & \mathbf{1} & \Delta t_{k+1,k}\mathbf{1} & \mathbf{E}_{24} & \mathbf{E}_{25}  & \mathbf{0}\\
				\mathbf{E}_{31} & \mathbf{0} & \mathbf{1} & \mathbf{E}_{34} & \mathbf{E}_{35} &  \mathbf{E}_{36}
			\end{bmatrix},\\
			\mathbf{E}_{11}=&\boldsymbol{\mathcal{J}}^{-1}_{C_{k+1}C_{k}}{\boldsymbol{\mathcal{T}}}_{C_{k+1}C_{k}},\\
			\mathbf{E}_{13}=&-\boldsymbol{\mathcal{J}}^{-1}_{C_{k+1}C_{k}},\\
			\mathbf{E}_{21}=&\frac{1}{2}\kjvelocity{C_{k+1}}^{\curlywedge}\boldsymbol{\mathcal{J}}^{-1}_{C_{k+1}C_{k}}{\boldsymbol{\mathcal{T}}}_{C_{k+1}C_{k}},\\
			\mathbf{E}_{24}=&-\frac{1}{2}\kjvelocity{C_{k+1}}^{\curlywedge}\boldsymbol{\mathcal{J}}^{-1}_{C_{k+1}C_{k}},\\
			\mathbf{E}_{25}=&-\boldsymbol{\mathcal{J}}^{-1}_{C_{k+1}C_{k}},\\
			\mathbf{E}_{31}=&\frac{1}{4}\kjvelocity{C_{k+1}}^{\curlywedge}\kjvelocity{C_{k+1}}^{\curlywedge}\boldsymbol{\mathcal{J}}^{-1}_{C_{k+1}C_{k}}{\boldsymbol{\mathcal{T}}}_{C_{k+1}C_{k}} \\
			&+ \frac{1}{2}\kjacceleration{C_{k+1}}^{\curlywedge}\boldsymbol{\mathcal{J}}^{-1}_{C_{k+1}C_{k}}{\boldsymbol{\mathcal{T}}}_{C_{k+1}C_{k}},\\
			\mathbf{E}_{34}=&\frac{1}{4}\kjvelocity{C_{k+1}}^{\curlywedge}\kjvelocity{C_{k+1}}^{\curlywedge}\boldsymbol{\mathcal{J}}^{-1}_{C_{k+1}C_{k}}+ \frac{1}{2}\kjacceleration{C_{k+1}}^{\curlywedge}\boldsymbol{\mathcal{J}}^{-1}_{C_{k+1}C_{k}},\\
			\mathbf{E}_{35}=&\frac{1}{2}\left(\boldsymbol{\mathcal{J}}^{-1}_{C_{k+1}C_{k}}\kjvelocity{C_{k+1}}^{\curlywedge}\right)^{\curlywedge} - \frac{1}{2}\kjvelocity{C_{k+1}}^{\curlywedge}\boldsymbol{\mathcal{J}}^{-1}_{C_{k+1}C_{k}},\\
			\mathbf{E}_{36}=&\boldsymbol{\mathcal{J}}^{-1}_{C_{k+1}C_{k}}.
		\end{aligned}
	\end{equation*}
	
	The operating point is perturbed according to the transform perturbations, $\{\delta\kjseel{}{}{k} \in \mathbb{R}^{6}\}$, velocity perturbations $\{\delta\kjvelocity{k} \in \mathbb{R}^{6}\}$, acceleration perturbations $\{\delta\kjacceleration{k} \in \mathbb{R}^{6}\}$, and landmark perturbations, $\{\delta\boldsymbol{\zeta}_j \in \mathbb{R}^{3}\}$, which are stacked to form the full state perturbation, $\kjstatepert{}$. 
	The indicator matrices $\mathbf{P}_{jk}$ and $\mathbf{P}_k$ are defined such that $\kjstatepert{}_{jk} = \mathbf{P}_{jk}\kjstatepert{}$ is the perturbation of the  $\left\{\kjtransform{\mathrm{op},C_kC_1}{}, \kjvelocity{\mathrm{op},C_k}, \kjacceleration{C_k}, \kjvecpoint{j_1C_1}{\mathrm{op},C_1}\right\}$ tuple and $\kjstatepert{}_{k} = \mathbf{P}_{k}\kjstatepert{}$ is the perturbation of the $\left\{\kjtransform{\mathrm{op},C_kC_1}{}, \kjvelocity{\mathrm{op},C_k},\kjacceleration{C_k}\right\}$ tuple.
	
	The optimal perturbation, $\kjstatepert{}^*$, to minimize the linearized cost is the solution to the linear equation 
	$\mathbf{A}\kjstatepert{}^* = \mathbf{b}$. 
	The operating point is then updated using \eqref{eq:estimation:ego:transform:update}, \eqref{eq:estimation:ego:wnoa:update}, and
	\begin{equation*}
		\kjacceleration{C_k} \leftarrow \kjacceleration{C_k} + \delta\kjacceleration{k}^{*},
	\end{equation*}
	and the cost is relinearized about the updated operating point until until a user-defined convergence criterion.
	
	\subsection{Geocentric Third-Party Estimation}\label{sec:estimation:geo}
	Egomotion estimation techniques must be adapted to estimate the geocentric trajectories of dynamic objects that invalidate the static tracklet assumption.
	Initial versions of the ideas in \cref{sec:estimation:geo:transform,sec:estimation:geo:wnoa} were first published in \citet{judd2018iros} and \citet{judd2020iros}, respectively, and they are further developed here.
	The ideas in \Cref{sec:estimation:geo:wnoj} are presented here for the first time.

	Pose-only estimates (\cref{sec:estimation:ego:transform}) in an egocentric frame can be directly converted to a geocentric frame (\cref{sec:estimation:geo:transform}).
	Inertial priors, such as the \ac{WNOA} (\cref{sec:estimation:ego:wnoa}) and \ac{WNOJ} (\cref{sec:estimation:ego:wnoj}) priors, are more complicated because bodies moving with a constant motion relative to a common inertial reference frame do not have constant motion relative to \emph{each other}.
	Applying these motion priors to third-party trajectories requires rederiving the motion model in \eqref{eq:estimation:ego:measurement} and its associated Jacobians in \eqref{eq:estimation:ego:transform:jacobian} (\cref{sec:estimation:geo:wnoa,sec:estimation:geo:wnoj}).

	\subsubsection{Pose-Only Estimator}\label{sec:estimation:geo:transform}
	\kjse{3} transforms initially estimated using egomotion estimation techniques (\cref{sec:estimation:ego:transform}) represent egomotion hypotheses calculated from the assumption that observed points are static.
	When the tracklets are moving, these estimates are the inverse of the objects' motion relative to the sensor, i.e., the inverse of their egocentric motion.
	
	Egomotion hypothesis trajectories, ${^\kjlabel{}}\kjtranset{C}$, can be transformed into true third-party geocentric trajectories by compensating for the identified egomotion estimate, $\kjtransform{C_{k}C_{1}}{}$, via
	\begin{equation}\label{eq:estimation:geo:transform:geo}
		\forall\kjlabel{}\in\kjlabelset{}\setminus C,\enspace \kjtransform{\kjlabel{}_{k}\kjlabel{}_{1}}{} = \kjtransformrigid{{\kjlabel_k}{\kjlabel_1}}{} \kjtransform{\kjlabel{}_{1}C_{1}}{} {^\kjlabel{}\kjtransform{{C_k}{C_1}}{-1}} \kjtransform{C_{k}C_{1}}{} \kjtransform{\kjlabel{}_{1}C_{1}}{-1},
	\end{equation}
	where $\kjtransformrigid{{\kjlabel_k}{\kjlabel_1}}{}$ is the object deformation matrix and is assumed to be identity, (i.e., rigid-body motion). 
	
	The initial sensor-to-object transform, $\kjtransform{\kjlabel{}_{1}C_{1}}{}$, relates the object frame from the sensor frame and is calculated blockwise from the initial translation and rotation.
	The initial translation, $\kjvecpoint{C_{1}\kjlabel{}_{1}}{\kjlabel{}_{1}}$, of the object frame is assigned to the observed centroid of the labeled points, $\kjtrackletset{\kjlabel{}}$, projected into the first observed frame,
	\begin{equation}\label{eq:estimation:geo:transform:centroid}
		\begin{aligned}
			\bar{\boldsymbol{p}}^{\kjlabel{}_{1}C_{1}}_{C_{1}} &= \frac{1}{\vert\kjtrackletset{\kjlabel{}}\vert}\sum_{j = 1}^{\vert\kjtrackletset{\kjlabel{}}\vert}{ {^{\kjlabel{}}\kjtransform{{C_{t_j}}{C_1}}{-1}} \kjhopoint{j_{t_j}C_{t_j}}{C_{t_j}}},\\
			\kjvecpoint{C_{1}\kjlabel{}_{1}}{\kjlabel{}_{1}} &=-\kjrotation{\kjlabel_1C_1}\kjvecpointmean{\kjlabel{}_{1}C_{1}}{C_{1}},
		\end{aligned}
	\end{equation}
	where $t_j$ is the first frame where the tracklet is observed.
	The initial rotation, $\kjrotation{\kjlabel_{1} C_1}$, is the initial frame orientation, which may be calculated from semantic information or be user specified (e.g., identity).
	
	Note that the egomotion estimate is only needed to convert third-party motions from the egocentric to geocentric frames and all motions are batch estimated first as egomotion hypotheses (\cref{sec:estimation:ego:transform}) before identifying the egomotion label.
	This means motion-based heuristics and other application-specific information can be used to identify the egomotion before converting the remaining third-party motions to geocentric estimates.
	
	\subsubsection{Pose-Velocity Estimator}\label{sec:estimation:geo:wnoa}
	The \ac{WNOA} prior is defined in an inertial frame, so third-party motions are estimated after identifying the egomotion.
	The techniques used to estimate the egomotion (\cref{sec:estimation:ego:wnoa}) must be adapted to estimate third-party motions in a geocentric frame.
	
	The transform model, $\mathbf{z}$, in \eqref{eq:estimation:ego:measurement} is adjusted to transform egocentrically observed points through a geocentrically estimated state,
	\begin{equation}\label{eq:estimation:geo:wnoa:motion}
		\mathbf{z}'\left(\mathbf{x}_{jk}\right) \coloneqq\kjtransform{{C}_{k}{C}_{1}}{} \kjtransform{\kjlabel{}_{1}{C}_{1}}{-1} \kjtransform{\kjlabel{}_{k}\kjlabel{}_{1}}{-1} \kjtransformrigid{\kjlabel{}_{k}\kjlabel{}_{1}}{} \kjtransform{\kjlabel{}_{1}C_{1}}{} \kjhopoint{{j_{1}}{C_{1}}}{C_{1}},
	\end{equation}
	where $\kjtransformrigid{\kjlabel{}_{k}\kjlabel{}_{1}}{}$ is the object deformation matrix (identity for rigid bodies), and $\kjtransform{{C}_{k}{C}_{1}}{}$ is the identified egomotion (\cref{sec:estimation:ego}).
	The motion model component of the measurement Jacobian, $\mathbf{G}_{jk}$, in \eqref{eq:estimation:ego:transform:jacobian} is given by the block row vector
	\looseness=-1
	\begin{equation}\label{eq:estimation:geo:wnoa:motion_jacobian}
		\begin{aligned}
			\frac{\partial\mathbf{z}'}{\partial\kjstatepert{}}\bigg\rvert_{\kjstate{}_{\mathrm{op},jk}}\hspace{-1mm} &=\begin{bmatrix} \mathbf{Z}_{1} & \mathbf{0}  & \mathbf{Z}_{3} \end{bmatrix},\\
			\mathbf{Z}_{1} &= -\kjtransform{C_{k}C_{1}}{} \kjtransform{\kjlabel{}_{1}C_{1}}{-1} \kjtransform{\mathrm{op},\kjlabel{}_{k}\kjlabel{}_{1}}{-1} \left(\kjtransformrigid{\kjlabel{}_{k}\kjlabel{}_{1}}{} \kjtransform{\kjlabel{}_{1}C_{1}}{} \kjhopoint{{j_{1}}{C_{1}}}{\mathrm{op},C_{1}}\right)^{\odot}\hspace{-1mm},\\
			\mathbf{Z}_{3} &= \kjtransform{C_{k}C_{1}}{} \kjtransform{\kjlabel{}_{1}C_{1}}{-1} \kjtransform{\mathrm{op},\kjlabel{}_{k}\kjlabel{}_{1}}{-1} \kjtransformrigid{{\kjlabel{}_{k}}\kjlabel{}_{1}}{} \kjtransform{\kjlabel{}_{1}C_{1}}{} \mathbf{D}.
		\end{aligned}
	\end{equation}
	
	The prior cost is the same as in \eqref{eq:estimation:ego:wnoa:jacobian}.
	The linearized cost in  \eqref{eq:estimation:ego:wnoa:linear_cost} is then used to estimate the geocentric trajectory of every third-party motion in the scene.
	
	Note that the egomotion must be estimated before the third-party motions, so motion-based heuristics cannot be used to help identify the egomotion label as in pose-only estimation.
	
	\subsubsection{Pose-Velocity-Acceleration Estimator}\label{sec:estimation:geo:wnoj}
	The \ac{WNOJ} prior is also defined in an inertial frame and is adapted using a similar procedure to the pose-velocity estimator (\cref{sec:estimation:geo:wnoa}).
	
	The transform model, $\mathbf{z}$, in \eqref{eq:estimation:ego:measurement} takes the same form as \eqref{eq:estimation:geo:wnoa:motion}.
	The motion model component of the measurement Jacobian, $\mathbf{G}_{jk}$, in \eqref{eq:estimation:ego:transform:jacobian} is given by the block row vector
	\begin{equation*}
		\begin{aligned}
			\frac{\partial\mathbf{z}'}{\partial\kjstatepert{}}\bigg\rvert_{\kjstate{}_{\mathrm{op},jk}} &=\begin{bmatrix} \mathbf{Z}_{1} & \mathbf{0}  & \mathbf{0}  & \mathbf{Z}_{3} \end{bmatrix},
		\end{aligned}
	\end{equation*}
	where $\mathbf{Z}_{1}$ and $\mathbf{Z}_{3}$ are defined in \eqref{eq:estimation:geo:wnoa:motion_jacobian}.
	The linearized cost takes the same form as \eqref{eq:estimation:ego:wnoa:linear_cost} and uses the same prior cost term as \eqref{eq:estimation:ego:wnoa:jacobian}.
	
	Note again that the egomotion must be estimated before the third-party motions and therefore motion-based heuristics cannot be used to identify the egomotion label.
	
	\section{Sliding-Window Multimotion Estimation}\label{sec:sliding}
	\cref{sec:mvo,sec:estimation} present a full-batch implementation of \ac{MVO}.
	Full-batch estimation requires all observations of a given data segment at once but makes motion segmentation and estimation more accurate because motions are more distinct over longer periods of time.
	It can also reason about temporary occlusions after they occur, making it easier to determine when an object becomes occluded or unoccluded, as well as track motions through those occlusions.
	Full-batch estimation is computationally expensive and is often used in offline estimation systems.
	
	\ac{MVO} can also be implemented as an online sliding-window pipeline that processes data as it is observed and is less computationally expensive than a full-batch implementation.
	Sliding-window estimation uses smaller batches of the $K$-most-recent observation frames, and maintaining consistent trajectories over consecutive batches presents several challenges.
	
	Motion segmentation is more difficult in a sliding-window pipeline because independent motions are often less distinct over short timescales.
	Motions must be associated across estimation windows to maintain label consistency because motions in separate windows are not guaranteed to be assigned the same label (\cref{sec:sliding:consistency}).
	Unlike full-batch estimation, sliding-window estimation must reason about temporary occlusions as they occur.
	This requires motions to be extrapolated from, and interpolated between, direct observations (\cref{sec:sliding:extrapolation}).
	Detecting when an object becomes unoccluded requires determining when a newly segmented motion is actually the reappearance of a previously occluded motion, i.e., \emph{motion closure} (\cref{sec:sliding:motion_closure}).
	Initial versions of these ideas were first presented in \citet{judd2019lhmp} and \citet{judd2020iros}, and they are further developed and fully detailed for the first time here.
	
	\subsection{Label Consistency}\label{sec:sliding:consistency}
	Sliding-window estimation requires consistently tracking motions across multiple windows.
	This involves associating motions in consecutive estimation windows to form a single trajectory over the entirety of the data.
	
	Each new window associates currently estimated motion labels, $\kjlabel{}_{k}$, with previously estimated motion labels, $\kjlabel{}_{k-1}'$, based on the overlap in their labeled tracklets,
	\begin{equation} 
		\kjlabel{}_k \coloneqq \argmax_{\kjlabel{}_{k-1}'} \vert\kjtrackletset{\kjlabel{}_{k}} \cup \kjtrackletset{\kjlabel{}_{k-1}'} \vert.
	\end{equation}
	Newly estimated motions without sufficient overlap with a previous motion are tracked as new object motions.
	Other motion- or appearance-based heuristics could also be employed to make this association more robust.
	
	The labeling and estimates from the current window are used to initialize the segmentation and egomotion hypotheses in the next window.
	Any current tracklets that also exist in the next window retain their labels, and the current trajectory estimates are extrapolated forward to the next time stamp to initialize the label set.
	This is done by extrapolating the geocentric estimates, $\kjtransform{\kjlabel{}_{k}\kjlabel{}_{1}}{}$, forward and then converting them into egomotion hypotheses by rearranging \eqref{eq:estimation:geo:transform:geo},
	\begin{equation}
		{^\kjlabel{}\kjtransform{{C_k}{C_1}}{}} = \kjtransform{C_{k}C_{1}}{} \kjtransform{\kjlabel{}_{1}C_{1}}{-1} \kjtransform{\kjlabel{}_{k}\kjlabel{}_{1}}{-1} \kjtransformrigid{{\kjlabel_k}{\kjlabel_1}}{}  \kjtransform{\kjlabel{}_{1}C_{1}}{}.
	\end{equation}
	Consistently observed motion labels are straightforward to propagate (\cref{sec:sliding:extrapolation}), but this is more difficult in the presence of occlusion (\cref{sec:sliding:motion_closure}).
	
	\subsection{Extrapolation and Interpolation}\label{sec:sliding:extrapolation}
	Unobserved motions can be estimated by modeling the expected motions of previous observations.
	This is less accurate than direct estimation, but can extend existing motions to initialize the next set of motion estimates as well as estimate motions through temporary occlusions.
	The accuracy of these estimates depends on the fidelity of the motion model to the true motion of the object, and inferring the motion of an object between two states, i.e., interpolation, is more accurate than from a single state, i.e., extrapolation.
	
	Pose extrapolation and interpolation require an estimate of the change in pose over time, i.e., the velocity.
	In the pose-only estimator (\cref{sec:estimation:ego:transform,sec:estimation:geo:transform}), the velocity is discretely calculated from the relative transform between the final two poses,
	\begin{equation*}
		\kjvelocity{\kjlabel{}_k} \coloneqq \frac{1}{\Delta t_{k,k-1}} \kjseel{\kjlabel{}_{k+1}}{\kjlabel{}_{k}}.
	\end{equation*}
	The body-centric velocity is continuously estimated in the pose-velocity (\cref{sec:estimation:ego:wnoa,sec:estimation:geo:wnoa}) and pose-velocity-acceleration estimators (\cref{sec:estimation:ego:wnoj,sec:estimation:geo:wnoj}).
	
	The pose-velocity-acceleration estimator also directly estimates the body-centric acceleration, which can be used for extrapolation and interpolation.
	This term is zero for the pose-only and pose-velocity estimators, i.e., constant-velocity extrapolation.

	\begin{figure*}[t]
		\centering
		\begin{subfigure}{.315\textwidth}
			\includegraphics[width=\textwidth]{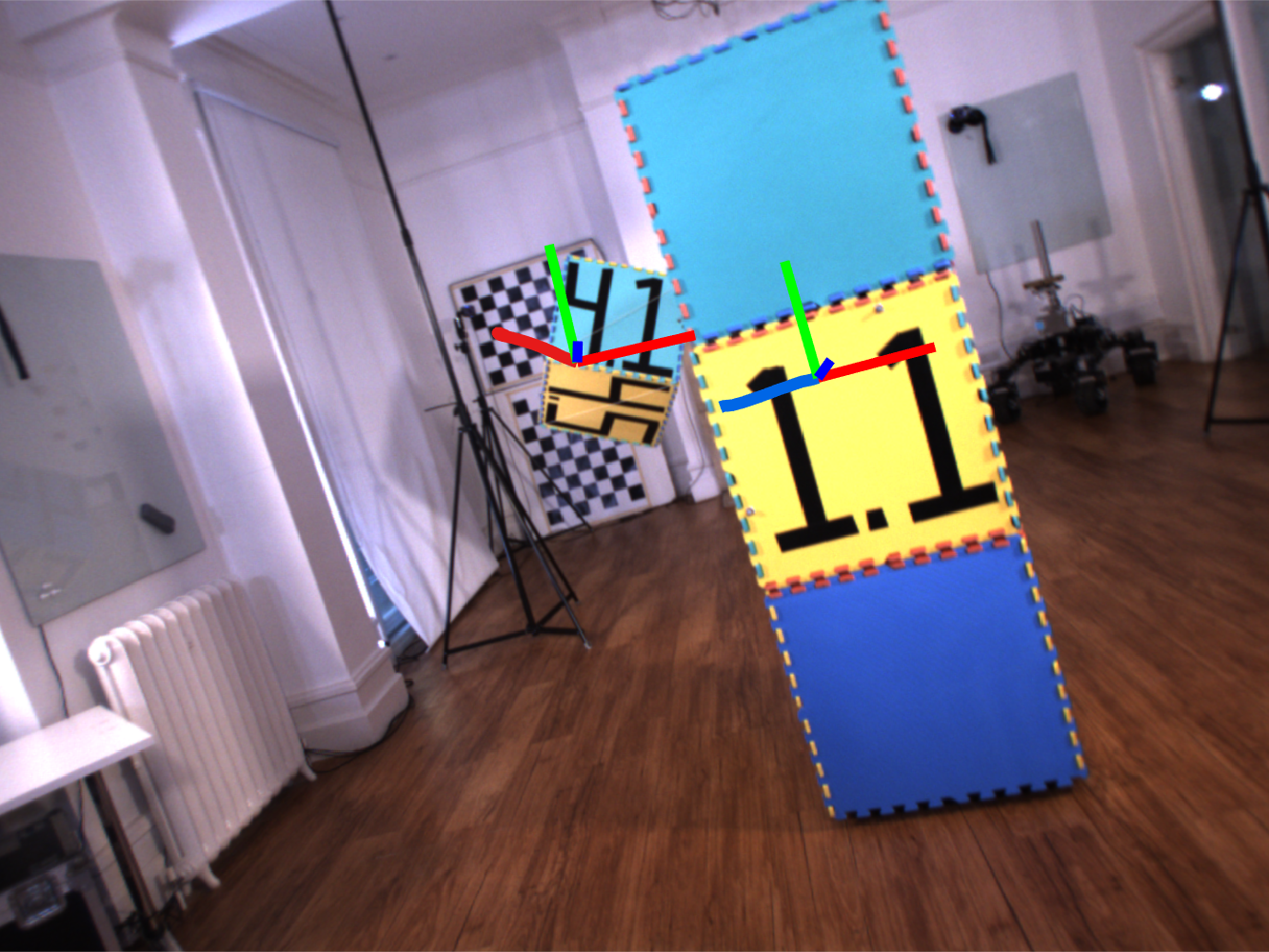}
		\end{subfigure}\hfill
		\begin{subfigure}{.315\textwidth}
			\includegraphics[width=\textwidth]{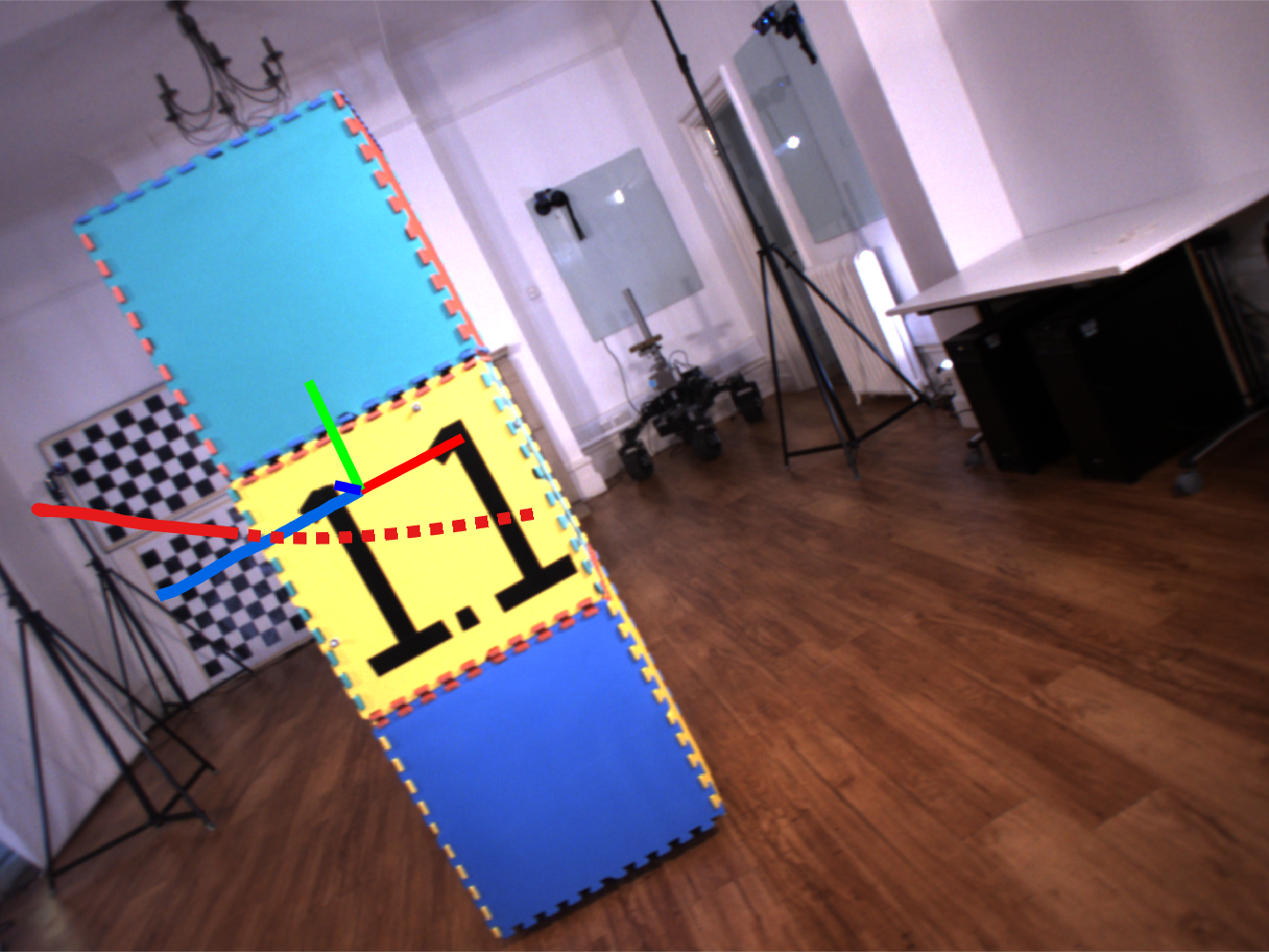}
		\end{subfigure}\hfill
		\begin{subfigure}{.315\textwidth}
			\includegraphics[width=\textwidth]{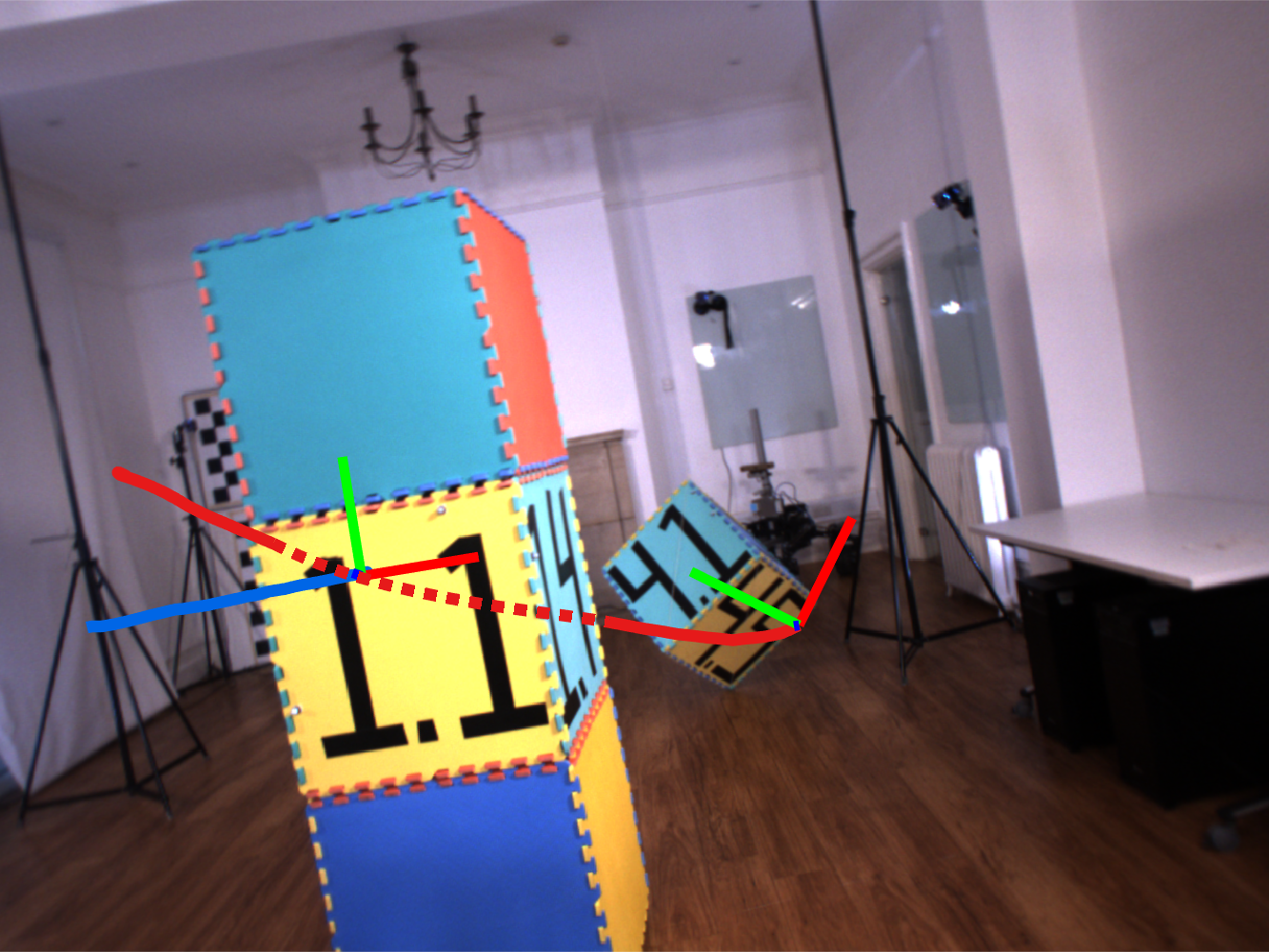}
		\end{subfigure}
		\caption{
			A demonstration of the \emph{motion closure} procedure showing trajectory estimates produced before (left), during (center), and after (right) an occlusion in the \texttt{occlusion\_2\_unconstrained} segment of the \acs{OMD}. 
			The trajectory of the swinging block (4, red) is directly estimated when it is visible and is extrapolated using the constant-velocity motion prior (dashed line) when the block is occluded by the moving tower (1, blue).
			When the block becomes unoccluded, it is rediscovered through motion closure and the estimates are interpolated to match the directly estimated trajectory.
		}
		\label{fig:sliding:marquee}
	\end{figure*}
	
	\subsubsection{Extrapolation}\label{sec:sliding:extrapolation:extrapolation}
	The local state, $\boldsymbol{\gamma}_{\kjlabel{}_k}$, of each estimator is extrapolated \citep{anderson2015,tang2019} via the transition function \eqref{eq:estimation:ego:wnoj:transition} to time $\tau$, 
	\begin{equation}\label{eq:sliding:wnoj_extrapolate}
		\check{\boldsymbol{\gamma}}_{\kjlabel{}_{k}}\left(\tau\right) = \mathbf{\Phi}\left(\tau,t_k\right){\boldsymbol{\gamma}}_{\kjlabel{}_{k}}\left(t_k\right) = \mathbf{\Phi}\left(\tau,t_k\right)\begin{bmatrix}
			\mathbf{0} \\ 
			\kjvelocity{\kjlabel{}_k} \\ 
			\kjacceleration{\kjlabel{}_k}
		\end{bmatrix},
	\end{equation} 
	where $\check{\cdot}$ denotes an extrapolated or interpolated value.
	Note that when the body-centric acceleration, $\kjacceleration{\kjlabel{}_k}$, is zero, as with the pose-only and the pose-velocity estimators, \eqref{eq:sliding:wnoj_extrapolate} is mathematically equivalent to \eqref{eq:estimation:ego:wnoa:transition} and \eqref{eq:estimation:ego:wnoa:transition_prior} with extra block zeros.
	\looseness=-1

	The extrapolated local state is transformed to the global pose, velocity, and acceleration state via
	\begin{equation} \label{eq:sliding:global}
		\begin{aligned}
			\kjtransform{\check{\kjlabel{}}}{}\left(\tau\right) &= \exp\left(\left(\begin{bmatrix}
				\mathbf{1} & \mathbf{0} & \mathbf{0} \end{bmatrix}\check{\boldsymbol{\gamma}}_{\kjlabel{}_{k}}\left(\tau\right)\right)^{\vee}\right)\kjtransform{\kjlabel{}}{}\left(t_k\right),\\
			\kjvelocity{\check{\kjlabel{}}}\left(\tau\right) &= \begin{bmatrix}
				\mathbf{0} & \mathbf{1} & \mathbf{0}\end{bmatrix}\check{\boldsymbol{\gamma}}_{\kjlabel{}_{k}}\left(\tau\right),\\
			\kjacceleration{\check{\kjlabel{}}}\left(\tau\right) &= \begin{bmatrix}
				\mathbf{0} & \mathbf{0} & \mathbf{1}\end{bmatrix}\check{\boldsymbol{\gamma}}_{\kjlabel{}_{k}}\left(\tau\right),
		\end{aligned}
	\end{equation}
	where the $\kjvelocity{\check{\kjlabel{}}}\left(\tau\right)$ term is omitted for the pose-only estimator and the $\kjacceleration{\check{\kjlabel{}}}\left(\tau\right)$ term is omitted for both the pose-only and pose-velocity estimators.
	
	Estimates can be extrapolated forward or backward in time.
	The extrapolation accuracy will degrade over time, especially if the true motion exhibits significant changes in velocity or acceleration.

	\subsubsection{Interpolation}\label{sec:sliding:extrapolation:interpolation}
	When the unobserved states are constrained by direct estimates on both sides, they can be estimated through interpolation.
	This is useful when tracking is resumed after an occlusion.
	The accuracy of the previously extrapolated estimates is improved because interpolation considers the directly observed data both before and after the occlusion.
	
	The pose-only state is linearly interpolated to an intermediate time, $t_j < \tau < t_k$, via
	\begin{equation*}
		\begin{aligned}
			\kjtransform{\check{\kjlabel{}}}{}\left(\tau\right) &= \exp\left(\kjseel{}{\tau j}^{\wedge}\right)\kjtransform{\kjlabel{}}{}\left(t_j\right)\\
			\kjseel{}{\tau j} &= \frac{\Delta t_{\tau j}}{\Delta t_{kj}}\kjseel{\kjlabel{}_k\kjlabel{}_j}{},
		\end{aligned}
	\end{equation*}
	which enforces a strict constant-velocity assumption.
	
	The pose-velocity and pose-velocity-acceleration states are interpolated \citep{anderson2015,tang2019} via  
	\begin{align*}
		\check{\boldsymbol{\gamma}}_{\kjlabel{}_j}\left(\tau\right) &= \boldsymbol{\Lambda}\left(\tau\right)\boldsymbol{\gamma}_{\kjlabel{}_j}\left(t_j\right) + \boldsymbol{\Omega}\left(\tau\right){\boldsymbol{\gamma}}_{\kjlabel{}_j}\left(t_{k}\right),\\
		\intertext{where}
		\boldsymbol{\gamma}_{\kjlabel{}_j}\left(t_k\right) &= \begin{bmatrix*}
			\kjseel{\kjlabel{}_k\kjlabel{}_j}{}\\
			\boldsymbol{\mathcal{J}}^{-1}_{\kjlabel{}_{k}\kjlabel{}_{j}}\kjvelocity{\kjlabel{}_{k}}\\
			-\frac{1}{2}\left(\boldsymbol{\mathcal{J}}^{-1}_{\kjlabel{}_{k}\kjlabel{}_{j}}\kjvelocity{\kjlabel{}_{k}}\right)^{\curlywedge}\kjvelocity{\kjlabel{}_{k}} + \boldsymbol{\mathcal{J}}^{-1}_{\kjlabel{}_{k}\kjlabel{}_{j}}\kjacceleration{\kjlabel{}_{k}}
		\end{bmatrix*},\\
		\boldsymbol{\Lambda}\left(\tau\right) &= \boldsymbol{\Phi}\left(\tau,t_j\right)-\boldsymbol{\Omega}\left(\tau\right)\boldsymbol{\Phi}\left(t_{k},t_j\right),\\
		\boldsymbol{\Omega}\left(\tau\right) &= \mathbf{Q}_{j}\left(\tau\right)\boldsymbol{\Phi}\left(t_{k},\tau\right)^{T}{\mathbf{Q}_{j}\left(t_k\right)^{-1}}.
	\end{align*}
	The block covariance matrix, $\mathbf{Q}_{j}\left(\tau\right)$, is given by \eqref{eq:estimation:ego:wnoa:covariance} for the \ac{WNOA} prior and \eqref{eq:estimation:ego:wnoj:covariance} for the \ac{WNOJ} prior.
	The interpolated state is then transformed to the global state via \eqref{eq:sliding:global}.
	
	Calculating the interpolation involves estimates both before and after the occlusion.
	This requires recognizing when a newly discovered motion can be explained by the reappearance of a previously occluded object, i.e., \emph{motion closure} (\cref{sec:sliding:motion_closure}).

	\subsection{Motion Closure} \label{sec:sliding:motion_closure}
	Online, sliding-window estimation requires actively recognizing when objects become occluded and unoccluded.
	While most existing tracking techniques address occlusions with appearance-based object detectors and simple motion models, \ac{MVO} tracks arbitrary objects through occlusions using motion-based tracking metrics (\cref{fig:sliding:marquee}).
	This \emph{motion closure} recognizes when a motion becomes unoccluded (\cref{sec:sliding:motion_closure:recognizing}), corrects the extrapolated pose using information from the new trajectory (\cref{sec:sliding:motion_closure:correcting}), and updates the previously extrapolated estimates through interpolation (\cref{sec:sliding:motion_closure:updating}).
	
	\subsubsection{Recognizing Previously Occluded Motions } \label{sec:sliding:motion_closure:recognizing}
	Motions are reacquired by comparing a newly discovered motion, ${\kjlabel{}'}$, to an extrapolated previously observed motion, ${\check{\kjlabel{}}}$, at a time, $t_k$.
	The two motions are considered the same and \emph{closed} if a weighted combination of the difference in their positions, $\kjvecpoint{\kjlabel{}_{k}'C_{k}}{C_{k}}$ and $\kjvecpoint{\check{\kjlabel{}}_{k}C_{k}}{C_{k}}$, and velocities, $\kjvelocity{\kjlabel{}'_{k}}$ and  $\kjvelocity{\check{\kjlabel{}}_{k}}$,
	is less than a user-defined threshold, $\epsilon_{\mathrm{mc}}$,
	\begin{equation*}
		\begin{aligned}
			\lambda_{\rm{mc}}\norm{\kjvecpoint{\check{\kjlabel{}}_{k}C_{k}}{C_{k}} - \kjvecpoint{\kjlabel{}_{k}'C_{k}}{C_{k}}} +\qquad\qquad\qquad&\\
			\left(1-\lambda_{\rm{mc}}\right)\norm{\kjvelocity{\check{\kjlabel{}}_{k}} - \kjadjoint{\check{\kjlabel{}}_{k}\kjlabel{}_{k}'}\kjvelocity{\kjlabel{}_{k}'}}& < \epsilon_{\mathrm{mc}}.
		\end{aligned}
	\end{equation*}
	The proportionality term, $\lambda_{\rm{mc}}$, governs how much the difference in position and the difference in velocity each contribute to whether the motions are closed.
	
	The position of the newly discovered motion, $\kjvecpoint{\kjlabel{}_{k}'C_{k}}{C_{k}}$, is calculated from the centroid of labeled points in the first frame the motion was observed as in \eqref{eq:estimation:geo:transform:centroid}.
	The position of the extrapolated motion, $\kjvecpoint{\check{\kjlabel{}}_{k}C_{k}}{C_{k}}$, can be found from the extrapolated transform, $\kjtransform{\check{\kjlabel{}}_kC_k}{}$, as in \eqref{eq:mvo:notation:transform} and \eqref{eq:mvo:notation:rotate}.
	
	The body-centric velocity of the newly discovered motion, $\kjvelocity{\kjlabel{}'_{k}}$, is  calculated directly by the estimator (\cref{sec:estimation}) and compared to the extrapolated velocity, $\kjvelocity{\check{\kjlabel{}}_{k}}$.
	This comparison is omitted for the pose-only estimator.

	This newly closed trajectory is then used to correct the extrapolated pose  (\cref{sec:sliding:motion_closure:correcting}).
	
	\subsubsection{Correcting Occluded States} \label{sec:sliding:motion_closure:correcting}
	The newly closed motion is adjusted to create a consistent, smooth trajectory.
	Information from both the extrapolated and newly discovered trajectories is used to correct the extrapolated states from the start of the occlusion to the time of closure, $t_k$.
	
	The final corrected pose, $\kjtransform{{\kjlabel{}}_{k}\kjlabel{}_{1}}{}$, is calculated by correcting the extrapolated pose, $\kjtransform{\check{\kjlabel{}}_{k}\kjlabel{}_{1}}{}$, to the same location as the newly discovered pose, $\kjtransform{\kjlabel{}_{k}'\kjlabel{}_{1}}{}$, using the correction transform, $\kjtransform{\kjlabel{}_{k}\check{\kjlabel{}}_{k}}{}$.
	This correction is calculated blockwise from the translation and rotation components as in \eqref{eq:mvo:notation:transform}.

	The translation component of the correction transform, $\kjvecpoint{\check{\kjlabel{}}_{k}\kjlabel{}_{k}}{\kjlabel{}_{k}}$, is the difference between the observed and extrapolated centroids, $\kjvecpoint{\kjlabel{}_{k}'C_{k}}{C_{k}}$ and $\kjvecpoint{\check{\kjlabel{}}_{k}C_{k}}{C_{k}}$, such that
	\begin{equation*}
		\begin{aligned}
			\kjvecpoint{\check{\kjlabel{}}_{k}\kjlabel{}_{k}}{\kjlabel{}_{k}} &=\mathbf{C}_{\kjlabel_{k}C_{k}}\kjvecpoint{\check{\kjlabel{}}_{k}\kjlabel{}_{k}}{C_{k}} = \mathbf{C}_{\kjlabel_{k}C_{k}}\left(\kjvecpoint{\check{\kjlabel{}}_{k}\kjlabel{}_{k}'}{C_{k}} + \kjvecpoint{\kjlabel{}_{k}'\kjlabel{}_{k}}{C_{k}}\right)\\ 
			\nonumber&=\mathbf{C}_{\kjlabel_{k}\check{\kjlabel}_{k}}\mathbf{C}_{\check{\kjlabel{}}_{k}C_k}\left(\kjvecpoint{\check{\kjlabel{}}_{k}C_{k}}{C_{k}} - \kjvecpoint{\kjlabel{}_{k}'C_{k}}{C_{k}} + \kjvecpoint{\kjlabel{}_{k}'\kjlabel{}_{k}}{C_{k}}\right),
		\end{aligned}
	\end{equation*}
	where $\kjvecpoint{\kjlabel{}_{k}'\kjlabel{}_{k}}{C_{k}} = \mathbf{0}$ because the newly observed centroid is taken as the position of the final corrected pose.
	
	The extrapolated camera-to-object rotation, $\mathbf{C}_{\check{\kjlabel{}}_{k}C_k}$, comes blockwise from 
	\begin{equation*}
		\begin{aligned}
			\kjtransform{\check{\kjlabel{}}_{k}C_{k}}{} = {\kjtransform{\check{\kjlabel{}}_{k}{\kjlabel{}}_{1}}{}} {\kjtransform{{\kjlabel{}}_{1}C_{1}}{}}
			\kjtransform{C_{k}C_{1}}{-1}
		\end{aligned}
	\end{equation*}
	using \eqref{eq:mvo:notation:transform}, 
	where $\kjframe{\kjlabel{}_1}$ is the object frame at the start of estimation, which is not the same as the start of the current sliding window.
	
	The rotation component of the correction transform, $\mathbf{C}_{\kjlabel{}_{k}\check{\kjlabel{}}_{k}}$, is not estimated and therefore taken as identity, which makes the rotation of the final corrected pose, $\mathbf{C}_{{\kjlabel{}}_{k}{\kjlabel{}}_{1}}$, equivalent to the rotation of extrapolated pose, $\mathbf{C}_{\check{\kjlabel{}}_{k}{\kjlabel{}}_{1}}$.
	Appearance-based methods could instead be used to estimate absolute object orientation after occlusion.
	\looseness=-1

	\subsubsection{Updating Closed Trajectories} \label{sec:sliding:motion_closure:updating}
	The previously estimated trajectory is updated to incorporate the motion closure.
	The correction transform, $\kjtransform{\kjlabel{}_{k}\check{\kjlabel{}}_{k}}{}$, is applied to the previously extrapolated state at the time of the closure, $\kjtransform{\check{\kjlabel{}}_{k}\kjlabel{}_{1}}{}$, to give the final corrected pose,
	\begin{equation*}
		\kjtransform{\kjlabel{}_{k}\kjlabel{}_{1}}{} = \kjtransform{\kjlabel{}_{k}\check{\kjlabel{}}_{k}}{}
		\kjtransform{\check{\kjlabel{}}_{k}\kjlabel{}_{1}}{}.
	\end{equation*} 
	The occluded states between the start of the occlusion and the corrected closure state, $\kjtransform{\kjlabel{}_{k}\kjlabel{}_{1}}{}$, are then updated via interpolation (\cref{sec:sliding:extrapolation:interpolation}).
	
	The states after the closure, $\tau > t_{k}$, are reestimated directly from measurements relative to the corrected pose, $\kjtransform{\kjlabel{}_{k}\kjlabel_{1}}{}$.
	The reestimate follows \cref{sec:estimation:geo} with an estimation window starting at $t_k$, and the associated initial sensor-to-object transform,
	\begin{equation*} 
		\kjtransform{\kjlabel{}_{k}C_{k}}{} = \kjtransform{\kjlabel{}_{k}{\kjlabel{}}_{1}}{}
		\kjtransform{{\kjlabel{}_{1}}C_{1}}{}
		\kjtransform{C_{k}C_{1}}{-1},
	\end{equation*} 
	is used as the initial sensor-to-object transform (i.e., $\kjtransform{\kjlabel{}_{1}C_{1}}{}$ in \eqref{eq:estimation:geo:transform:geo} and \eqref{eq:estimation:geo:wnoa:motion}) and the above $\kjtransform{\kjlabel_{1}C_1}{}$ is the original sensor-to-object transform at the start of estimation, found using \eqref{eq:estimation:geo:transform:centroid}.
	This uses both the past knowledge and current observations to estimate a smooth motion throughout the occlusion and maintain motion consistency for the entire trajectory.
	
	\begin{table}[t]
		\caption{\acs{MVO} Parameters used in the \acs{OMD} and KITTI experiments.}
		\label{tab:evaluation:parameters}
		\centering
		\begin{tabular}{p{0.56\columnwidth}|cc}
			\toprule
			Parameter & \acs{OMD} & KITTI\\
			\midrule
			Window length $\left(K\right)$ & 8  & 8 \\
			Graph neighbors $\left(k\right)$ & 4 & 4\\
			RANSAC threshold $\left(\epsilon_{\rm{th}}\right)$ & 4 & 6 \\
			RANSAC iterations ($N_{\rm{RSAC}}$) & 100 & 100\\
			Outlier cost $\left(\alpha\right)$ & 100 & 100\\
			Outlier cost $\left(\beta\right)$ & 5 & 5\\
			Assignment proportionality $\left(\lambda_{\mathrm{sm}}\right)$ & 0.5 & 0.5\\
			Per-label cost $\left(\mu_{\kjlabel{}}\right)$ & 1000 & 1000\\
			Min support size $\left(N_{\text{th}}\right)$ & 20 & 20\\
			Min trajectory length $\left(K_{\text{th}}\right)$ & 3 & 3\\
			Motion closure threshold $\left(\epsilon_\text{mc}\right)$ & 3 & 6\\
			Motion closure prop. $\left(\lambda_{\rm{mc}}\right)$ & 0.25 & 0.25\\
			Convergence iterations $\left(\kjvarconvergence{}\right)$ & 3 & 3\\
			\bottomrule
		\end{tabular}
	\end{table}
	
	\section{Evaluation}\label{sec:evaluation}
	The performance of \ac{MVO} is evaluated on real-world stereo data from the \ac{OMD} \citep{judd2019ral} and the KITTI dataset \citep{geiger2012}.
	It is evaluated both quantitatively and qualitatively and the accuracy of the different estimators is discussed.
	\ac{MVO} is also compared to \ac{VDO-SLAM} \citep{zhang2020arxiv}, a dynamic \ac{SLAM} system that also estimates the \kjse{3} motion of all objects in the scene, which has reported better performance than CubeSLAM \citep{yang2019} and ClusterVO \citep{huang2020}.
	
	Before evaluation, each estimated trajectory must first be calibrated to the ground-truth trajectory frame (\cref{sec:evaluation:calibration}).
	The accuracy of \ac{MVO} and \ac{VDO-SLAM} are evaluated using two metrics that give different context to their performance (\cref{sec:evaluation:metrics}).
	
	The \ac{OMD} contains several challenging multimotion scenes with ground-truth trajectory data for each motion in the scene.
	The \kjdatasegment{swinging\_4\_unconstrained} segment of the \ac{OMD} directly tests the ability to simultaneously segment and estimate multiple independent motions in a scene (\cref{sec:evaluation:swinging}).
	The \kjdatasegment{occlusion\_2\_unconstrained} segment exhibits significant occlusion and is used to evaluate the applicability of the motion priors for extrapolation and interpolation (\cref{sec:evaluation:occlusion}).

	The KITTI dataset is a driving dataset that includes ground-truth GPS/INS data for the sensor platform, but no third-party motion data.
	While it does not allow for the same quantitative evaluation as the \ac{OMD}, it is a commonly used benchmark for evaluating navigation algorithms and the segments \kjdatasegment{Drive 0005} and \kjdatasegment{Odometry 03} are used here to qualitatively evaluate performance in real-world driving scenarios (\cref{sec:evaluation:kitti}).
	
	The \ac{MVO} parameters used in the experiments are given in \cref{tab:evaluation:parameters}.
	Feature detection and matching were performed using LIBVISO2 \citep{geiger2011} and the Gauss-Newton minimization was performed with Ceres \citep{ceres-solver} using analytical derivatives (\cref{sec:estimation}). 
	Note that the significant differences in sensor and depth of field between the \ac{OMD} segments and the KITTI segments required slightly different values for two parameters.
	These parameter tunings would each be appropriate across their respective datasets.
	
	\subsection{Error Calibration}\label{sec:evaluation:calibration}
	Calculating the error between an estimate and the ground truth at a time, $t_k$, requires calibrating the trajectories at some earlier reference time, $t_j < t_k$.
	This makes the calibration dependent on the chosen reference time.
	
	The calibration depends on the initial transform between the estimated and ground truth object frames, 
	\begin{equation*}
		\kjtransform{\kjlabel{}_1^{\text{GT}}\kjlabel{}_1^{}}{} = \kjtransform{\kjlabel{}^{\text{GT}}_1A^{\text{GT}}_1}{} \kjtransform{A^{\text{GT}}C}{} \kjtransform{\kjlabel{}^{}_1C^{}_1}{-1}.
	\end{equation*}
	where $\kjtransform{A^{\text{GT}}C}{}$ is the transform relating the ground-truth sensor apparatus frame from the left camera frame. 
	The transform, $\kjtransform{A^{\text{GT}}C}{}$, is static and provided in each dataset.
	The estimated object frame is assigned by each estimator and is arbitrary relative to the ground-truth object frame.
	
	The calibration at any future time is given by propagating the initial calibration forward,
	\begin{equation}\label{eq:evluation:calibration:arbitrary}
		\kjtransform{\kjlabel{}_{j}^{\text{GT}}\kjlabel{}_{j}^{}}{} = \kjtransform{\kjlabel{}^{\text{GT}}_{j}\kjlabel{}^{\text{GT}}_{1}}{} \kjtransform{\kjlabel{}_1^{\text{GT}}\kjlabel{}_1^{}}{} \kjtransform{\kjlabel{}^{}_{j}\kjlabel{}^{}_{1}}{-1},
	\end{equation}
	where
	\begin{equation*}
		\kjtransform{\kjlabel{}_{j}^{}\kjlabel{}_{1}^{}}{} =\kjtransformrigid{\kjlabel{}^{}_{j}\kjlabel{}_1^{}}{} \kjtransform{\kjlabel{}^{\text{GT}}_1\kjlabel{}^{}_1}{-1}		\kjtransformrigid{\kjlabel{}_{j}^{\text{GT}}\kjlabel{}_1^{\text{GT}}}{-1} \kjtransform{\kjlabel{}^{\text{GT}}_{j}\kjlabel{}^{\text{GT}}_1}{} \kjtransform{\kjlabel{}^{\text{GT}}_{1}\kjlabel{}^{}_{1}}{},
	\end{equation*}
	is the ground-truth object motion expressed in the estimated frame, and $\kjtransformrigid{}{}$ is again the body deformation matrix and assumed to be identity.

	\subsection{Error Metrics}\label{sec:evaluation:metrics}
	The accuracy of the estimation techniques is assessed using two \kjse{3} error metrics.
	Global odometric error measures the estimation accuracy of the entire trajectory to a given point.
	Relative RMS error measures the error over consecutive time steps.
	Both errors provide useful context for quantifying the accuracy of the trajectory estimation.
	
	Each metric is defined using the error between the estimated motion, $\kjtransform{\kjlabel{}_k\kjlabel{}^{}_j}{}$, and the ground-truth motion, $\kjtransform{\kjlabel{}_k^{\text{GT}}\kjlabel{}_j^{\text{GT}}}{}$, at any time $t_k$ and calibrated via the pose at any time $t_j$, given as
	\begin{equation}\label{eq:evaluation:error}
		\mathrm{err}_{\kjlabel{}}\left(t_{j}, t_{k}\right) \coloneqq\kjtransformrigid{\kjlabel{}_k^{\text{GT}}\kjlabel{}_j^{\text{GT}}}{}\kjtransform{\kjlabel{}^{\text{GT}}_j\kjlabel{}^{}_j}{}		\kjtransformrigid{\kjlabel{}_k^{}\kjlabel{}_j^{}}{-1} \kjtransform{\kjlabel{}^{}_k\kjlabel{}^{}_j}{} \kjtransform{\kjlabel{}_j^{\text{GT}}\kjlabel{}^{}_j}{-1} \kjtransform{\kjlabel{}^{\text{GT}}_k\kjlabel{}^{\text{GT}}_j}{-1},
	\end{equation}
	where the calibration matrices are given in \eqref{eq:evluation:calibration:arbitrary}.
	
	\paragraph{Global Error}
	The global error measures the difference between an estimated pose relative to the initial pose and the ground truth.
	It evaluates the overall estimation accuracy along a trajectory and is calculated at a given time using \eqref{eq:evaluation:error},
	\begin{equation*}
		\text{GE}\left(\kjlabel{}, t_{k}\right)\coloneqq \ln\left(\mathrm{err}_{\kjlabel{}}\left(t_{k}, t_{1}\right)\right)^{\vee}.
	\end{equation*}
	The maximum global errors for each estimator are given in \cref{tab:evaluation:global:swinging,tab:evaluation:global:occlusion,tab:evaluation:global:kitti} and illustrated in  \cref{fig:results:swinging:discrete,fig:results:swinging:wnoa,fig:results:swinging:wnoj,fig:results:occlusion:discrete,fig:results:occlusion:wnoa,fig:results:occlusion:wnoj,fig:results:kitti,fig:results:kitti_od}.
	
	The global error on a specific data segment is contextualized by the total ground-truth path length (\cref{tab:evaluation:global:total_path_swinging,tab:evaluation:global:total_path_occlusion,tab:evaluation:global:total_path_kitti}) and maximum trajectory displacement (\cref{tab:evaluation:global:swinging,tab:evaluation:global:occlusion,tab:evaluation:global:kitti}).
	The total path contextualizes how far a motion travels in all directions, but does not completely describe the trajectory or the difficulty in estimating it.
	A repetitive motion, such as swinging blocks, unwinds accumulated error and makes the total path length less relevant for evaluating the estimator.
	The maximum trajectory displacement contextualizes how far a motion travels in any given direction.
	Unlike the total path length, this metric does not differentiate between a single traversal of a given path and multiple traversals.
	Together, both values are useful in understanding the global error of a given estimator on a particular data segment.

	\paragraph{Relative Error}
	The global error is sensitive to the time at which an error occurs.
	An estimation error that occurs at the beginning of a trajectory results in significantly more global error than an equivalent error occurring at the end of the trajectory.
	In contrast, the relative error measures how much the estimated transform between consecutive time stamps deviates from the ground truth.
	The translational and rotational \acl{RMSRE}\acused{RMSRE}s are given by
	\begin{equation*}
		\begin{aligned}
			\text{RMSRE}_{\text{xyz}}\left(\kjlabel{}\right) &\coloneqq \sqrt{\frac{1}{K}\sum^{K}_{k=2}\norm{\text{xyz}\left(\kjsetran{\kjlabel{},k}{}\right)}},\\
			\text{RMSRE}_{\text{angle}}\left(\kjlabel{}\right) &\coloneqq \sqrt{\frac{1}{K}\sum^{K}_{k=2}\norm{\kjsoel{\kjlabel{},k}}},
		\end{aligned}
	\end{equation*}
	where $\text{xyz}\left(\cdot\right)$ converts the \kjsealg{3} translational element to the equivalent Euclidean translation, and
	\begin{equation*}
		\kjseel{\kjlabel{},k}{} \coloneqq \begin{bmatrix}
			\kjsetran{\kjlabel{},k}{}\\
			\kjsoel{\kjlabel{},k} 
		\end{bmatrix} \coloneqq \ln\left(\mathrm{err}_{\kjlabel{}}\left(t_{k}, t_{k-1}\right)\right)^{\vee}.
	\end{equation*}
	This provides a single measure of the average relative error independent of when any individual error occurs.
	
	The \ac{RMSRE} is compared to the \ac{RMS} ground-truth frame-to-frame motion in \cref{tab:evaluation:relative:swinging,tab:evaluation:relative:occlusion,tab:evaluation:relative:kitti}.
	The ground-truth relative motion illustrates how much each object moves between each time stamp, which can become arbitrarily small for high-frame-rate sensors, even for quickly moving objects.
	Small ground-truth relative motions can result in estimators having arbitrarily small relative error.
	
	Both global and relative errors are decomposed into their translational (i.e., xyz) and rotational (i.e., roll-pitch-yaw) components.
	All errors are reported for geocentric trajectory estimates, so a portion of the error for each motion is due to error in the camera motion estimate.

	\ac{VDO-SLAM} third-party motion estimates were provided as calibrated relative motions.
	These transforms are recalibrated and combined to match the global transform representation of \ac{MVO} for evaluation.
	
	\begin{figure}[t]
		\begin{subfigure}{\columnwidth}
			\includegraphics[width=\textwidth]{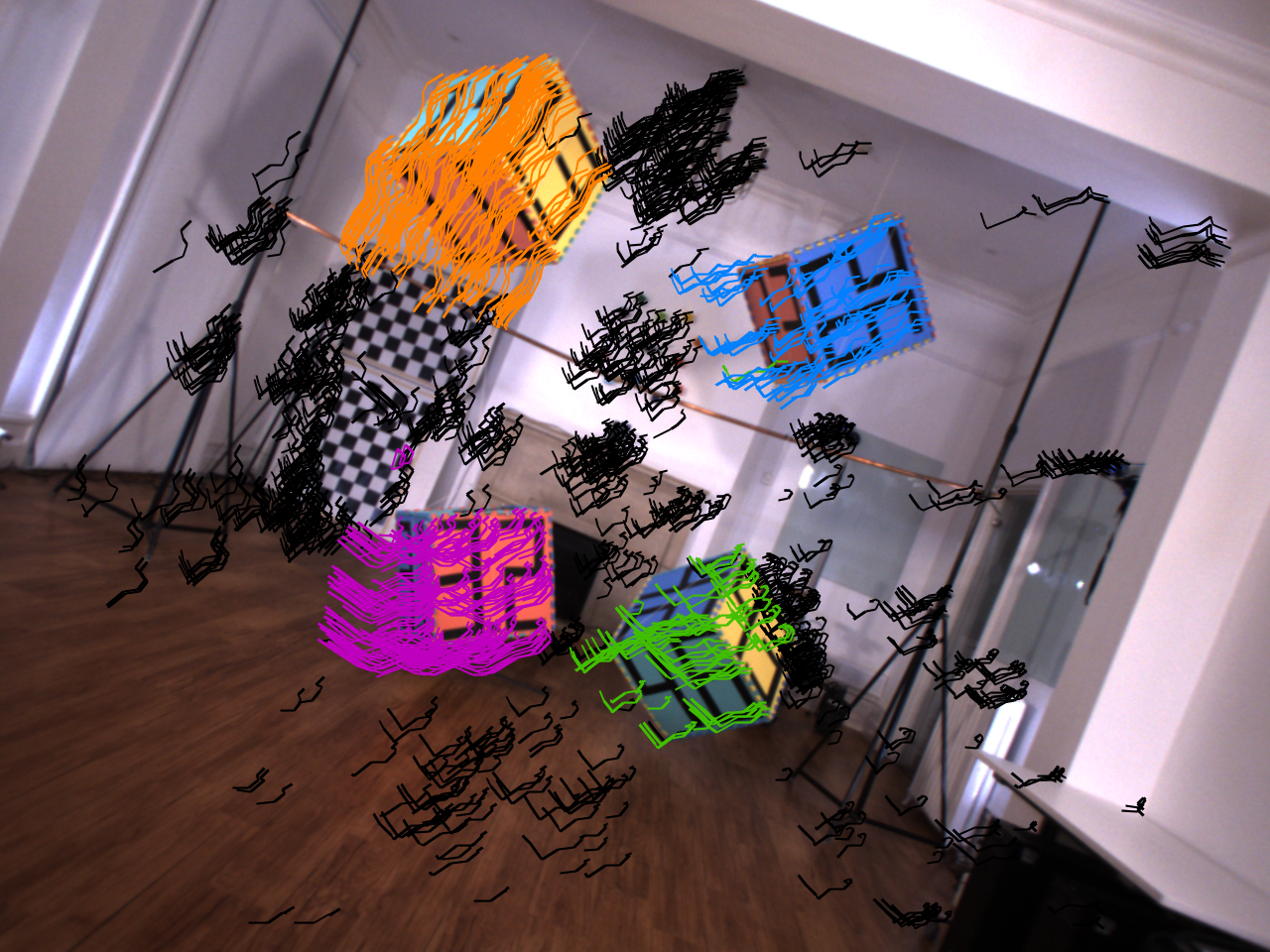}
		\end{subfigure}\\
		\begin{subfigure}{\columnwidth}
			\includegraphics[width=\textwidth]{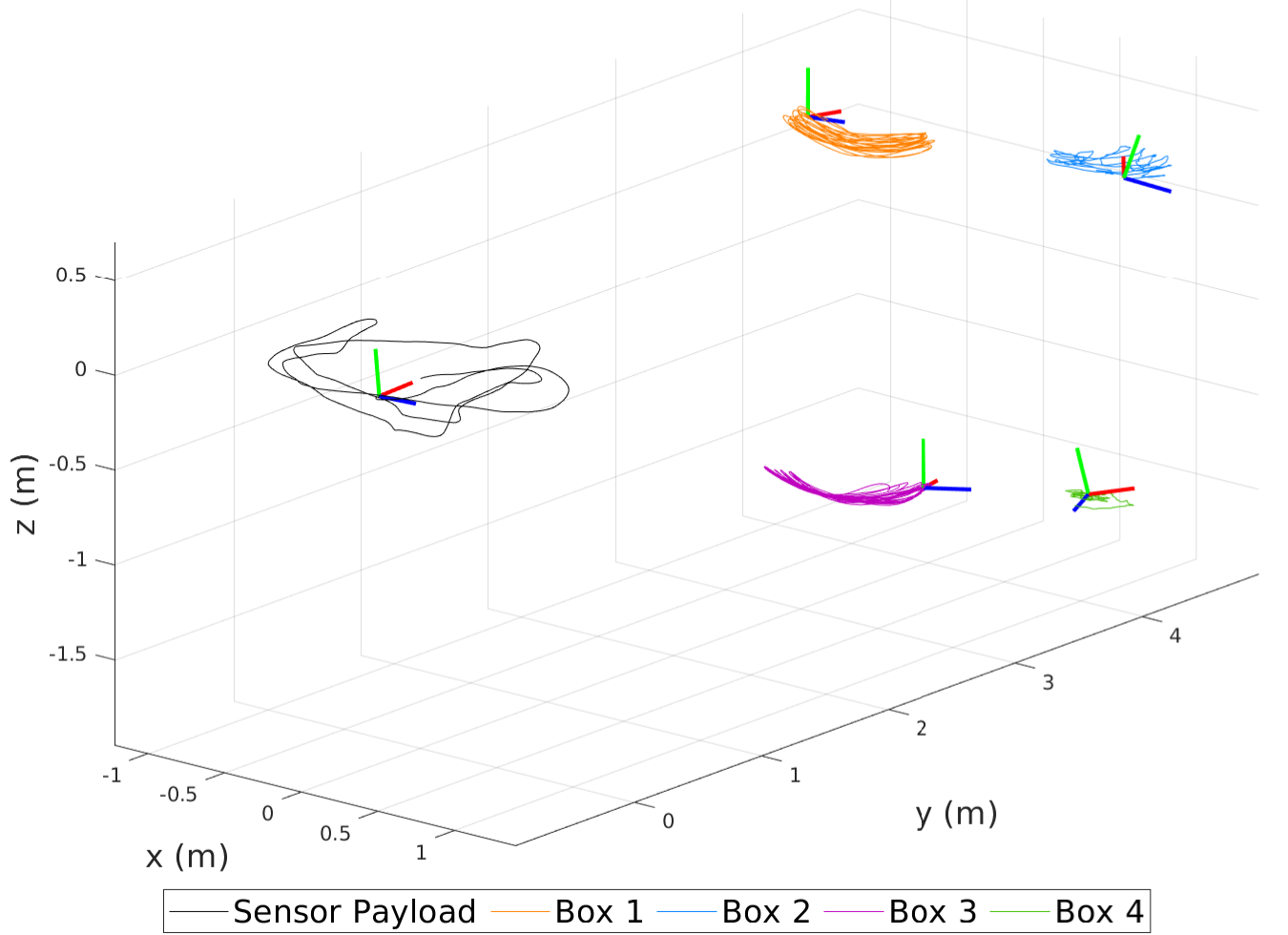}
		\end{subfigure}
		\caption{Motion segmentation (top) and trajectories (bottom) produced by \acs{MVO} using the pose-velocity estimator for the \kjdatasegment{swinging\_4\_unconstrained} data segment from the \acs{OMD}. 
			The egomotion (black, bottom) of the camera is estimated from the static points in the scene (black, top). 
			The motions of the swinging blocks (1--4) are segmented and estimated simultaneously with the egomotion.
			The results for each \acs{MVO} estimator for this segment are illustrated in Extensions 1--3 (\cref{app:extensions}).
		}
		\label{fig:marquee:swinging}
	\end{figure}
	
	\subsection{Multimotion Estimation}\label{sec:evaluation:swinging}
	The \kjdatasegment{swinging\_4\_unconstrained} segment of the \ac{OMD} presents complex \kjse{3} motions without significant occlusion.
	It tests the ability of estimators to simultaneously segment and estimate multiple motions.
	It consists of four blocks that both translate and rotate in pendular motions (\cref{fig:marquee:swinging}). 
	The blocks are observed by a dynamic camera, resulting in five independent \kjse{3} motions.
	A 500-frame section starting at frame 10 was used in this evaluation.
	
	Two of the blocks, Block 1 (top left) and Block 3 (bottom left), have minimal rotation and swing primarily along the ground-truth world y- and x-axis, respectively.
	The other two blocks, Block 2 (top right) and Block 4 (bottom right), rotate significantly about the positive ground-truth body z-axis.
	Block 4 rotates counter-clockwise and Block 2 rotates clockwise while also swinging, which creates a complex motion.
	\looseness=-1

	\begin{figure}[t]
		\begin{subfigure}{\columnwidth}
			\includegraphics[width=\textwidth]{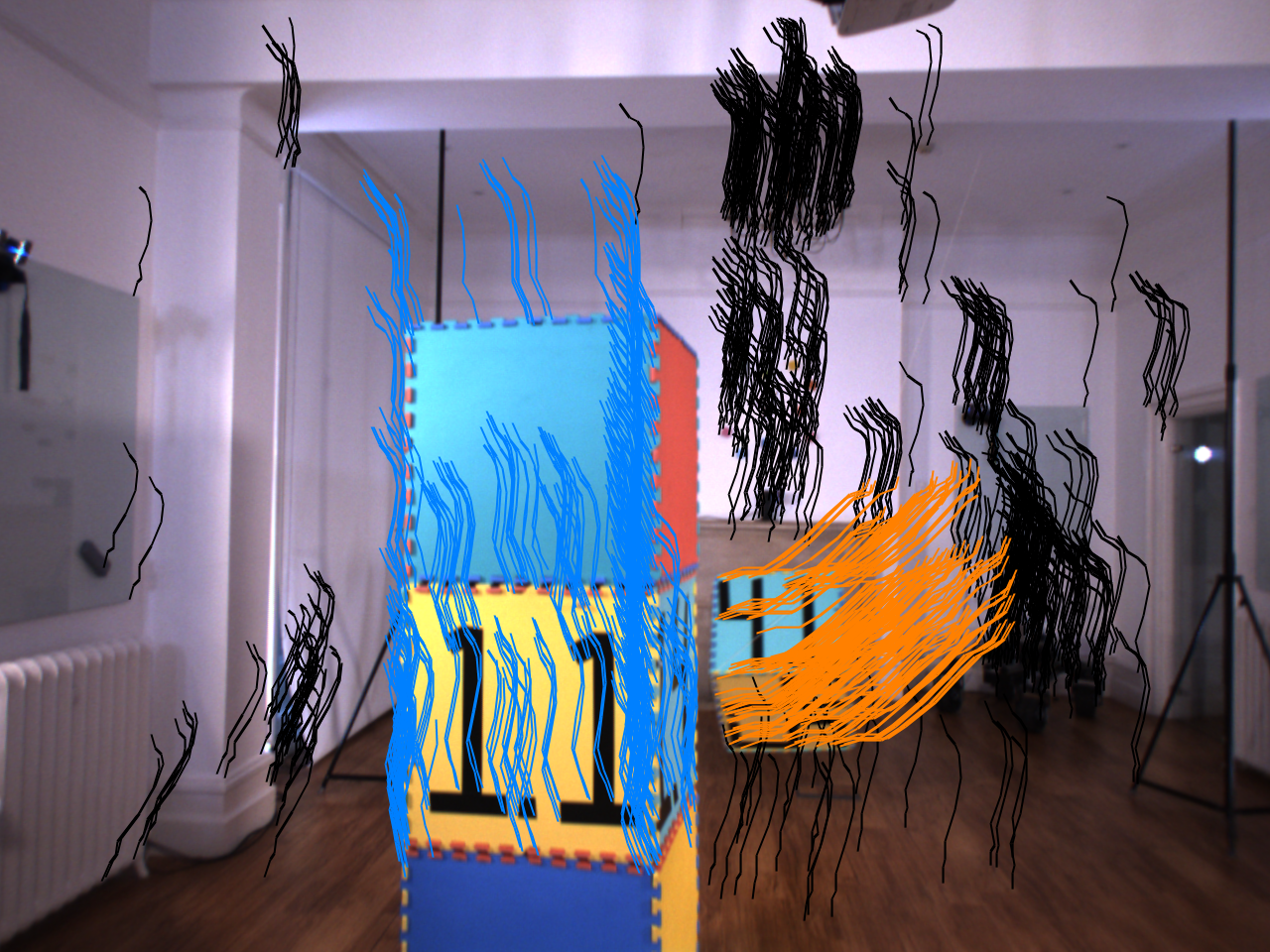}
		\end{subfigure}\\
		\begin{subfigure}{\columnwidth}
			\includegraphics[width=\textwidth]{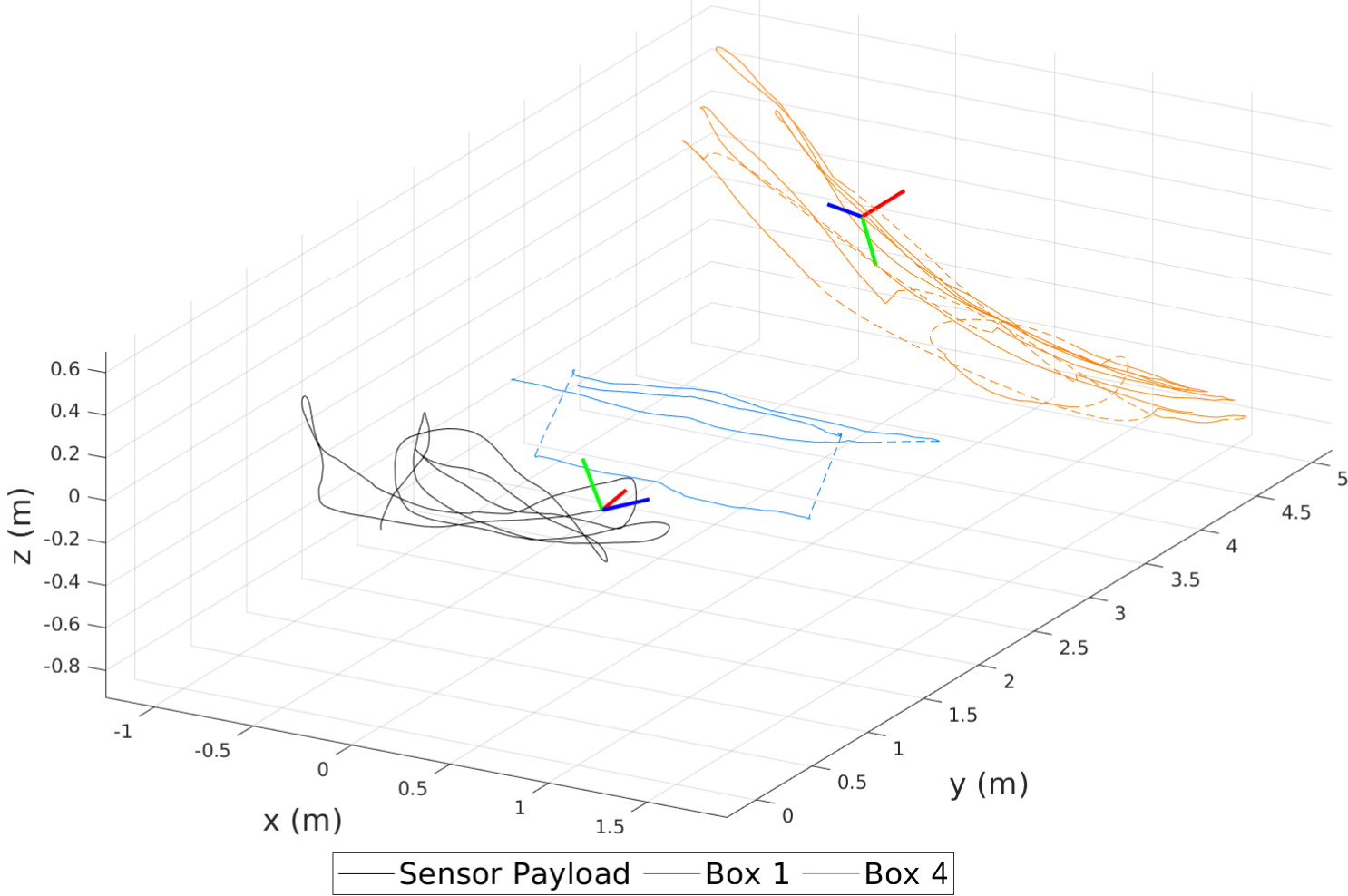}
		\end{subfigure}
		\caption{Motion segmentation (top) and trajectories (bottom) produced by \acs{MVO} using the pose-velocity estimator for the \kjdatasegment{occlusion\_2\_unconstrained} data segment from the \acs{OMD}. 
			The egomotion (black, bottom) of the camera is estimated from the static points in the scene (black, top). 
			The motions of the swinging block (4, orange) and the block tower (1, blue) are segmented and estimated simultaneously with the egomotion.
			The results for each \acs{MVO} estimator for this segment are illustrated in Extensions 4--6 (\cref{app:extensions}).}
		\label{fig:marquee:occlusion}
	\end{figure}
	
	\subsection{Tracking Through Occlusion}\label{sec:evaluation:occlusion}
	The \kjdatasegment{occlusion\_2\_unconstrained} segment of the \ac{OMD} presents multiple \kjse{3} motions with significant occlusions (\cref{fig:marquee:occlusion}).
	It tests the ability of estimators to segment and estimate repeatedly occluded motions.
	It includes three independent motions, a sliding block tower (Block 1) that occasionally becomes static, a swinging block (Block 4) that is repeatedly occluded by the tower, and the camera egomotion.
	A 500-frame section starting at frame 1750 was used in this evaluation.
	The segment is particularly challenging because it includes multiple instances of the swinging block changing direction when partially or fully occluded.
	\looseness=-1

% prevents blank space in previous page. Seems to only work to the next columnbreak.
\afterpage{
	\begin{landscape}
		
		\begin{table}
			\caption{Maximum translational and roll-pitch-yaw errors for each \acs{MVO} estimator and \acs{VDO-SLAM} for the \kjdatasegment{swinging\_4\_unconstrained} data segment from the \acs{OMD}.
				The maximum ground-truth trajectory displacement is also included for comparison.
				Translational xyz errors are given in meters (m) and rotational roll-pitch-yaw errors are given in degrees ($^\circ$) and are not bounded to $\pm180^\circ$.
				Bolded values indicate the best results for each estimated quantity.
		}
			\label{tab:evaluation:global:swinging}
			\centering
			\begin{tabular}{l|cccc|cccc|cccc|cccc|cccc}
				\toprule
				\multicolumn{1}{c}{} & \multicolumn{4}{c}{Pose-Only} & \multicolumn{4}{c}{Pose-Velocity} & \multicolumn{4}{c}{Pose-Velocity-Acceleration}& \multicolumn{4}{c}{VDO-SLAM} & \multicolumn{4}{c}{Max Ground Truth Displacement} \\
				Obj. & xyz & roll & pitch & yaw & xyz & roll & pitch & yaw  & xyz & roll & pitch & yaw & xyz & roll & pitch & yaw & xyz & roll & pitch & yaw \\
				\midrule
				Ego & \textbf{0.08} & -1.46 & \textbf{-0.63} & 1.16  & \textbf{0.08} & -1.48 & \textbf{-0.63} & 1.19 & \textbf{0.08} & \textbf{-1.40} & -0.72 & \textbf{1.15} & 0.46 & -2.74 & 2.07 & 2.74 & 1.07 & -5.76 & -23.44 & -14.58 \\
				B. 1 & \textbf{0.09} & 10.39 & -3.99 & \textbf{-1.31} & 0.10 & 9.29 & \textbf{-3.01} & -1.43 & 0.57 & \textbf{2.83} & 7.24 & -3.96 & 1.06 & -33.13 & 38.80 & 24.90 & 0.87 & 27.77 & -33.96 & -18.96 \\
				B. 2 & 0.31 & \textbf{11.12} & -5.46 & -67.07  & 0.22 & 15.39 & -7.47 & -80.16 & \textbf{0.19} & 11.63 & \textbf{-4.68} & \textbf{-62.68} & 0.40 & 12.29 & 13.62 & -191.42 &  0.26 & -9.38 & 11.16 & -1115.85 \\
				B. 3 & 0.13 & \textbf{2.59} & -3.02 & 3.65  & \textbf{0.12} & 2.73 & \textbf{-2.65} & \textbf{2.50} & 0.14 & 8.46 & 5.51 & -8.57 & 1.30 & 10.16 & -14.57 & 7.08 & 0.71 & -30.88 & -35.15 & 36.19 \\
				B. 4 & 0.55 & 14.80 & 7.06 & 91.69 & 0.41 & 9.52 & -9.57 & 112.42 & \textbf{0.19} & \textbf{8.67} & \textbf{3.65} & \textbf{83.54} & 0.76 & -38.78 & 27.03 & 211.81 & 0.22 & -4.10 & 6.34 & 1484.20 \\
				\bottomrule
			\end{tabular}
		\vspace{-0.85mm}
		\end{table}
		
		\begin{table}
			\caption{Maximum translational and roll-pitch-yaw errors for each \acs{MVO} estimator for the \kjdatasegment{occlusion\_2\_unconstrained} data segment from the \acs{OMD}, as well as the maximum extrapolated errors for each block when occluded. 
				The maximum ground-truth trajectory displacement is also included for comparison.
				Translational xyz errors are given in meters (m) and rotational roll-pitch-yaw errors are given in degrees ($^\circ$) and are not bounded to $\pm180^\circ$.
				Bolded values indicate the best results for each estimated quantity.}
			\label{tab:evaluation:global:occlusion}
			\centering
			\begin{tabular}{lr|cccc|cccc|cccc|cccc}
				\toprule
				\multicolumn{2}{c}{} & \multicolumn{4}{c}{Pose-Only} & \multicolumn{4}{c}{Pose-Velocity} & \multicolumn{4}{c}{Pose-Velocity-Acceleration}&  \multicolumn{4}{c}{Max Ground Truth Displacement} \\
				\multicolumn{2}{c}{Object} & xyz & roll & pitch & yaw & xyz & roll & pitch & yaw  & xyz & roll & pitch & yaw & xyz & roll & pitch & yaw \\
				\midrule
				\multicolumn{2}{l|}{Ego} & 0.32 & \textbf{-4.64} & 1.79 & -4.31 & \textbf{0.12} & -5.02 & -2.12 & \textbf{3.10} & 0.21 & -4.95 & \textbf{-1.64} & -3.43 & 1.32 & -8.40 & 24.81 & -13.48 \\
				\multicolumn{2}{l|}{Block 1} & 0.53 & -3.69 & -20.32 & 33.56 & \textbf{0.46} & \textbf{1.89} & \textbf{3.89} & \textbf{-8.57} & 0.56 & -10.10 & -9.68 & 12.13 & 1.68 & -3.04 & 1.00 & 16.31 \\
				& extrapolated & 1.32 & -3.69 & -20.32 & 33.56 & \textbf{0.46} & \textbf{1.89} & \textbf{4.61} & \textbf{-8.57} & 0.64 & -10.10 & -9.68 & 12.697 & & & &  \\
				\multicolumn{2}{l|}{Block 4} & 0.59 & \textbf{129.16} & 74.63 & 65.37 & \textbf{0.58} & 135.70 & \textbf{58.90} & \textbf{61.27} & 0.92 & -184.36 & -83.17 & 161.91  & 2.22 & -54.24 & -38.62 & 32.82 \\
				& extrapolated & \textbf{1.13} & \textbf{129.16} & 74.63 & 65.37 & 1.39 & 135.70 & \textbf{58.90} & \textbf{61.96} & 3.88 & -184.36 & -83.17 & 161.91  & & & & \\
				\bottomrule
			\end{tabular}
		\vspace{-0.85mm}
		\end{table}

		\begin{table}
			\parbox{.32\linewidth}{
				\caption{Total ground-truth translational and roll-pitch-yaw path length for the  \kjdatasegment{swinging\_4\_unconstrained} data segment from the \acs{OMD}.
					Translational xyz motions is given in meters (m) and rotational roll-pitch-yaw motion is given in degrees ($^\circ$).}
				\label{tab:evaluation:global:total_path_swinging}
				\centering
				\begin{tabular}{l|cccc}
					\toprule
					\multicolumn{1}{c}{} & \multicolumn{4}{c}{Total Path} \\
					Object & xyz & roll & pitch & yaw \\
					\midrule
					Ego & 13.71 & 181.92 & 262.91 & 180.29 \\
					Block 1 & 26.65 & 662.60 & 967.34 & 195.32 \\
					Block 2 & 8.12 & 334.43 & 141.51 & 1115.85 \\
					Block 3 & 18.91 & 280.44 & 1116.73 & 135.41 \\
					Block 4 & 2.79 & 74.30 & 63.84 & 1484.20 \\
					\bottomrule
				\end{tabular}
			}
			\hfill
			\parbox{.32\linewidth}{
				\caption{Total ground-truth translational and roll-pitch-yaw path length for the  \kjdatasegment{occlusion\_2\_unconstrained} data segment from the \acs{OMD}.
					Translational xyz motions is given in meters (m) and rotational roll-pitch-yaw motion is given in degrees ($^\circ$).}
				\label{tab:evaluation:global:total_path_occlusion}
				\centering
				\begin{tabular}{l|cccc}
					\toprule
					\multicolumn{1}{c}{} & \multicolumn{4}{c}{Total Path} \\
					Object & xyz & roll & pitch & yaw \\
					\midrule
					Ego & 12.26 & 319.06 & 314.98 & 241.32\\
					Block 1 &  6.72 & 55.68 & 27.78 & 268.58 \\
					Block 4 & 21.72 & 530.96 & 496.46 & 113.44  \\
					\bottomrule
					\multicolumn{5}{c}{}\\
					\multicolumn{5}{c}{}
				\end{tabular}
			}
			\hfill
			\parbox{.32\linewidth}{
				\caption{Total ground-truth translational and roll-pitch-yaw path length for the \kjdatasegment{Drive 0005} and \kjdatasegment{Odometry 03} data segments from the KITTI Vision Benchmark Suite.
					Translational xyz motions is given in meters (m) and rotational roll-pitch-yaw motion is given in degrees ($^\circ$).}
				\label{tab:evaluation:global:total_path_kitti}
				\centering
				\begin{tabular}{l|cccc}
					\toprule
					\multicolumn{1}{c}{} & \multicolumn{4}{c}{Total Path} \\
					Object & xyz & roll & pitch & yaw \\
					\midrule
					\kjdatasegment{Drive 0005} & 69.43 & 16.47 & 15.59 & 137.59\\
					\kjdatasegment{Odometry 03} & 561.09 & 98.69 & 90.45 & 275.66 \\
					\bottomrule
					\multicolumn{5}{c}{}\\
					\multicolumn{5}{c}{}
				\end{tabular}
			}
		\end{table}

	\end{landscape}
}

\afterpage{
	\begin{landscape}
					
		\begin{table}
			\caption{Maximum translational and roll-pitch-yaw errors for each \acs{MVO} estimator and \acs{VDO-SLAM} for the egomotion in the \kjdatasegment{Drive 0005} and \kjdatasegment{Odometry 03} data segments from the KITTI Vision Benchmark Suite.
				The maximum ground-truth trajectory displacement is also included for comparison.
				Translational xyz errors are given in meters (m) and rotational roll-pitch-yaw errors are given in degrees ($^\circ$) and are not bounded to $\pm180^\circ$.
				Bolded values indicate the best results for each estimated quantity. 
			\acs{VDO-SLAM} results for the \kjdatasegment{Odometry 03} segment were not available.}
			\label{tab:evaluation:global:kitti}
			\centering
			\begin{tabular}{l|cccc|cccc|cccc|cccc|cccc}
				\toprule
				\multicolumn{1}{c}{} & \multicolumn{4}{c}{Pose-Only} & \multicolumn{4}{c}{Pose-Velocity} & \multicolumn{4}{c}{Pose-Velocity-Acceleration}& \multicolumn{4}{c}{VDO-SLAM} & \multicolumn{4}{c}{Max Ground Truth Displacement} \\
				Segment & xyz & roll & pitch & yaw & xyz & roll & pitch & yaw  & xyz & roll & pitch & yaw & xyz & roll & pitch & yaw & xyz & roll & pitch & yaw \\
				\midrule
				\kjdatasegment{Drive 0005} & \textbf{3.17} & \textbf{-0.71} & \textbf{-0.29} & \textbf{-0.59} & 3.26 & -1.07 & 0.54 & -0.65 & 3.22 & -0.77 & -0.33 & -1.03 & 4.17 & 1.94 & -2.03 & 2.42 & 62.25 & -2.42 & 1.56 & 48.40 \\
				\kjdatasegment{Odometry 03} & \textbf{29.0} & -2.19 & -1.25 & \textbf{-5.60} & 31.4 & -2.34 & \textbf{1.00} & -6.27 & 30.1 & \textbf{-2.14} & 1.67 & -5.72 &  &  &  &  & 511.87 & 4.09 & -7.78 & -92.04 \\
				\bottomrule
			\end{tabular}
			\vspace{-0.85mm}
		\end{table}
	
		\begin{table}
			\caption{\acs{RMS} (and maximum) relative translational and axis-angle errors for each \acs{MVO} estimator and \acs{VDO-SLAM} for the \kjdatasegment{swinging\_4\_unconstrained} data segment from the \acs{OMD}.
				The \acs{RMS} ground-truth frame-to-frame motion is also included for comparison.
				Translational xyz errors are given in meters (m) and rotational roll-pitch-yaw errors are given in degrees ($^\circ$).
				Bolded values indicate the best results for each estimated quantity.}
			\label{tab:evaluation:relative:swinging}
			\centering
			\begin{tabular}{l|cc|cc|cc|cc|cc}
				\toprule
				\multicolumn{1}{c}{} & \multicolumn{2}{c}{Pose-Only} & \multicolumn{2}{c}{Pose-Velocity} & \multicolumn{2}{c}{Pose-Velocity-Acceleration}& \multicolumn{2}{c}{VDO-SLAM} & \multicolumn{2}{c}{\acs{RMS} Ground Truth} \\
				Object & xyz & axis-angle & xyz & axis-angle  & xyz & axis-angle & xyz & axis-angle & xyz & axis-angle \\
				\midrule
				Ego & 0.007 (0.018) & 0.293 {(\textbf{1.010})} & 0.002 {(\textbf{(0.007})} & \textbf{0.240} {(1.590)} & \textbf{0.002} {(\textbf{0.007})} & 0.263 {(1.752)} & 0.010 {(0.032)} & 0.579 {(2.568)} & 0.029 {(0.046)} & 0.931 {(2.512)} \\
				Block 1 & \textbf{0.012} {(0.045)} & 1.218 {(5.579)} & 0.005 {(\textbf{0.014})} & \textbf{0.228} {(\textbf{0.924})} & 0.005 {(0.018)} & 0.322 {(1.394)} & 0.023 {(0.113)} & 1.191 {(4.334)} & 0.058 {(0.103)} & 2.587 {(5.210)} \\
				Block 2 & 0.007 {(0.024)} & \textbf{0.350} {(\textbf{1.683})} & 0.007 {(0.032)} & 0.632 {(2.819)} & \textbf{0.005} {(\textbf{0.019})} & 0.453 {(1.975)} & 0.012 {(0.041)} & 0.733 {(4.161)} & 0.018 {(0.030)} & 2.423 {(3.223)}\\
				Block 3 & 0.013 {(0.055)} & 1.361 {(6.057)} & 0.005 {(0.014)} & \textbf{0.234} {(\textbf{1.220})} & \textbf{0.005} {(\textbf{0.012})} & 0.343 {(1.243)} & 0.012 {(0.040)} & 0.682 {(1.630)} & 0.042 {(0.077)} & 2.637 {(5.520)}\\
				Block 4 & \textbf{0.003} {(\textbf{0.009})} & \textbf{0.241} {(\textbf{1.583})} & 0.007 {(0.025)} & 0.747 {(3.304)} & 0.005 {(0.018)} & 0.537 {(2.476)} & 0.019 {(0.088)} & 0.884 {(3.308)} & 0.006 {(0.011)} & 2.989 {(3.608)} \\
				\bottomrule
			\end{tabular}
		\end{table} 

		\begin{table}
			\caption{\acs{RMS} {(and maximum) relative} translational and axis-angle errors for each \acs{MVO} estimator for the \kjdatasegment{occlusion\_2\_unconstrained} data segment from the \acs{OMD}.
				The \acs{RMS} ground-truth frame-to-frame motion is also included for comparison.
				Translational xyz errors are given in meters (m) and rotational roll-pitch-yaw errors are given in degrees ($^\circ$).
				Bolded values indicate the best results for each estimated quantity.}
			\label{tab:evaluation:relative:occlusion}
			\centering
			\begin{tabular}{lr|cc|cc|cc|cc}
				\toprule
				\multicolumn{2}{c}{} & \multicolumn{2}{c}{Pose-Only} & \multicolumn{2}{c}{Pose-Velocity} & \multicolumn{2}{c}{Pose-Velocity-Acceleration}& \multicolumn{2}{c}{\acs{RMS} Ground Truth} \\
				\multicolumn{2}{c}{Object} & xyz & axis-angle  & xyz & axis-angle & xyz & axis-angle & xyz & axis-angle \\
				\midrule
				\multicolumn{2}{l|}{Ego} & 0.009 {(0.125)} & 0.997 {(6.652)} & \textbf{0.005} {(\textbf{0.033})} & \textbf{0.995} {(\textbf{6.642})} & 0.006 {(0.050)} & 1.006 {(6.714)} & 0.026 {(0.015)} & 0.697 {(2.036)} \\
				\multicolumn{2}{l|}{Block 1} & 0.019 {(0.186)} & 0.887 {(9.871)} & \textbf{0.014} {(\textbf{0.101})} & \textbf{0.343} {(\textbf{2.659})} & 0.019 {(0.147)} & 0.673 {(7.195)} & 0.015 {(0.026)} & 1.829 {(5.332)} \\
				& extrapolated & 0.026 {(0.780)} & 0.993 {(7.156)} & \textbf{0.014} {(\textbf{0.101})} & \textbf{0.379} {(\textbf{2.658})} & 0.019 {(0.147)} & 0.659 {(6.610)} &  & \\
				\multicolumn{2}{l|}{Block 4} & 0.026 {(0.158)} & \textbf{1.123} {(7.537)} & 0.027 {(\textbf{0.124})} & 1.198 {(\textbf{4.732})} & \textbf{0.026} {(0.133)} & 2.321 {(12.97)} & 0.050 {(0.119)} & 1.436 {(5.773)} \\
				& extrapolated & 0.029 {(1.123)} & \textbf{1.303} {(\textbf{4.411})} & \textbf{0.029} {(\textbf{0.133})} & 1.346 {(6.100)} & 0.044 {(0.471)} & 2.643 {(7.569)} & &  \\
				\bottomrule
			\end{tabular}
		\end{table}		
		
		\begin{table}
			\caption{\acs{RMS} {(and maximum) relative} translational and axis-angle errors for each \acs{MVO} estimator and \acs{VDO-SLAM} for the \kjdatasegment{Drive 0005} and \kjdatasegment{Odometry 03} data segments from the KITTI Vision Benchmark Suite.
				The \acs{RMS} ground-truth frame-to-frame motion is also included for comparison.
				Translational xyz errors are given in meters (m) and rotational roll-pitch-yaw errors are given in degrees ($^\circ$).
				\acs{VDO-SLAM} results for the \kjdatasegment{Odometry 03} segment were not available.}
			\label{tab:evaluation:relative:kitti}
			\centering
			\begin{tabular}{l|cc|cc|cc|cc|cc}
				\toprule
				\multicolumn{1}{c}{} & \multicolumn{2}{c}{Pose-Only} & \multicolumn{2}{c}{Pose-Velocity} & \multicolumn{2}{c}{Pose-Velocity-Acceleration}& \multicolumn{2}{c}{VDO-SLAM} & \multicolumn{2}{c}{\acs{RMS} Ground Truth} \\
				Segment & xyz & axis-angle & xyz & axis-angle  & xyz & axis-angle & xyz & axis-angle & xyz & axis-angle \\
				\midrule
				\kjdatasegment{Drive 0005} & \textbf{0.050} {(\textbf{0.093})} & 0.083 {(0.180)} & 0.052 {(0.101)} & \textbf{0.077} {(\textbf{0.166})} & 0.051 {(0.130)} & 0.081 {(0.279)} & 0.063 {(0.160)} & 0.087 {(0.189)} & 0.467 {(0.713)} & 0.992 {(1.587)} \\
				\kjdatasegment{Odometry 03} & {0.156} {({0.973})} & 0.080 {(0.294)} & \textbf{0.145} {(\textbf{0.723})} & \textbf{0.090} {(\textbf{0.825})} & 0.164 {(1.253)} & 0.091 {(0.959)} &  &  & 0.738 {(0.972)} & 0.642 {(3.260)} \\
				\bottomrule
			\end{tabular}
		\end{table}
		
	\end{landscape}
}

%%%%%%%
% Kitti Odometry 03
% Total Path GT
% Xyz, r, p, y
% 561.0850   98.6786   90.4538  275.6607
% xyz max, xyz percent, r, p, y 
% 511.8684    0.9123  4.0867   -7.7810  -92.0421 
% relative max xyz, rms xyz, max rpy, rms rpy
%  0.9721    0.7379    3.2591    0.6422

% Discrete
% xyz max, xyz percent, r, p, y 
% 29.0334    0.0517   -2.1943    1.2526   -5.6007
% relative max xyz, rms xyz, max rpy, rms rpy
% 0.9732    0.1562    0.2944    0.0802

% WNOA
% xyz max, xyz percent, r, p, y 
% 31.3936    0.0560   -2.3426    1.0018   -6.2694
% relative max xyz, rms xyz, max rpy, rms rpy
% 0.7228    0.1451    0.8246    0.0898

% WNOJ
% xyz max, xyz percent, r, p, y 
% 30.0651 0.0536  -2.1432    1.6676   -5.7181
% relative max xyz, rms xyz, max rpy, rms rpy
% 1.2529    0.1641    0.9587    0.0905
	
	The sliding block tower is partially occluded multiple times when it partially leaves the field of view of the camera, further complicating the estimation.
	It also occasionally stops moving and becomes part of the static background, which is a form of indirect occlusion in motion-based tracking.
	\Ac{VDO-SLAM} was not evaluated on this segment because it is not designed for this level of occlusion.
	\looseness=-1

	\subsection{Urban Driving}\label{sec:evaluation:kitti}
	The KITTI Vision Benchmark Suite \citep{geiger2012} is a collection of datasets and benchmarks dedicated to autonomous driving scenarios.
	The suite contains data and evaluation metrics for object detection and tracking, depth and flow calculations, and \ac{VO}.
	The datasets and benchmarks within KITTI are widely used to develop and evaluate computer vision techniques, but it does not contain any ground-truth third-party trajectory data.
	This makes it less suitable for quantitatively evaluating multimotion estimation algorithms; however, it is still a useful qualitative tool for evaluation in a real-world scenario.

	\begin{figure}[t]
		\begin{subfigure}{0.48\textwidth}
			\includegraphics[width=\textwidth]{figures/kitti/kitti_segmentation.png}
		\end{subfigure}\\
		\begin{subfigure}{0.48\textwidth}
			\includegraphics[width=\textwidth]{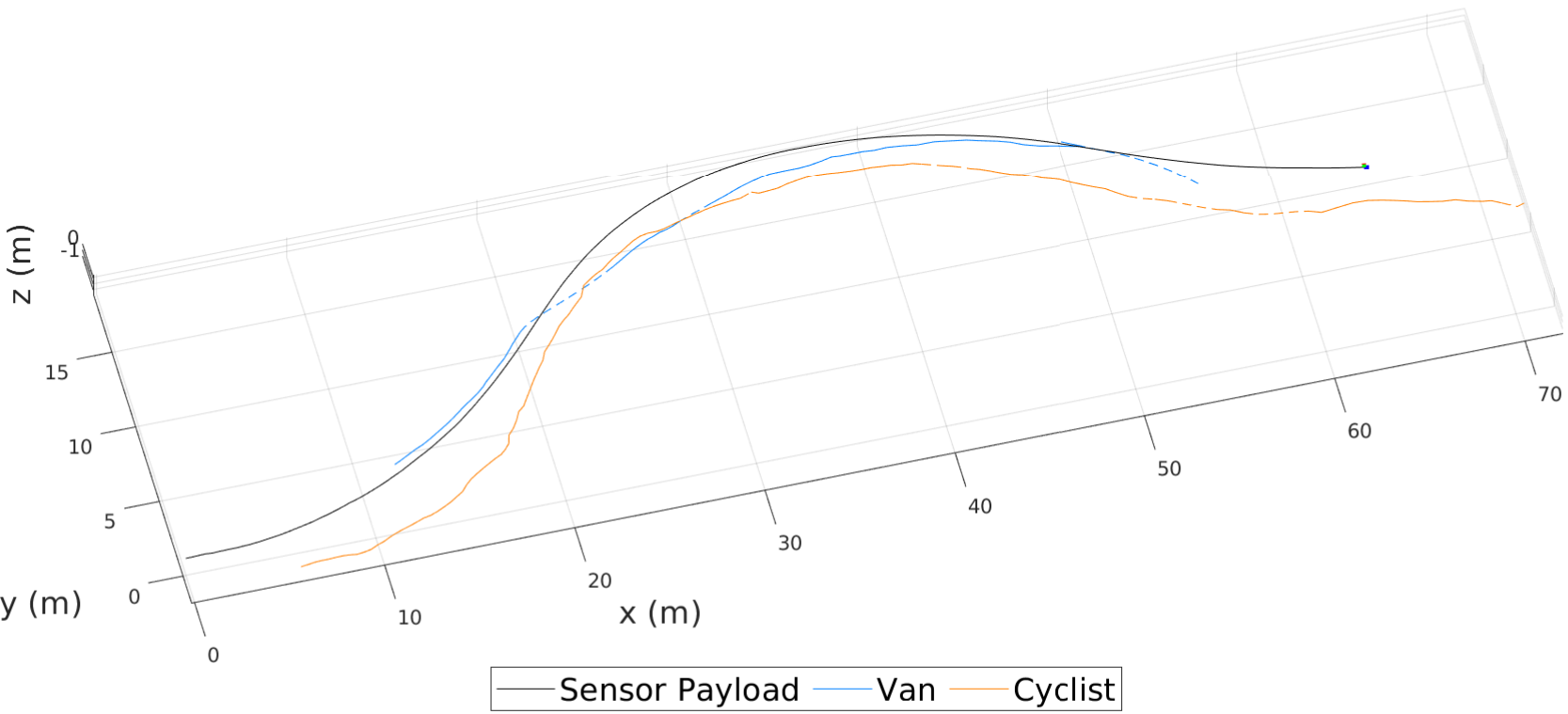}
		\end{subfigure}
		\caption{Motion segmentation (top) and trajectories (bottom) produced by \acs{MVO} using the pose-velocity estimator for the \kjdatasegment{Drive 0005} data segment from the KITTI Vision Benchmark Suite. 
			The egomotion (black, bottom) of the camera is estimated from the static points in the scene (black, top). 
			The motions of the cyclist (orange) and the van (blue) are segmented and estimated simultaneously with the egomotion.
			These results are illustrated in Extension 7 (\cref{app:extensions}).
		}
		\label{fig:marquee:kitti}
	\end{figure}  
	
	The KITTI \kjdatasegment{Drive 0005} segment (\cref{fig:marquee:kitti}) is a 154-frame sequence that contains two independent motions, a cyclist and van, observed by a car-mounted camera as they travel around two consecutive turns.
	The van is temporarily occluded multiple times as it rounds a turn, is obscured by the cyclist, or is overexposed by sunlight.
	These all contribute to a challenging tracking and estimation problem that is compounded by the significant distance between the van and the camera.
	\looseness=-1	
	
	The KITTI \kjdatasegment{Odometry 03} segment, is a longer 800-frame sequence that only contains one independent motion, another car, observed for part of the segment.
	This segment is useful for evaluating egomotion estimation over a significant distance but is not not particularly interesting for multimotion estimation.

	\section{Discussion}\label{sec:discussion}
	Qualitative and quantitative evaluations show that \ac{MVO} consistently estimates ego- and third-party motions, even in the presence of temporary occlusions.
	\ac{MVO} demonstrates good estimation and tracking accuracy on the evaluated toy problems and a real-world driving scenario without relying on appearance-based detectors or preprocessing segmentation stages.

	\ac{MVO} uses sparse feature tracking which does not require prior knowledge about object appearance but has known limitations (\cref{sec:discussion:limitations}).
	The distribution of features on each object affects the estimation of its motion, and small or distant objects can be difficult to segment and estimate.
	Similarly, \ac{MVO} demonstrates that motion-based tracking through occlusions is a reliable way to address the \ac{MEP} without making appearance-based assumptions, but accurately estimating the rotation of a previously occluded object without modeling its shape is challenging.
	Some of the limitations of \ac{MVO} are inherent to the estimators it employs, each bringing its own advantages and disadvantages (\cref{sec:discussion:comparison}).
	Despite its limitations, \ac{MVO} outperforms \ac{VDO-SLAM} in both egomotion and third-party trajectory estimation accuracy (\cref{sec:discussion:vdo}).
	\looseness=-1

	\vspace{-1mm}
	\subsection{Limitations of Sparse Approaches}\label{sec:discussion:limitations}
	\ac{MVO} is a sparse, feature-based approach and therefore its estimation accuracy depends on the distribution of features on each dynamic object.
	This distribution affects the nature of the estimated motion, the observed shape of the object, and the motion extrapolation during occlusions.
	
	Features are only observed on object surfaces facing the sensor, which makes it difficult to estimate certain types of motions, such as translations along the optical axis and rotations not around the optical axis.
	For example, Block 2 and Block 4 in the \kjdatasegment{swinging\_4\_unconstrained} segment  (\cref{fig:results:swinging:discrete,fig:results:swinging:wnoa,fig:results:swinging:wnoj}) both rotate significantly around the z-axis, leading to larger rotational errors.
	Not only are these rotations outside the image plane, but it is also difficult to differentiate between an object rotating in place and orbiting about an offset axis.
	These challenges could be overcome by using longer estimation windows or incorporating 3D fusion techniques \citep{runz2017}.
	Despite this, the yaw errors for both Block 2 and Block 4 are less than $7.5\%$ of their total yaw rotation for all \ac{MVO} estimators.
	
	The observable shape (and centroid) of an object also changes over time as it and the sensor move.
	This shifts the observable features on the object and the geocentric trajectory estimate, which degrades the interpolated estimates.
	This is often most evident before and after an occlusion, as seen in the estimates for the block tower in the \kjdatasegment{occlusion\_2\_unconstrained} segment  (\cref{fig:results:occlusion:discrete:1,fig:results:occlusion:wnoa:1,fig:results:occlusion:wnoj:1}).
	The block tower often extends outside the camera frustum while stopped (i.e., occluded for motion-based tracking), which leads to a large discrete jump in its observed centroid when it is tracked again.
	These discrete jumps create interpolation errors during motion closure and lead to considerably higher global errors than would otherwise be expected given the small relative error (\cref{tab:evaluation:global:occlusion,tab:evaluation:relative:occlusion})
	
	The transition into occlusion itself presents estimation challenges.
	As the object becomes more occluded, the number of tracked features decreases and the quality of the segmentation and estimation degrades.
	This is evident in the egomotion and block tower estimates for the \kjdatasegment{occlusion\_2\_unconstrained} segment, which are occasionally corrupted when the the block tower slows to a stop and an increasing number of its features are incorrectly classified as part of the background (\cref{fig:results:occlusion:discrete,fig:results:occlusion:wnoa,fig:results:occlusion:wnoj}).
	
	The transition into occlusion also makes extrapolation harder, because the erroneous estimates before an occlusion are extrapolated forward and make motion closure more difficult.
	This is evident in the swinging block estimates for the \kjdatasegment{occlusion\_2\_unconstrained} segment, which are often poorly extrapolated (\cref{fig:results:occlusion:discrete:4,fig:results:occlusion:wnoa:4,fig:results:occlusion:wnoj:4}).
	Objects rarely become occluded or unoccluded instantaneously and performance in the transitional periods could be improved by predicting occlusions \citep{mitzel2010} or more sophisticated centroid calculation \citep{setterfield2018}.
	
	\subsection{Estimator Comparison}\label{sec:discussion:comparison}
	\ac{MVO} is a direct extension of \ac{VO} to the \ac{MEP} and can therefore leverage significant developments in egomotion estimation.
	This includes different \kjse{3} estimation techniques, such as the discrete, \ac{WNOA}, and \ac{WNOJ} motion models presented here.
	The associated estimators have advantages and disadvantages in \ac{MVO}.
	
	The pose-only estimator is the most general of those discussed in this paper.
	Its lack of motion prior makes it simple, but it can estimate physically implausible motions (e.g., discrete jumps) and it is not as robust to occlusions.
	The pose-velocity and pose-velocity-acceleration estimators constrain the trajectory estimate with a \ac{WNOA} and \ac{WNOJ} prior, respectively.
	This makes estimation more complex but penalizes physically implausible motions and allows for more accurate extrapolation and interpolation through occlusions.
	The \ac{WNOA} prior assumes a locally constant velocity that is  appropriate for many real-world situations but does not accurately model motions with large changes in direction.
	The \ac{WNOJ} prior assumes a locally constant acceleration that is applicable to more real-world situations, including objects with large changes in direction; however, it often does not sufficiently constrain its estimates to accurately extrapolate motions for long periods of time.
	
	\subsubsection{Swinging Blocks Segment}
	All estimators performed similarly well in the \kjdatasegment{swinging\_4\_unconstrained} segment, which is the simplest scenario without occlusion (\cref{sec:evaluation:swinging}).
	They all consistently segmented each motion and accurately estimated the motion of the camera and Blocks 1 and 3 but underestimated the yaw rotation of Blocks 2 and 4 (\cref{fig:results:swinging:discrete,fig:results:swinging:wnoa,fig:results:swinging:wnoj,tab:evaluation:global:swinging,tab:evaluation:global:total_path_swinging,tab:evaluation:relative:swinging}).
	This yaw rotation is an out-of-plane rotation that is hard to observe and its estimation may be improved with appearance-based object models.
	The global and relative estimation errors for the different estimators and \ac{VDO-SLAM} are given in \cref{tab:evaluation:global:swinging,tab:evaluation:relative:swinging} and shows that the \ac{MVO} estimators all perform similarly while outperforming \ac{VDO-SLAM}.
	
	\subsubsection{Occlusion Segment}\label{sec:discussion:comparison:occlusion}
	The estimators perform differently in the presence of occlusions, as seen in the \kjdatasegment{occlusion\_2\_unconstrained} results (\cref{sec:evaluation:occlusion}).
	\ac{MVO} consistently estimates the motions of both the camera and the moving blocks through motion closure with all estimators; however, the newly calculated centroid of a motion after an occlusion does not always match that of the original motion.
	This discrepancy causes jumps in the trajectory, as the original centroid is projected forward to a different location on the object than previously estimated.
	This affects the calculation of the correction transform (\cref{sec:sliding:motion_closure:correcting}) and is a major source of error in the estimates of the block tower, as it is often partially outside the view of the camera.
	The global and relative estimation errors for the different estimators are given in \cref{tab:evaluation:global:occlusion,tab:evaluation:relative:occlusion} and show that the pose-velocity estimator was generally the most accurate.
	
	The pose-only estimator performed reasonably well for directly observed motions but did not perform well on occlusions  (\cref{fig:results:occlusion:discrete,tab:evaluation:global:occlusion,tab:evaluation:global:total_path_occlusion,tab:evaluation:relative:occlusion}).
	This is because its linear extrapolation and interpolation are sensitive to estimation noise, which often increases as an object becomes occluded (\cref{sec:discussion:limitations}).
	The egomotion estimate is also affected by the sliding block transitioning between being stationary and dynamic more than the other estimators because of the lack of motion prior.
	
	The pose-velocity estimator performed well for the entire segment and was the best at extrapolating occluded motions (\cref{fig:results:occlusion:wnoa,tab:evaluation:global:occlusion,tab:evaluation:global:total_path_occlusion,tab:evaluation:relative:occlusion}).
	Its locally constant-velocity assumption reasonably models many motions and allows for gradual changes in direction when directly observed.
	When changes in motion are not observed (i.e., during occlusion), the extrapolation can be inaccurate and motion closure may fail.
	This is expected because a \ac{WNOA} prior only allows for small changes in velocity between consecutive frames, so it distributes large changes in motion over longer periods of time.
	
	The pose-velocity-acceleration estimator was the best at interpolating occluded motions, but occasionally underconstrained its extrapolated estimates (\cref{fig:results:occlusion:wnoj,tab:evaluation:global:occlusion,tab:evaluation:global:total_path_occlusion,tab:evaluation:relative:occlusion}).
	Its locally constant-acceleration assumption directly models changes in velocity, even during occlusion.
	This more expressive prior results in better interpolations after motion closure than the \ac{WNOA} prior but is less robust when extrapolating motions.
	This is because even small nonzero accelerations are extrapolated to significant motions over a moderate-length occlusion, as evidenced by the high extrapolation error for the swinging block.
	
	All estimators had high rotational error after occlusion because motion closure does not correct the extrapolated rotation (\cref{sec:sliding:motion_closure}).
	This results in similar maximum rotation errors when extrapolating and interpolating (\cref{tab:evaluation:global:swinging,tab:evaluation:global:occlusion,tab:evaluation:global:kitti}).
	Accurately recognizing the relative rotation of an object after an occlusion is difficult without modeling the appearance or structure of the object.
	This may be achieved by tracking the visual features of the object or by building a model of its structure.
	
	\subsubsection{KITTI Segments}\label{sec:discussion:comparison:kitti}
	\ac{MVO} consistently estimates the motions of the camera and the cyclist throughout the \kjdatasegment{Drive 0005} segment, and the estimators perform similarly for egomotion estimation (\cref{sec:evaluation:kitti}).
	The van is directly or indirectly occluded several times, but \ac{MVO} reliably extrapolates and tracks its motion through motion closure for most of the segment until the van becomes too far away for accurate stereo triangulation.
	\Ac{VDO-SLAM} demonstrates a similar ability to track the cyclist and the van.
	
	The egomotion estimation error for the different estimators and \ac{VDO-SLAM} on the \kjdatasegment{Drive 0005} and \kjdatasegment{Odometry 03} segments is given in \cref{tab:evaluation:global:kitti,tab:evaluation:relative:kitti} and is less than $5.6\%$ of path length for all \ac{MVO} estimators.
	These errors represent a similar percentage of the ground-truth motion as in the toy problems (\cref{tab:evaluation:global:kitti,tab:evaluation:global:total_path_kitti,tab:evaluation:relative:kitti,fig:results:kitti,fig:results:kitti_od}).
	The segment does not include ground-truth trajectory data for third-party motions, so it is not useful for comparing the estimators beyond egomotion accuracy.
	
	\vspace{0.5\baselineskip}
	
	The decision of which estimator to use in any given application can be dependent on the requirements and characteristics of the situation.
	Based on the performance demonstrated on these segments, the \ac{WNOA} is likely the most reliable prior for the general \ac{MEP}.
	The \ac{WNOJ} prior proved most accurate for interpolation, and it could likely be combined with the \ac{WNOA} estimator for improved performance in cases with significant occlusion of nonconstant motions.
	
	\subsection{\acs{MVO} Comparison}\label{sec:discussion:vdo}
	\ac{MVO} and \ac{VDO-SLAM} demonstrate similar tracking performance, but \ac{MVO} outperforms \ac{VDO-SLAM} for metric estimation of both ego- and third-party motions for the \kjdatasegment{swinging\_4\_unconstrained} (\cref{sec:evaluation:swinging}) and \kjdatasegment{Drive 0005} (\cref{sec:evaluation:kitti}).
	\ac{MVO} achieves this without relying on learning-based preprocessing steps or other appearance-based information and can track motions through significant occlusions.
	
	\ac{VDO-SLAM} prioritizes estimating relative object motions in outdoor scenes with minimal rotation, such as autonomous driving scenarios, and is not designed for significant occlusion, so it is not evaluated on \kjdatasegment{occlusion\_2\_unconstrained}.
	It is most comparable to \ac{MVO} in egomotion accuracy, especially for the \kjdatasegment{Drive 0005} segment (\cref{tab:evaluation:global:kitti,tab:evaluation:relative:kitti}), but this design focus limits its applicability to the general \ac{MEP}.
	\looseness=-1

	Unlike \ac{VDO-SLAM}, \ac{MVO} is reliant on motion, so objects that temporarily move similarly may be given the same label, such as when the sliding block tower becomes stationary (\cref{sec:evaluation:occlusion}).
	Extrapolating lost motions even when they are similar to existing motions allows temporarily similar motions to be resegmented once they diverge, providing a form of \emph{motion permanence}.
	This allows the block tower to be successfully tracked through motion-based occlusions, but \ac{MVO} still had higher egomotion errors in the \kjdatasegment{occlusion\_2\_unconstrained} segment than the \kjdatasegment{swinging\_4\_unconstrained} segment because the block tower's transitions between dynamic and stationary motion created noise in the egomotion label.
	\ac{VDO-SLAM} uses front-end instance segmentation to avoid the difficulty of distinguishing between similarly moving objects, instead identifying objects to track based on appearance.
	Incorporating similar appearance-based techniques into \ac{MVO} could improve its performance when independent objects temporarily move similarly.
	
	\Ac{MVO} and \ac{VDO-SLAM} qualitatively demonstrate a similar ability to track both the cyclist and van in the \kjdatasegment{Drive 0005} segment.
	Both consistently segment and estimate the cyclist until failing to track the van at the same time in the segment, though metric comparisons of their third-party estimates are not possible for this dataset due to its lack of ground-truth data.
	\ac{VDO-SLAM} achieves this by relying on separate preprocessing algorithms to perform instance segmentation and optical flow calculation.
	\ac{MVO} has no such dependencies, segmenting the scene and estimating trajectories based on the bulk motion of tracked feature points.
	This means it is applicable to a variety of different environments without significant tuning or prior knowledge of object number, appearance, or motion.
	
	\section{Conclusion}\label{sec:conclusion}
	This paper presents \ac{MVO} as an extension of the traditional \ac{VO} pipeline to address
	the \ac{MEP}.
	It uses multilabeling techniques to simultaneously segment and estimate all rigid motions in a scene.
	A variety of motion-based trajectory states are explored, and their effect on the ability to estimate independent \kjse{3} motions and track them through occlusions is discussed. 
	
	A stereo implementation of \ac{MVO} is evaluated both quantitatively and qualitatively using the \ac{OMD} \citep{judd2019ral} and the KITTI Vision Benchmark Suite \citep{geiger2012}.
	\ac{MVO} achieves egomotion estimation accuracy comparable to similarly defined egomotion-only \ac{VO} systems, and it outperforms an appearance-based multimotion system \citep{zhang2020arxiv} in addressing the \ac{MEP}.
	
	The pipeline operates directly on tracked 3D points and is agnostic to the type of sensor that generates them, so a variety of sensors and corresponding estimators can be employed in \ac{MVO} \citep{judd2019thesis}.
	\ac{MVO} also relies exclusively on motion-based estimation and tracking techniques.
	Future work will include extending \ac{MVO} to other sensors, such as LiDAR and event cameras, parallelizing the batch estimation in order to run at real time, and introducing application-specific, appearance-based techniques to improve performance in specific environments.
	
	\section{Acknowledgments}
	We would like to thank Paul Amayo for providing his implementation of CORAL and the authors of VDO-SLAM for providing their results and discussing them.
	We would also like to thank the editorial board for considering this manuscript and our reviewers for their time and constructive criticism.
	
	This research was funded by UK Research and Innovation and EPSRC through Robotics and Artificial Intelligence for Nuclear (RAIN) [EP/R026084/1], ACE-OPS: From Autonomy to Cognitive assistance in Emergency OPerationS [EP/S030832/1], and the Autonomous Intelligent Machines and Systems (AIMS) Centre for Doctoral Training (CDT) [EP/S024050/1].

	\appendix
	\section{Index to Multimedia Extensions}\label{app:extensions}
	The multimedia extensions in \cref{tab:multimedia} are available at \url{https://www.youtube.com/c/roboticesp}.
	
	\begin{table}[h]
		\caption{Index to Multimedia Extensions.}\label{tab:multimedia}
		\begin{tabular}{c|c|p{0.55\columnwidth}}
			\toprule
			Extension & \multicolumn{1}{c|}{\begin{tabular}{@{}c@{}}Media\\Type\end{tabular}} & Description\\
			\midrule
			1 & Video & Results of the pose-only estimator on the \kjdatasegment{swinging\_4\_unconstrained} segment of the \acs{OMD} \\
			2 & Video & Results of the pose-velocity estimator on the \kjdatasegment{swinging\_4\_unconstrained} segment of the \acs{OMD} \\
			3 & Video & Results of the pose-velocity-acceleration estimator on the \kjdatasegment{swinging\_4\_unconstrained} segment of the \acs{OMD} \\
			4 & Video & Results of the pose-only estimator on the \kjdatasegment{occlusion\_2\_unconstrained} segment of the \acs{OMD} \\
			5 & Video & Results of the pose-velocity estimator on the \kjdatasegment{occlusion\_2\_unconstrained} segment of the \acs{OMD} \\
			6 & Video & Results of the pose-velocity-acceleration estimator on the \kjdatasegment{occlusion\_2\_unconstrained} segment of the \acs{OMD} \\
			7 & Video & Results of the pose-velocity estimator on the \kjdatasegment{Drive 0005} segment of KITTI\\
			\bottomrule
		\end{tabular}
	\end{table}
	
	\section{Nonlinear Stereo Camera Model}
	\label{app:stereo}
	A calibrated stereo camera pair can be described by its intrinsic calibration matrix,
	\begin{equation*}
		\mathbf{K} = \begin{bmatrix}
			f_u & 0 & u_{0} & 0\\
			0 & f_v & v_{0} & 0\\
			0 & 0 & 1 & 0
		\end{bmatrix},
	\end{equation*}
	and its horizontal baseline $b$.
	The horizontal and vertical focal lengths, $f_u$ and $f_v$, affect the camera field and depth of view.
	The principal point coordinates, $u_{0}$ and $v_{0}$, define the camera center in the image plane.
	The nonlinear stereo camera projection, $\mathbf{s}_{\text{stereo}}\left(\cdot\right)$, projects a 3D world point onto the image plane according to
	\begin{equation*}
		\begin{bmatrix}
			u\\
			v\\
			d\\
			1
		\end{bmatrix} = \mathbf{s}_{\text{stereo}}\left(\begin{bmatrix}x\\y\\z\end{bmatrix}\right) \coloneqq \begin{bmatrix}
			\frac{f_{u}x}{z} - u_0\\
			\frac{f_{v}y}{z} - v_0\\
			\frac{f_{u}b}{z}\\
			1
		\end{bmatrix},
	\end{equation*}
	where $u$ and $v$ are the horizontal and vertical left-image coordinates and $d$ is the horizontal disparity between the image coordinate in the left and right image.
	
	This projection function has the corresponding Jacobian,
	\begin{equation*}
		\mathbf{S}_{jk} \coloneqq \left.\frac{\delta\mathbf{s}}{\delta\mathbf{z}}\right|_{\mathbf{z}\left(\mathbf{x}_{\text{op},jk}\right)}=\begin{bmatrix}
			\frac{f_{u}}{z_{3}} & 0 & -\frac{f_{u}z_{1}}{z_{3}^2} & 0\\
			0 & \frac{f_{v}}{z_{3}} & -\frac{f_{v}z_{2}}{z_{3}^2} & 0\\
			0 & 0 & -\frac{f_{u}b}{z_{3}^2} & -\frac{f_{u}b}{z_{3}}
		\end{bmatrix},
	\end{equation*}
	where $\mathbf{z}\left(\mathbf{x}_{\text{op},jk}\right) = \begin{bmatrix}z_1 & z_2 & z_3 & 1\end{bmatrix}^{T}$.
	This Jacobian is used in the bundle adjustment optimization in \cref{sec:estimation} applied to the stereo camera data in \cref{sec:evaluation}.

	\begin{figure}[t]
		\begin{subfigure}{0.48\textwidth}
			\includegraphics[width=\textwidth]{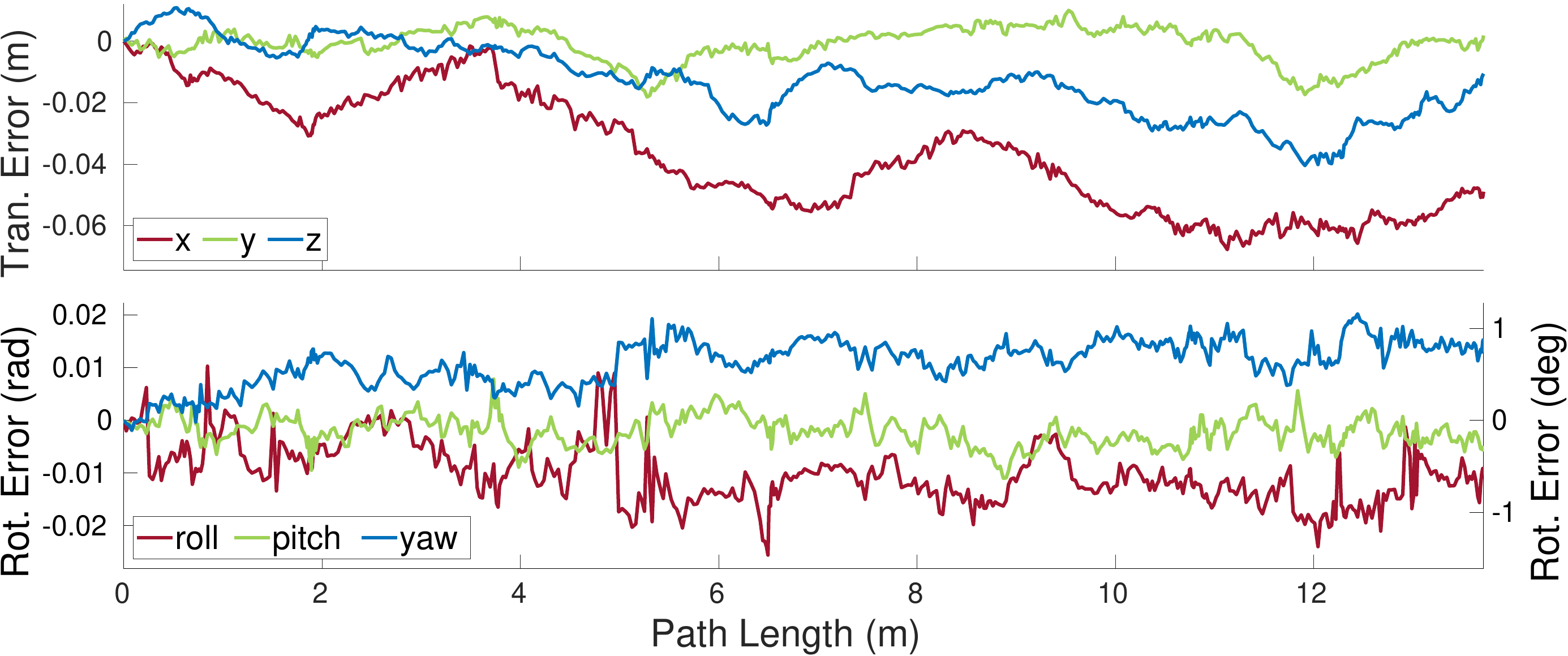}
			\vspace{-6mm}
			\caption{Camera Egomotion}
			\label{fig:results:swinging:discrete:ego}%
			\vspace{6mm}
		\end{subfigure}\\
		\begin{subfigure}{0.48\textwidth}
			\includegraphics[width=\textwidth]{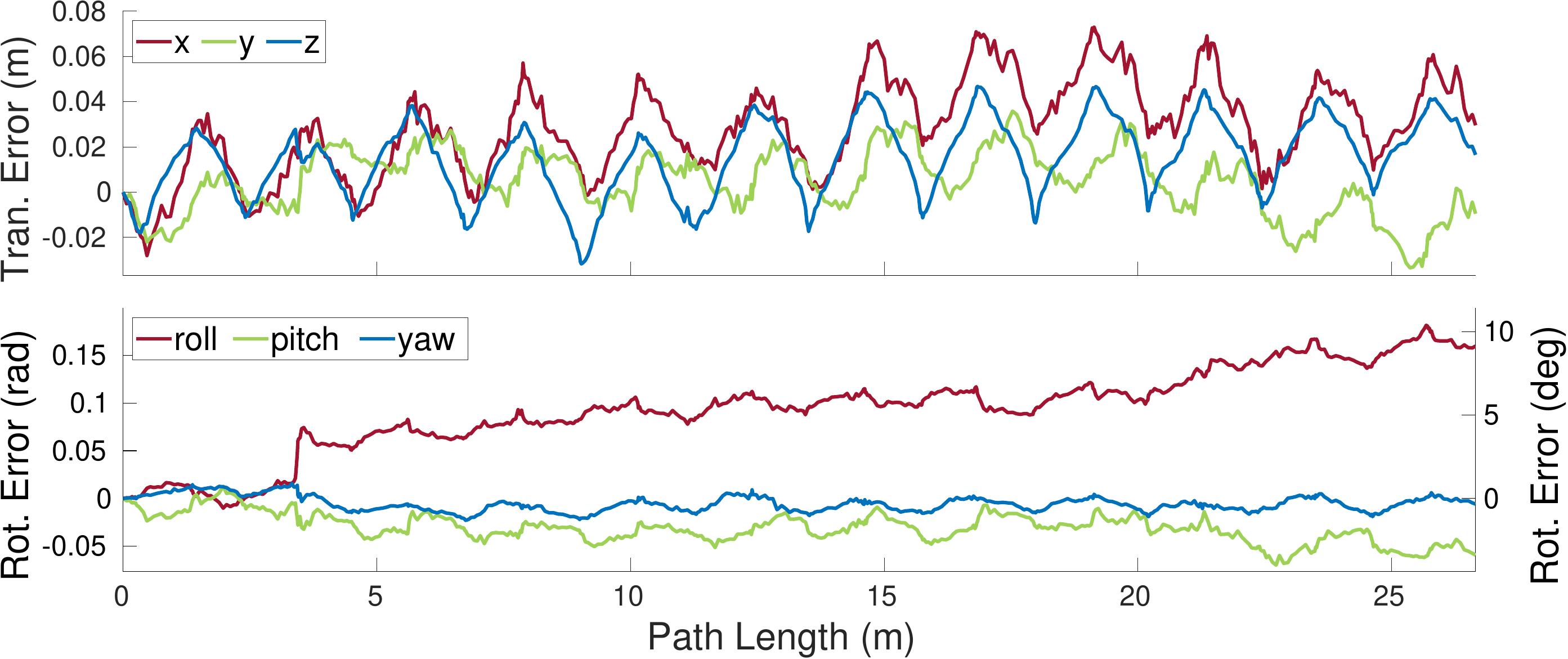}
			\vspace{-6mm}
			\caption{Block 1}%
			\label{fig:results:swinging:discrete:1}%
			\vspace{6mm}
		\end{subfigure}\\
		\begin{subfigure}{0.48\textwidth}
			\includegraphics[width=\textwidth]{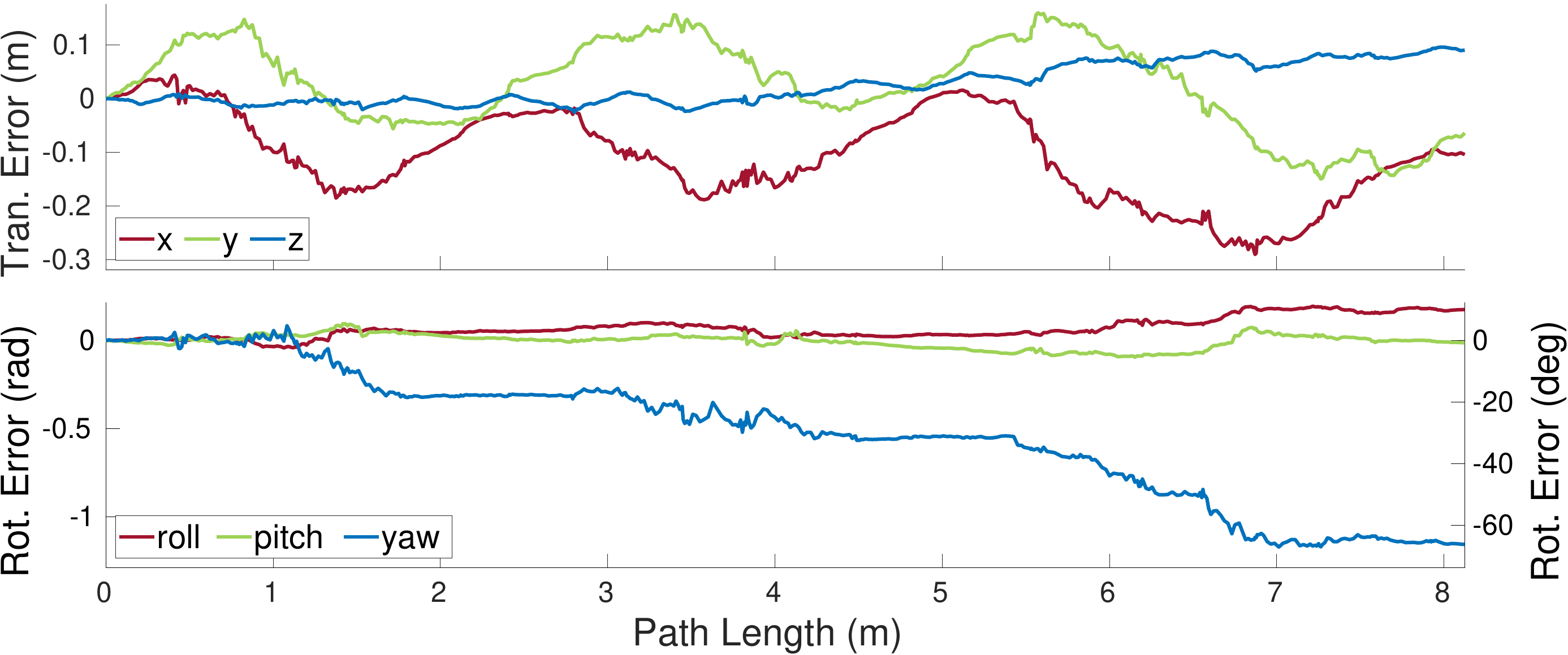}
			\vspace{-6mm}
			\caption{Block 2}%
			\label{fig:results:swinging:discrete:2}%
			\vspace{6mm}
		\end{subfigure}\\
		\begin{subfigure}{0.48\textwidth}
			\includegraphics[width=\textwidth]{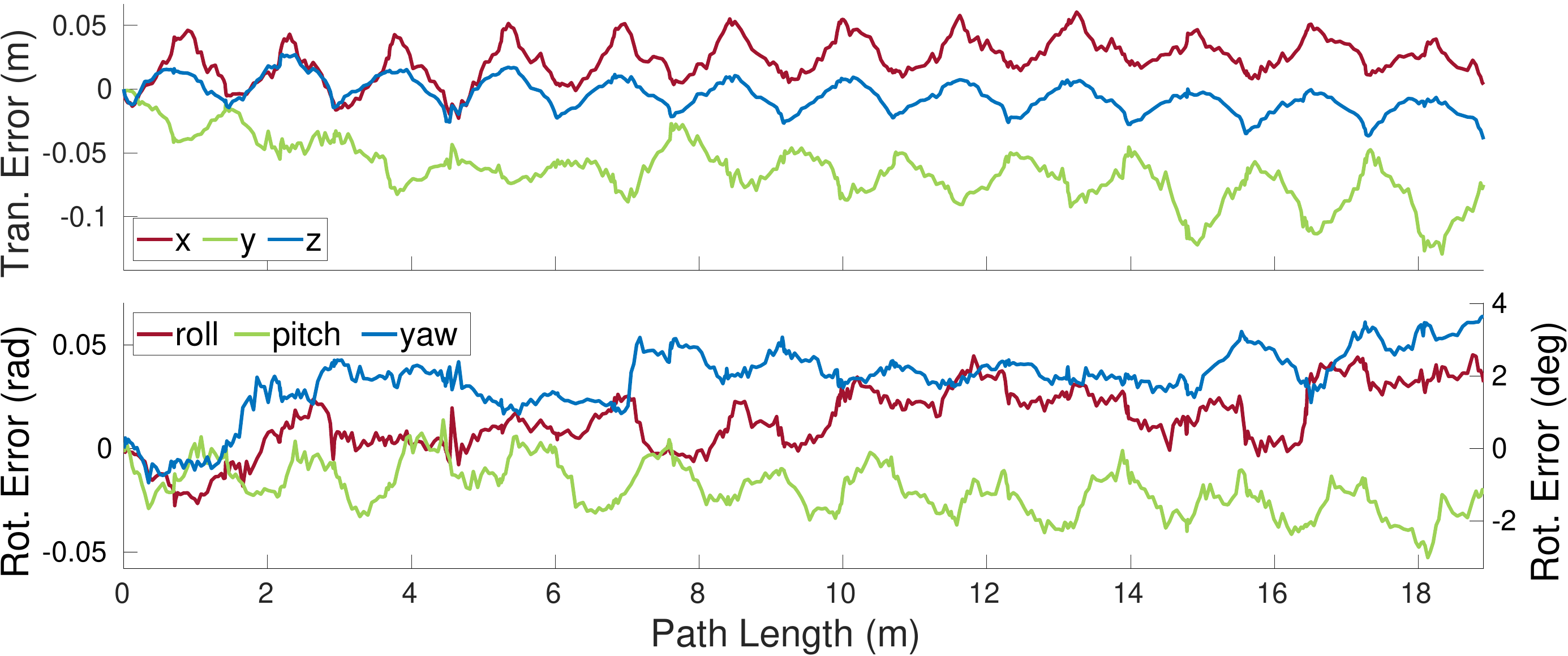}
			\vspace{-6mm}
			\caption{Block 3}%
			\label{fig:results:swinging:discrete:3}%
			\vspace{6mm}
		\end{subfigure}\\
		\begin{subfigure}{0.48\textwidth}
			\includegraphics[width=\textwidth]{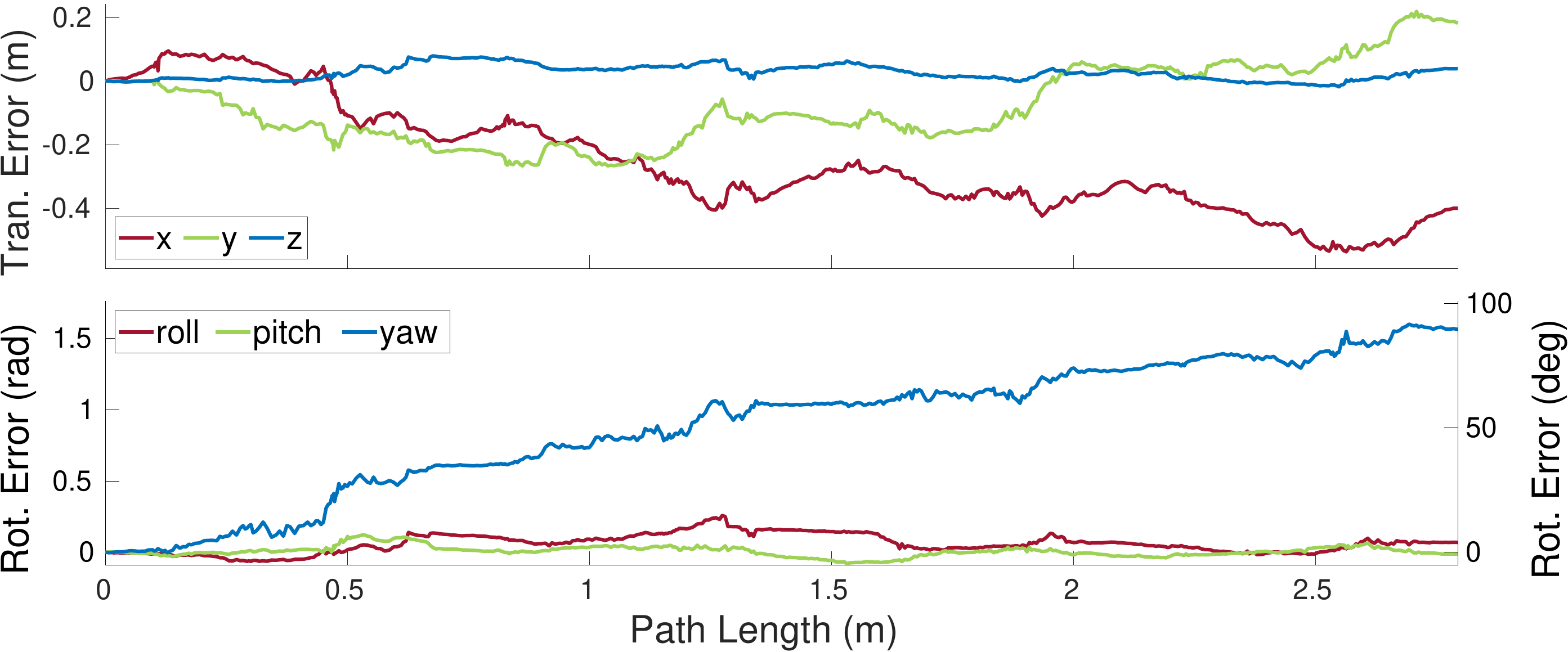}
			\vspace{-6mm}
			\caption{Block 4}%
			\label{fig:results:swinging:discrete:4}%
		\end{subfigure}
		\caption{The translational and rotational errors estimated by \acs{MVO} for the camera (a) and the swinging blocks (b--e) for a 500-frame section of the \kjdatasegment{swinging\_4\_unconstrained} segment from the \acs{OMD} using the pose-only estimator.
			Each object motion is compared to ground-truth trajectory data and errors are reported in an arbitrary geocentric frame with the z-axis up and arbitrary x- and y-axes.
		}%
		
		\label{fig:results:swinging:discrete}%
	\end{figure}
	
	\begin{figure}[t]
		\begin{subfigure}{0.48\textwidth}
			\includegraphics[width=\textwidth]{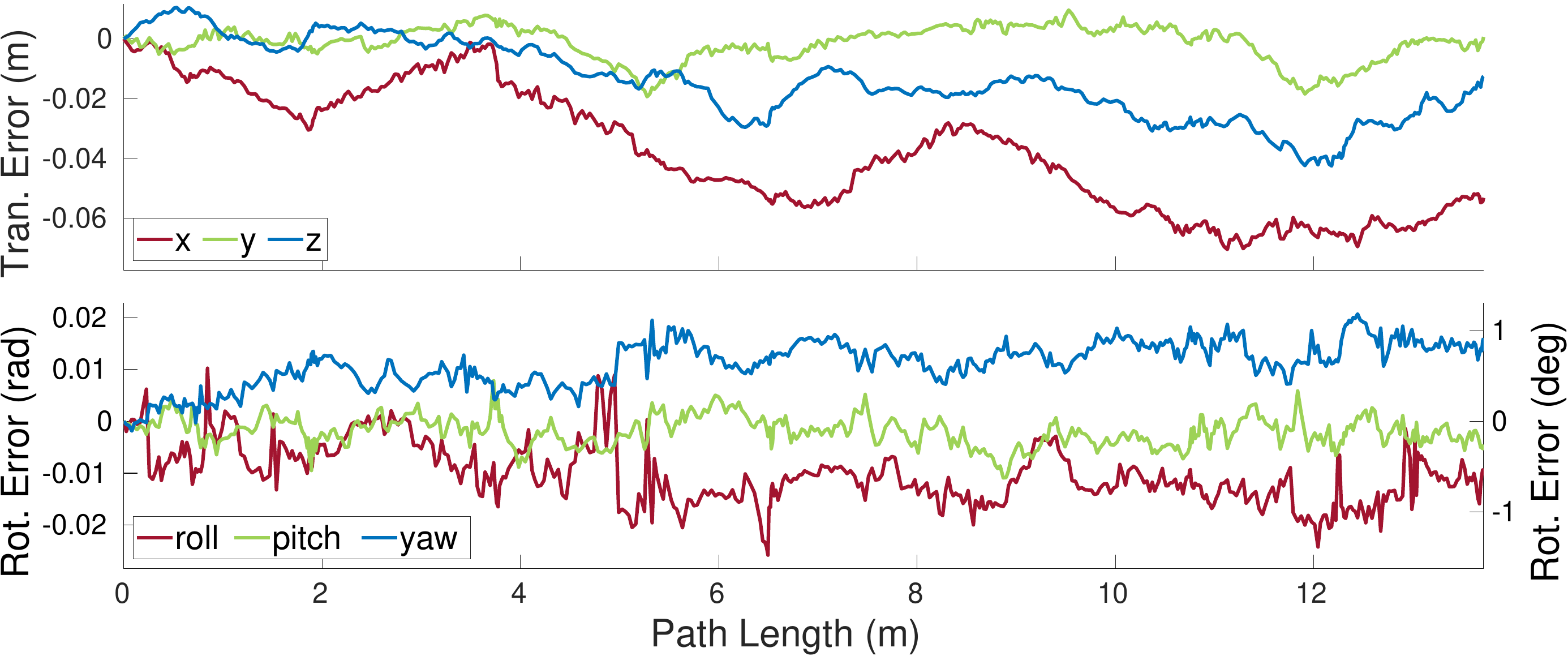}
			\vspace{-6mm}
			\caption{Camera Egomotion}
			\label{fig:results:swinging:wnoa:ego}
			\vspace{6mm}
		\end{subfigure}\\
		\begin{subfigure}{0.48\textwidth}
			\includegraphics[width=\textwidth]{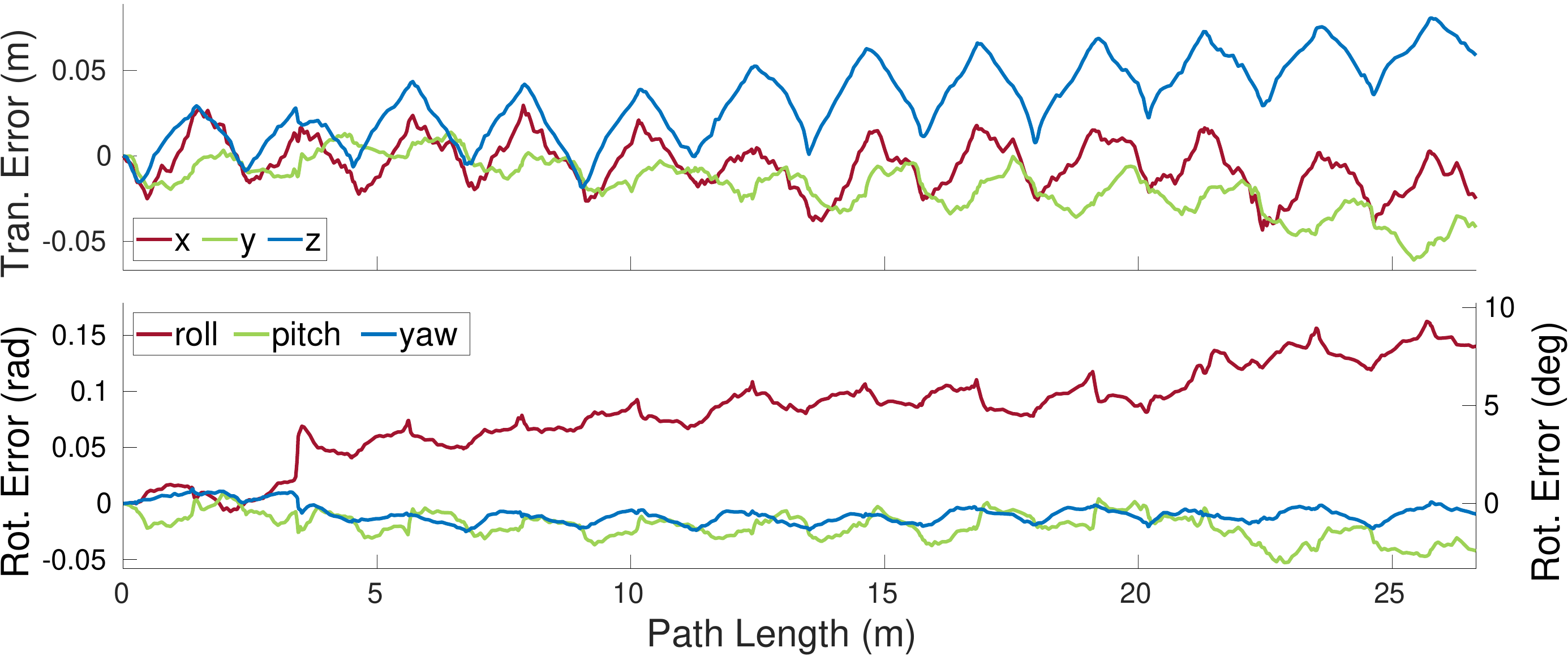}
			\vspace{-6mm}
			\caption{Block 1}
			\label{fig:results:swinging:wnoa:1}
			\vspace{6mm}
		\end{subfigure}\\
		\begin{subfigure}{0.48\textwidth}
			\includegraphics[width=\textwidth]{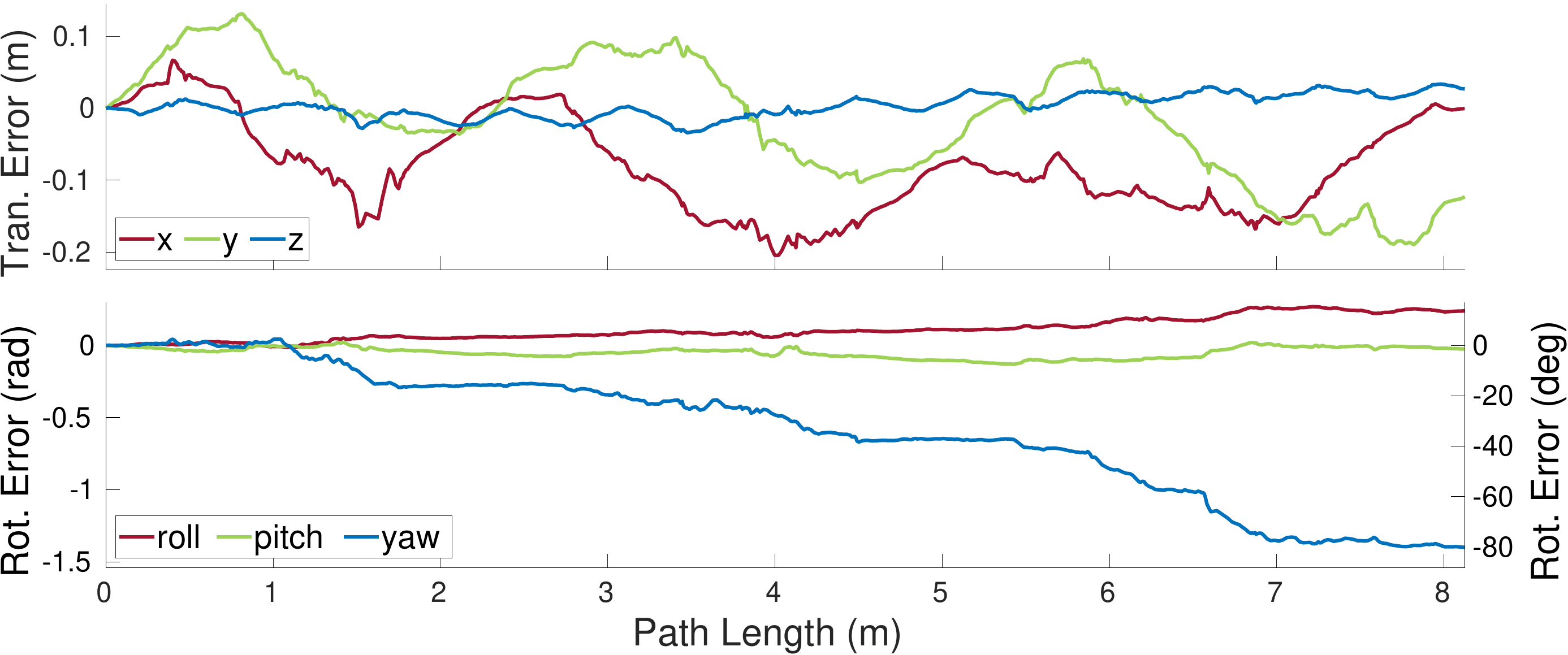}
			\vspace{-6mm}
			\caption{Block 2}
			\label{fig:results:swinging:wnoa:2}
			\vspace{6mm}
		\end{subfigure}\\
		\begin{subfigure}{0.48\textwidth}
			\includegraphics[width=\textwidth]{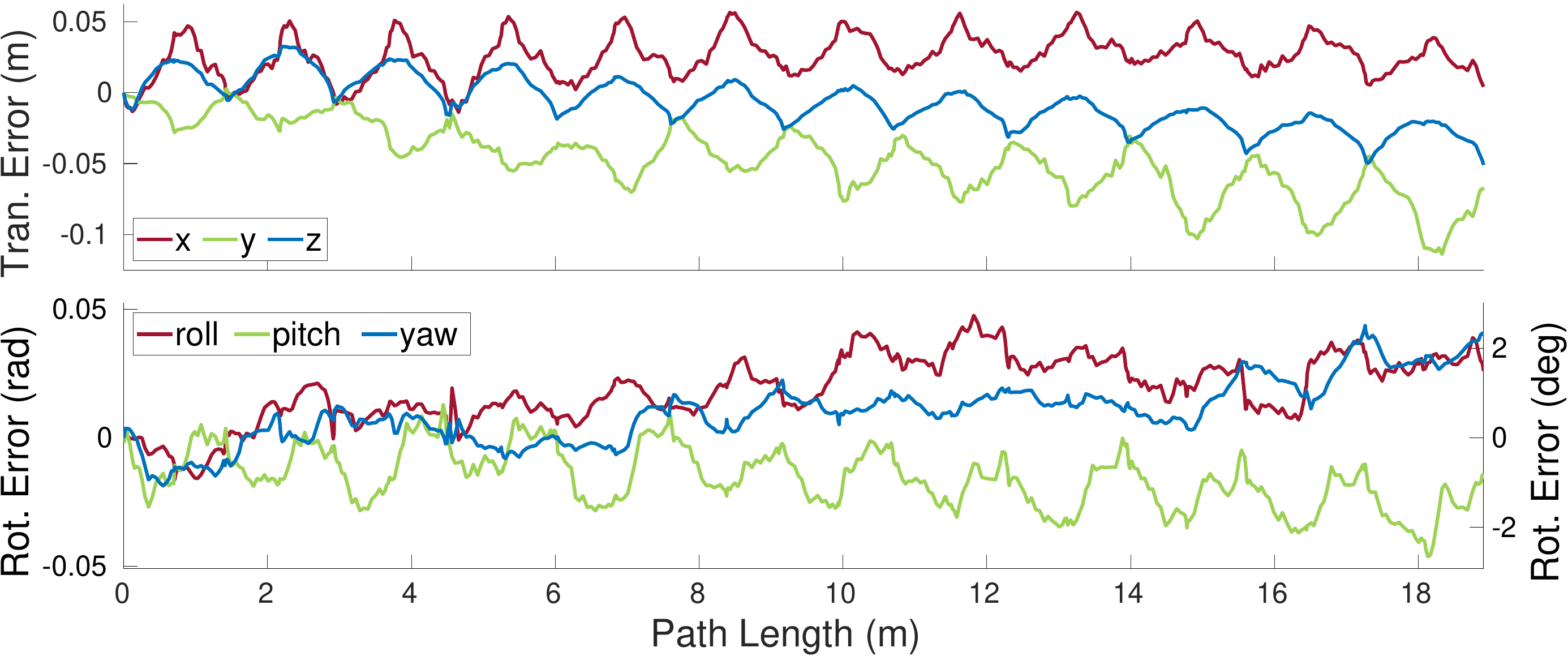}
			\vspace{-6mm}
			\caption{Block 3}
			\label{fig:results:swinging:wnoa:3}
			\vspace{6mm}
		\end{subfigure}\\
		\begin{subfigure}{0.48\textwidth}
			\includegraphics[width=\textwidth]{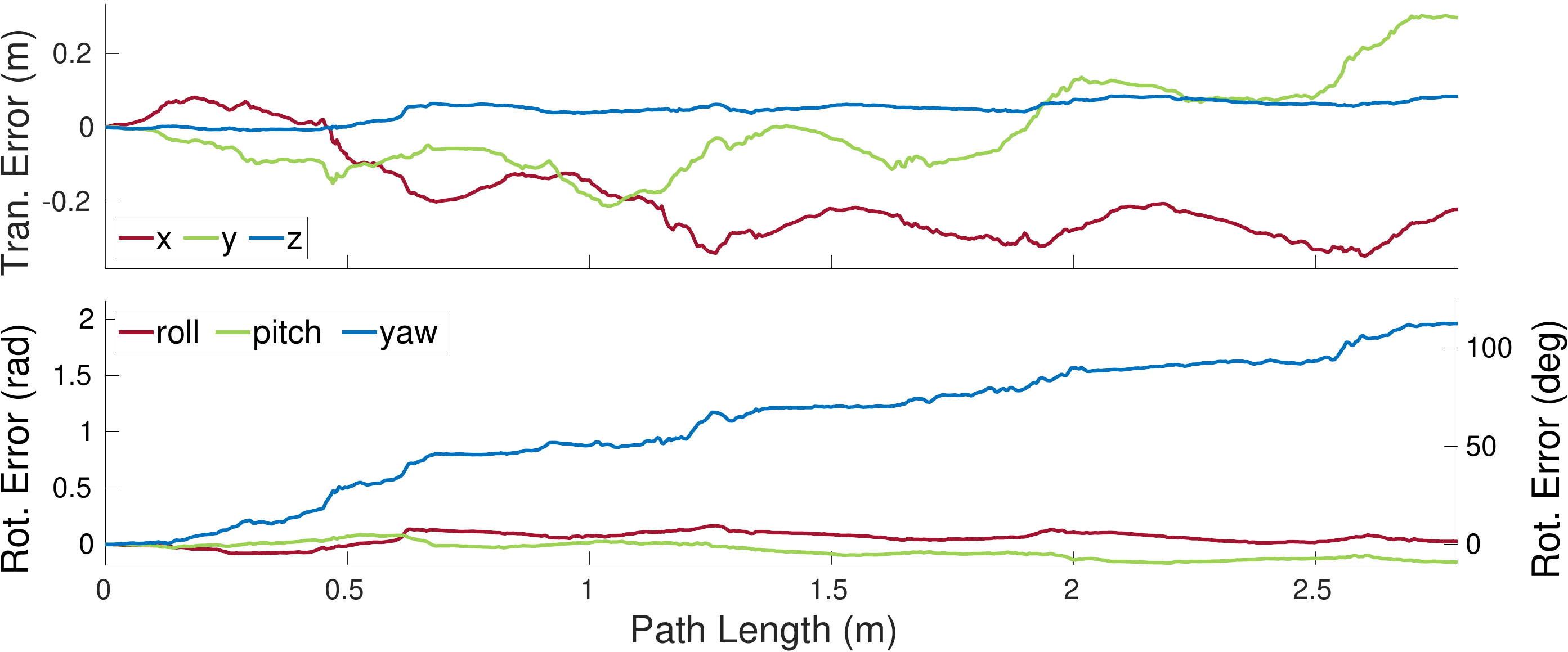}
			\vspace{-6mm}
			\caption{Block 4}
			\label{fig:results:swinging:wnoa:4}
		\end{subfigure}
		\caption{The translational and rotational errors estimated by \acs{MVO} for the camera (a) and the swinging blocks (b--e) for a 500-frame section of the \kjdatasegment{swinging\_4\_unconstrained} segment from the \acs{OMD} using the pose-velocity estimator.
			Each object motion is compared to ground-truth trajectory data and errors are reported in an arbitrary geocentric frame with the z-axis up and arbitrary x- and y-axes.
		}%
		\label{fig:results:swinging:wnoa}
	\end{figure}
	
	\begin{figure}[t]
		\begin{subfigure}{0.48\textwidth}
			\includegraphics[width=\textwidth]{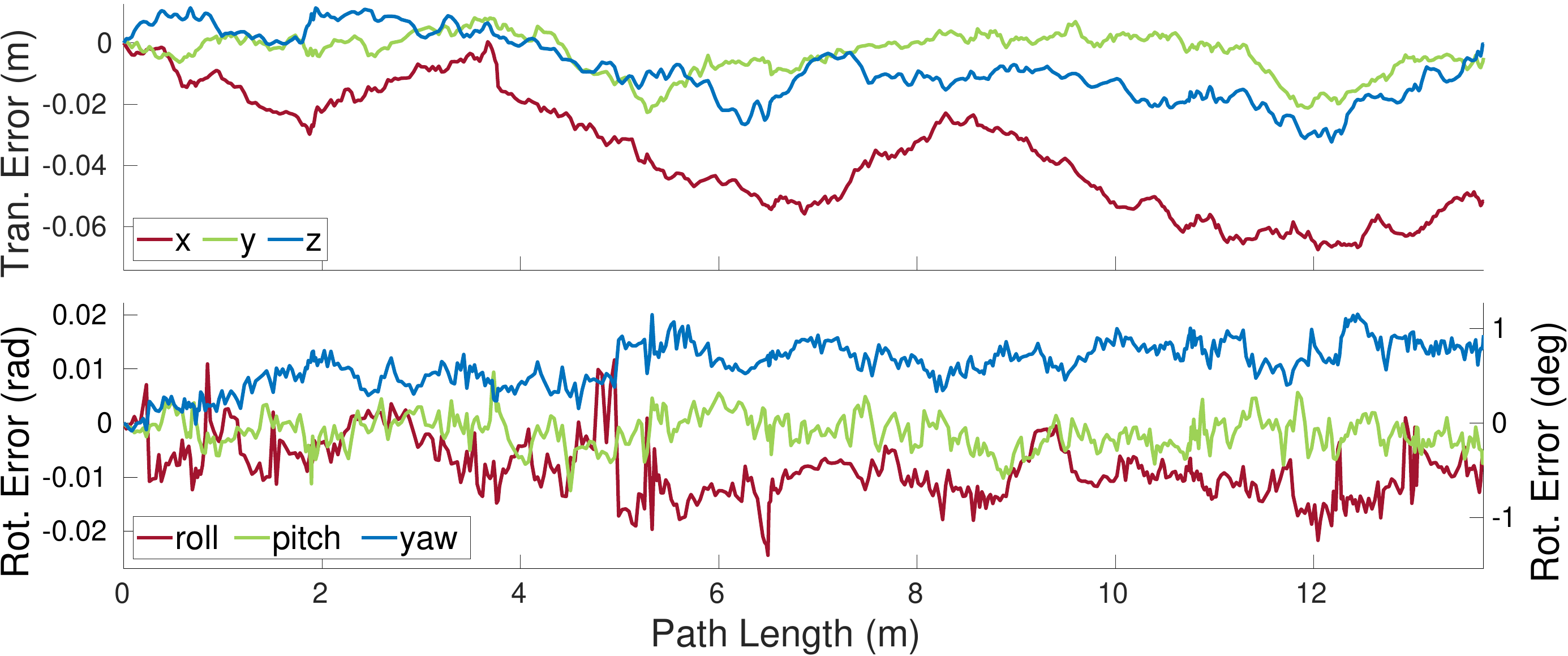}
			\vspace{-6mm}
			\caption{Camera Egomotion}
			\label{fig:results:swinging:wnoj:ego}
			\vspace{6mm}
		\end{subfigure}
		\\
		\begin{subfigure}{0.48\textwidth}
			\includegraphics[width=\textwidth]{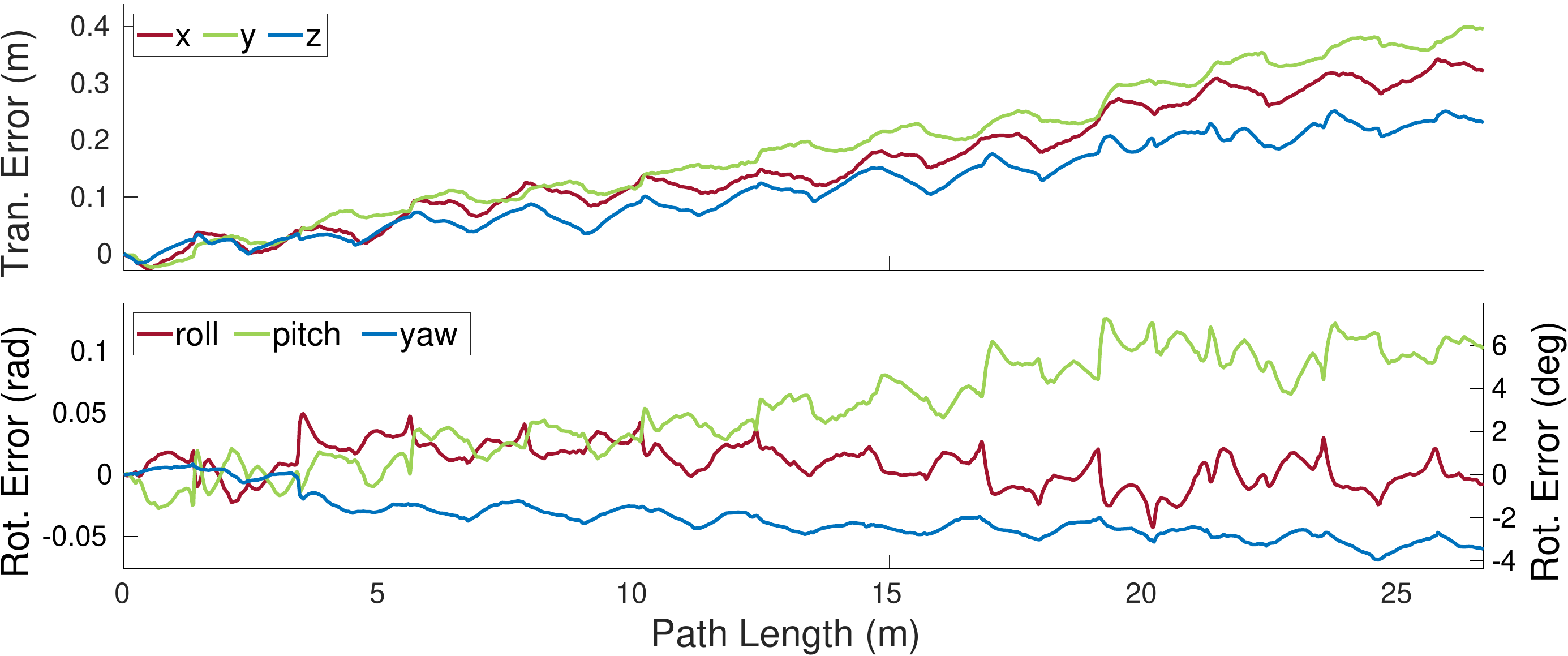}
			\vspace{-6mm}
			\caption{Block 1}
			\label{fig:results:swinging:wnoj:1}
			\vspace{6mm}
		\end{subfigure}
		\\
		\begin{subfigure}{0.48\textwidth}
			\includegraphics[width=\textwidth]{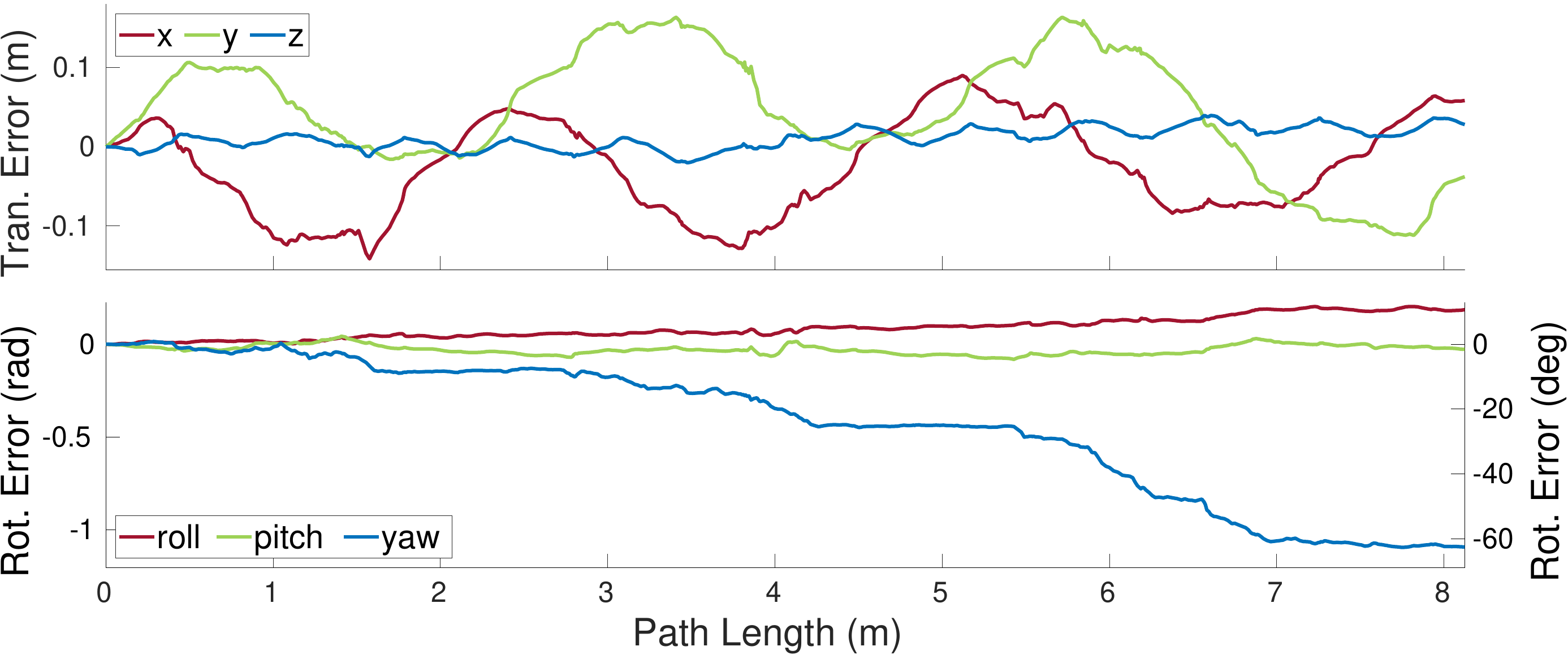}
			\vspace{-6mm}
			\caption{Block 2}
			\label{fig:results:swinging:wnoj:2}
			\vspace{6mm}
		\end{subfigure}
		\\
		\begin{subfigure}{0.48\textwidth}
			\includegraphics[width=\textwidth]{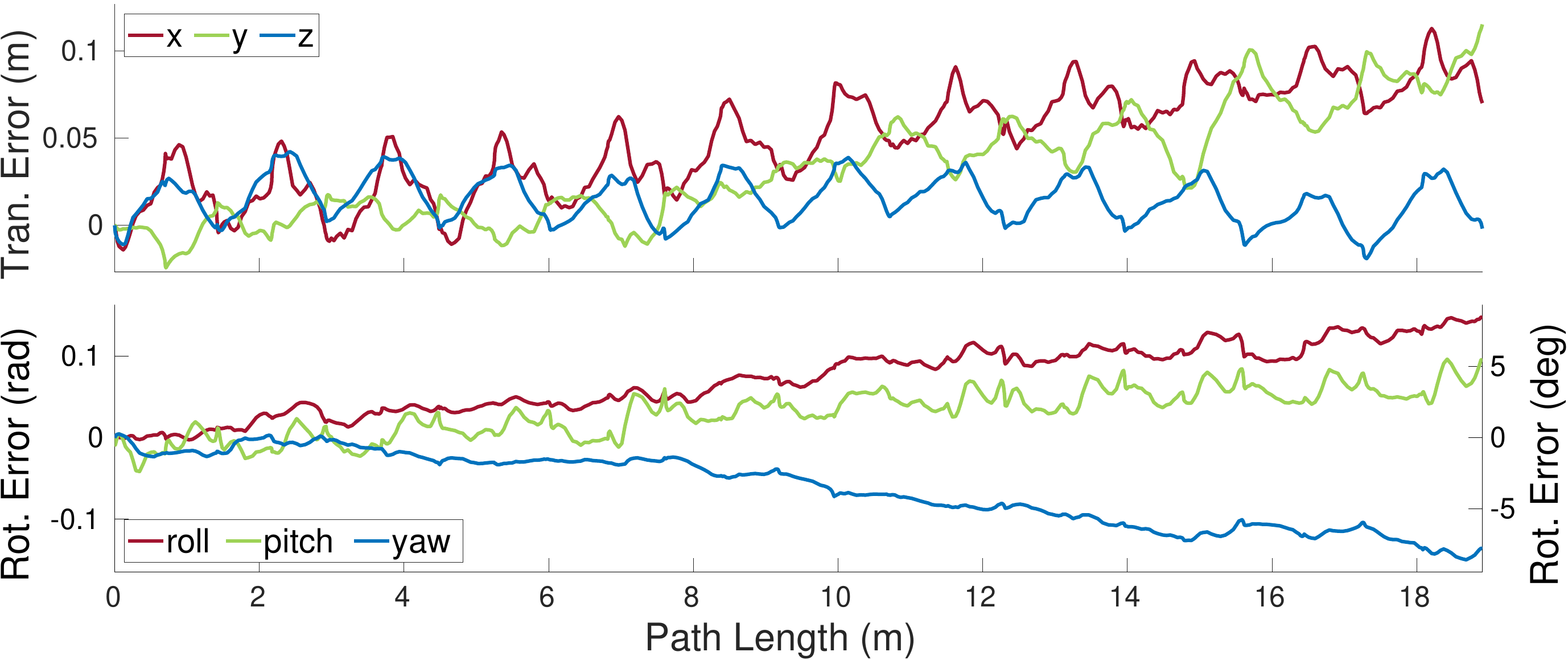}
			\vspace{-6mm}
			\caption{Block 3}
			\label{fig:results:swinging:wnoj:3}
			\vspace{6mm}
		\end{subfigure}
		\\
		\begin{subfigure}{0.48\textwidth}
			\includegraphics[width=\textwidth]{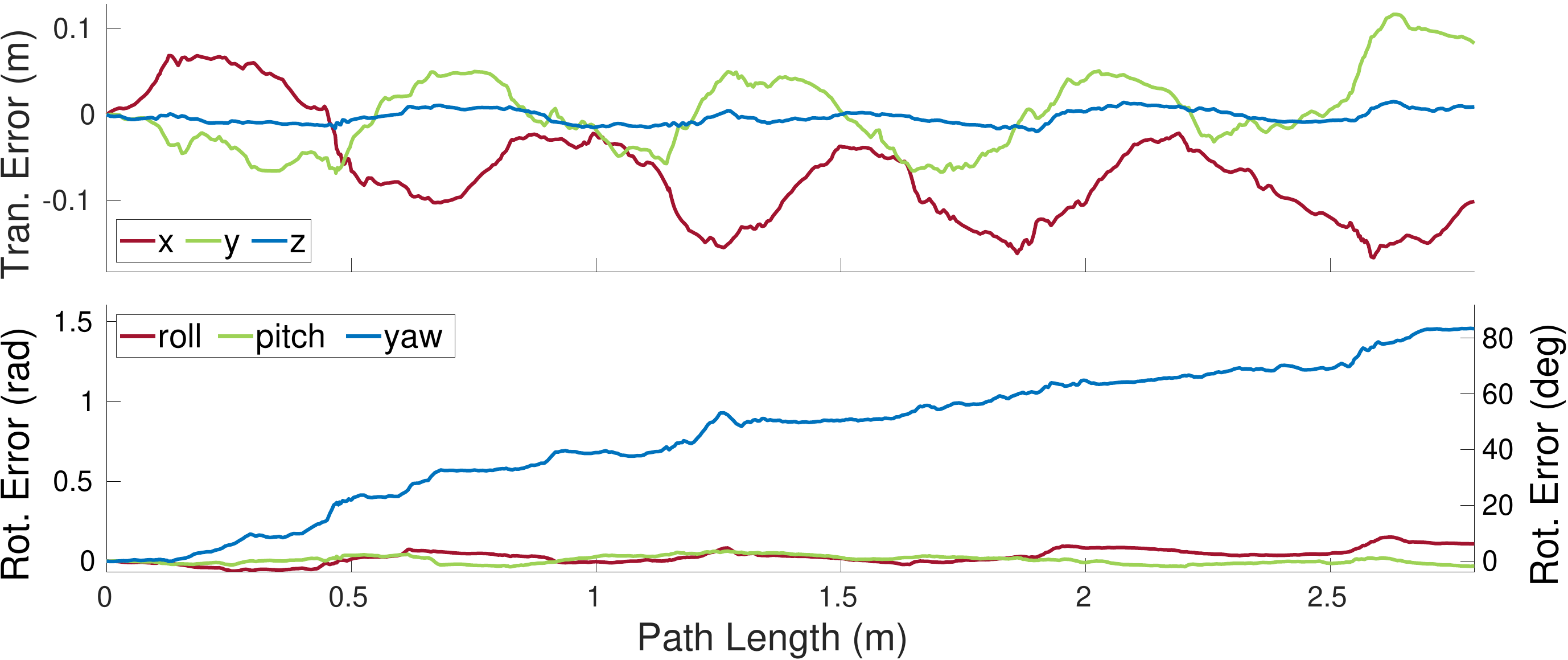}
			\vspace{-6mm}
			\caption{Block 4}
			\label{fig:results:swinging:wnoj:4}
		\end{subfigure}
		\caption{The translational and rotational errors estimated by \acs{MVO} for the camera (a) and the swinging blocks (b--e) for a 500-frame section of the \kjdatasegment{swinging\_4\_unconstrained} segment from the \acs{OMD} using the pose-velocity-acceleration estimator.
			Each object motion is compared to ground-truth trajectory data and errors are reported in an arbitrary geocentric frame with the z-axis up and arbitrary x- and y-axes.
		}%
		\label{fig:results:swinging:wnoj}
	\end{figure}
	
	\begin{figure}[t]
		\begin{subfigure}{0.48\textwidth}
			\includegraphics[width=\textwidth]{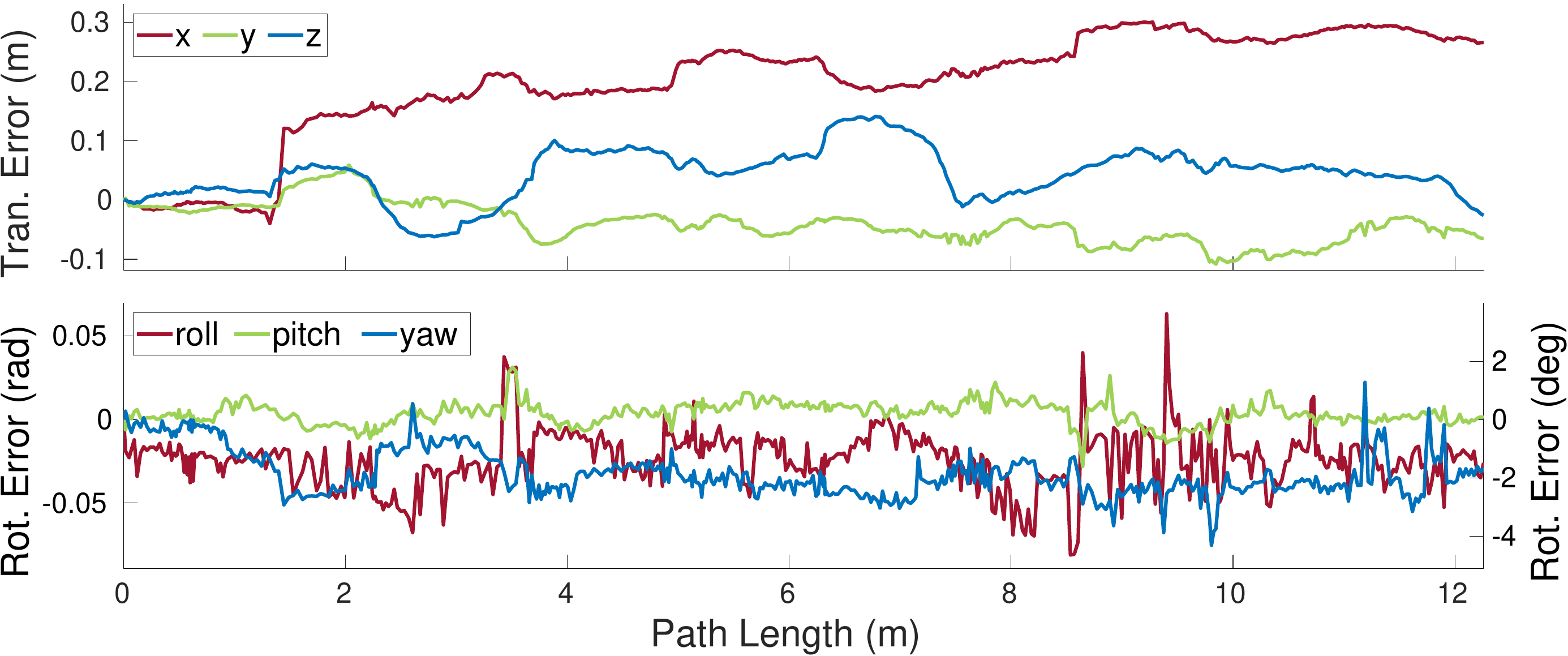}
			\vspace{-6mm}
			\caption{Camera Egomotion}
			\label{fig:results:occlusion:discrete:ego}
			\vspace{6mm}
		\end{subfigure}\\
		\begin{subfigure}{0.48\textwidth}
			\includegraphics[width=\textwidth]{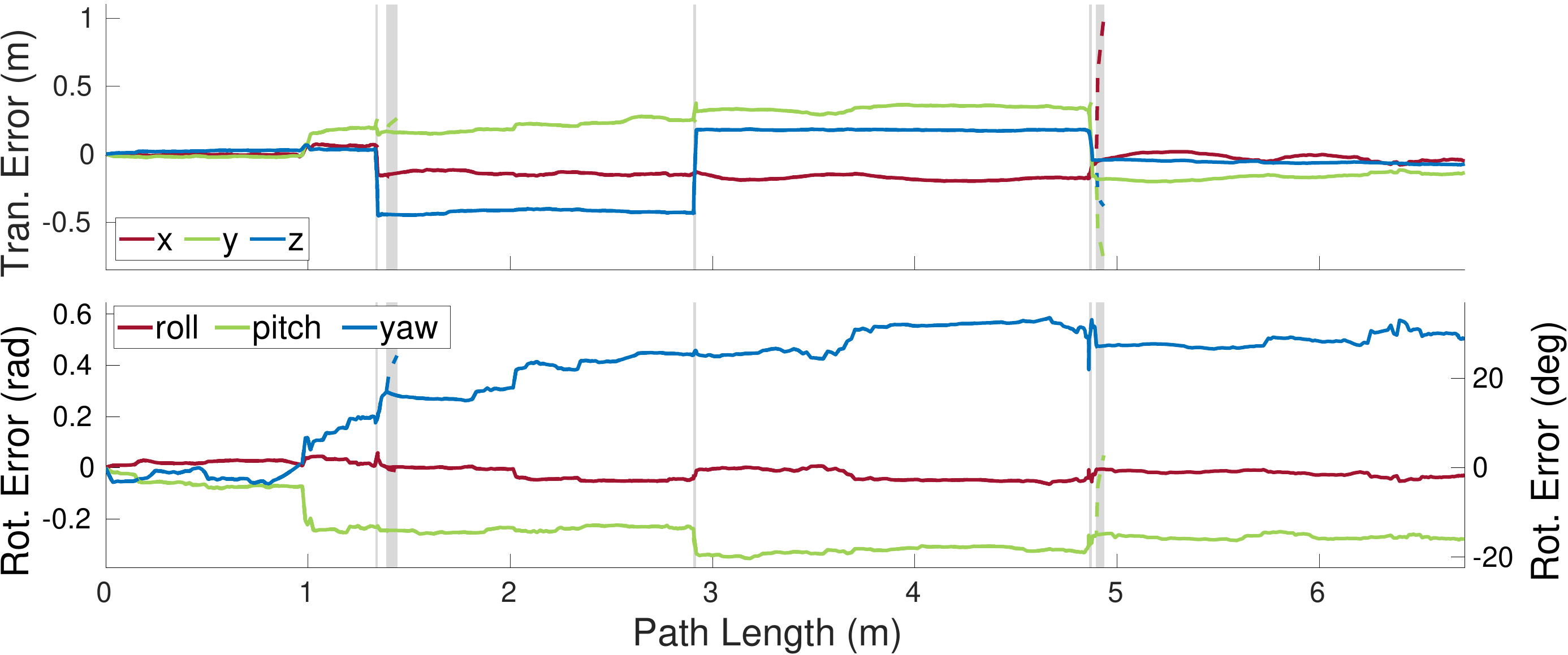}
			\vspace{-6mm}
			\caption{Block Tower 1}
			\label{fig:results:occlusion:discrete:1}
			\vspace{6mm}
		\end{subfigure}\\
		\begin{subfigure}{0.48\textwidth}
			\includegraphics[width=\textwidth]{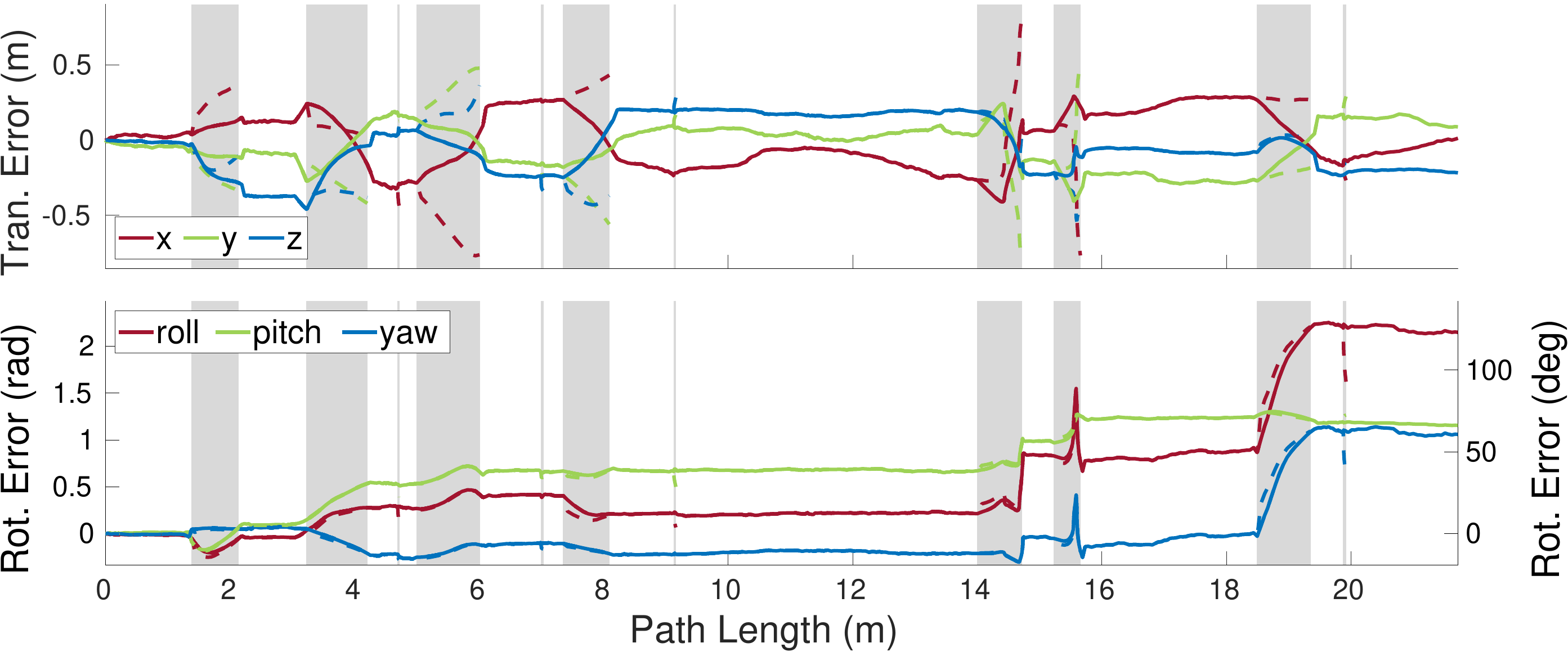}
			\vspace{-6mm}
			\caption{Swinging Block 4}
			\label{fig:results:occlusion:discrete:4}
		\end{subfigure}
		\caption{The translational and rotational errors estimated by \acs{MVO} for the camera (a), the block tower (b), and the swinging block (c) for a 500-frame section of the \kjdatasegment{occlusion\_2\_unconstrained} segment from the \acs{OMD} using the pose-only estimator.
			Each object motion is compared to ground-truth trajectory data and errors are reported in an arbitrary geocentric frame with the z-axis up and arbitrary x- and y-axes.
			Grey regions show when the swinging block was occluded by the tower, or when the tower was stationary and effectively part of the background.
			Dashed lines represent the extrapolation error and the solid lines represent the error in the direct or interpolated estimates.
		}%
		\label{fig:results:occlusion:discrete}
	\end{figure}
	
	\begin{figure}[t]
		\begin{subfigure}{0.48\textwidth}
			\includegraphics[width=\textwidth]{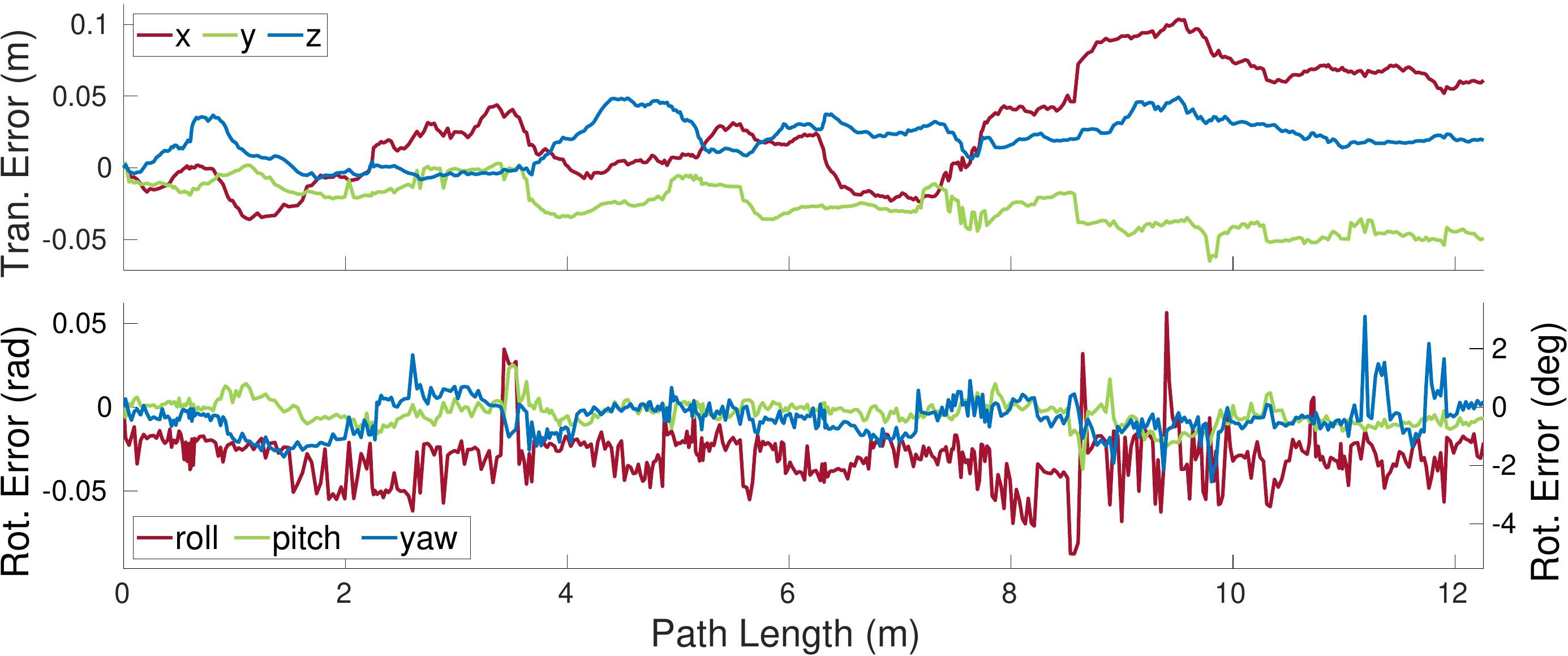}
			\vspace{-6mm}
			\caption{Camera Egomotion}
			\label{fig:results:occlusion:wnoa:ego}
			\vspace{6mm}
		\end{subfigure}\\
		\begin{subfigure}{0.48\textwidth}
			\includegraphics[width=\textwidth]{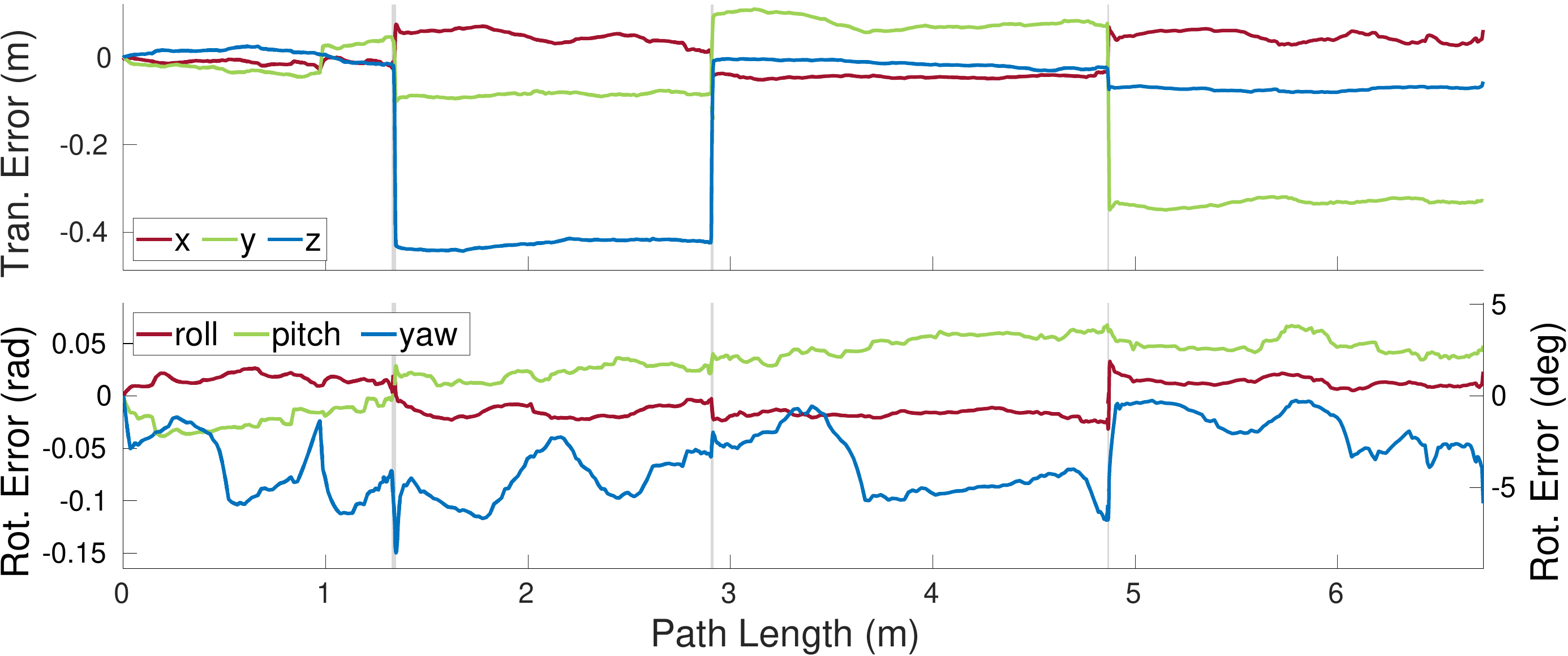}
			\vspace{-6mm}
			\caption{Block Tower 1}
			\label{fig:results:occlusion:wnoa:1}
			\vspace{6mm}
		\end{subfigure}\\
		\begin{subfigure}{0.48\textwidth}
			\includegraphics[width=\textwidth]{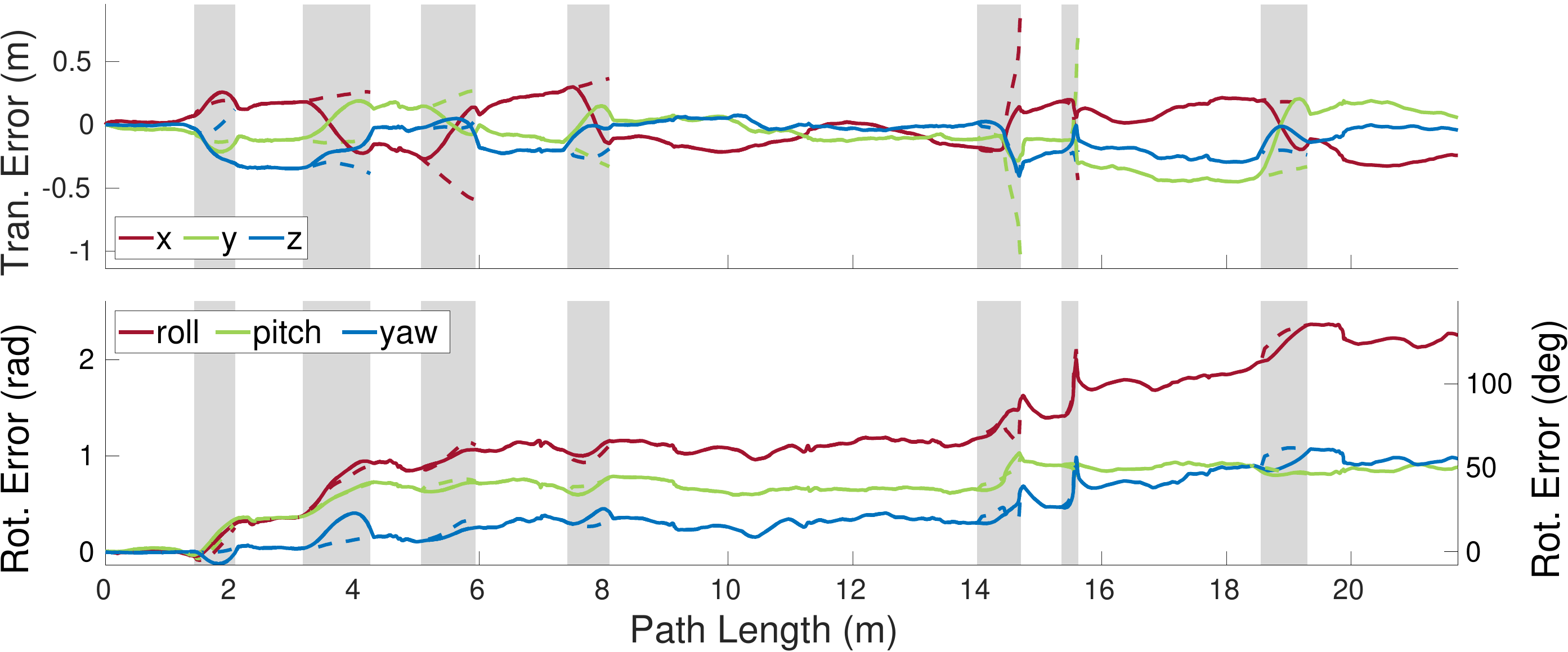}
			\vspace{-6mm}
			\caption{Swinging Block 4}
			\label{fig:results:occlusion:wnoa:4}
		\end{subfigure}
		\caption{The translational and rotational errors estimated by \acs{MVO} for the camera (a), the block tower (b), and the swinging block (c) for a 500-frame section of the \kjdatasegment{occlusion\_2\_unconstrained} segment from the \acs{OMD} using the pose-velocity estimator.
			Each object motion is compared to ground-truth trajectory data and errors are reported in an arbitrary geocentric frame with the z-axis up and arbitrary x- and y-axes.
			Grey regions show when the swinging block was occluded by the tower, or when the tower was stationary and effectively part of the background.
			Dashed lines represent the extrapolation error and the solid lines represent the error in the direct or interpolated estimates.
		}%
		\label{fig:results:occlusion:wnoa}
	\end{figure}
	
	\begin{figure}[t]
		\begin{subfigure}{0.48\textwidth}
			\includegraphics[width=\textwidth]{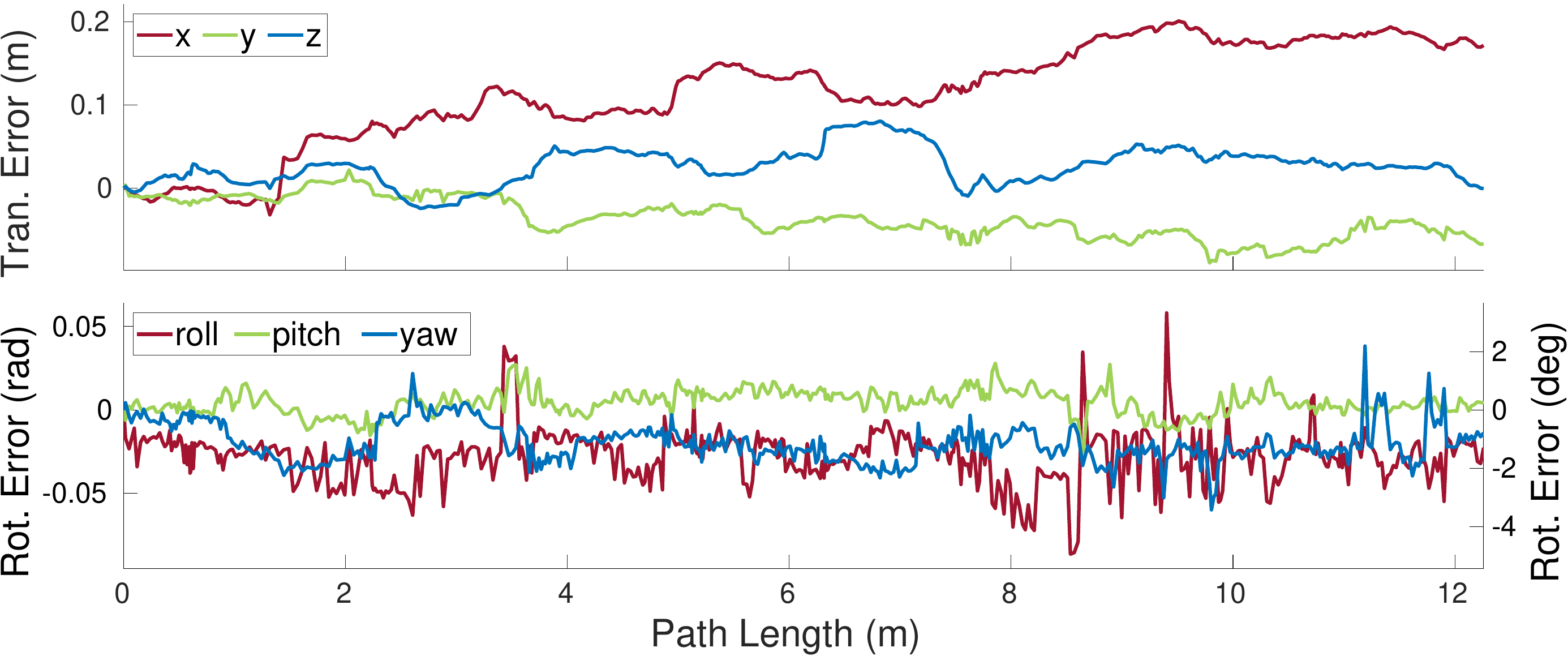}
			\vspace{-6mm}
			\caption{Camera Egomotion}
			\label{fig:results:occlusion:wnoj:ego}
			\vspace{6mm}
		\end{subfigure}\\
		\begin{subfigure}{0.48\textwidth}
			\includegraphics[width=\textwidth]{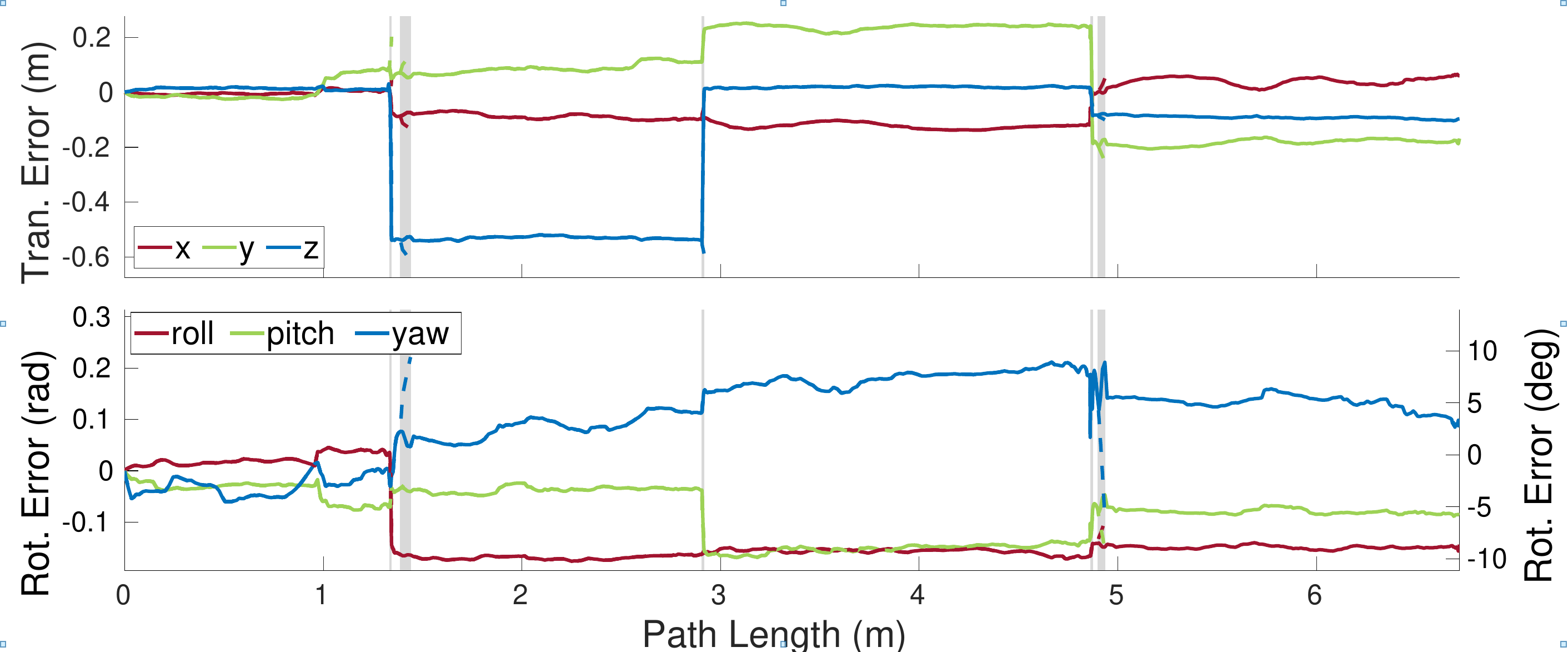}
			\vspace{-6mm}
			\caption{Block Tower 1}
			\label{fig:results:occlusion:wnoj:1}
			\vspace{6mm}
		\end{subfigure}\\
		\begin{subfigure}{0.48\textwidth}		\includegraphics[width=\textwidth]{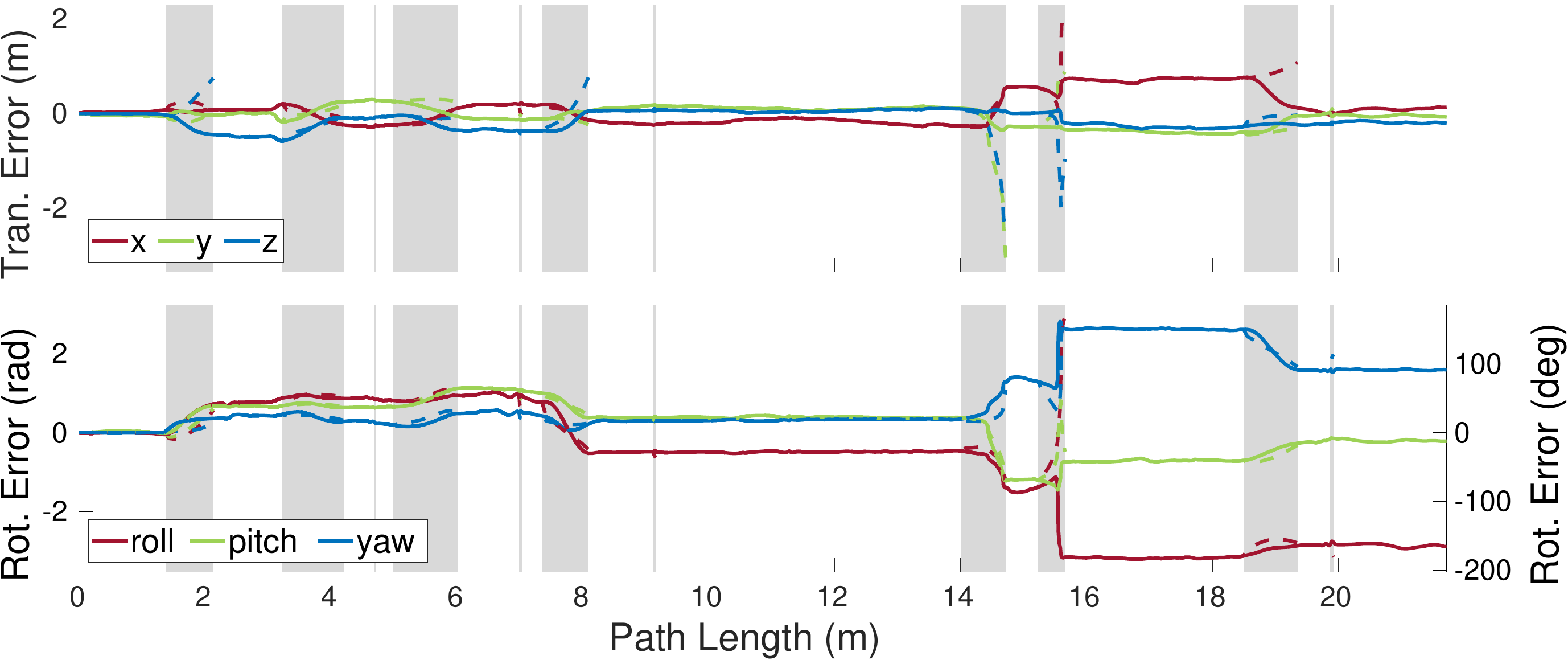}
			\vspace{-6mm}
			\caption{Swinging Block 4}
			\label{fig:results:occlusion:wnoj:4}
		\end{subfigure}
		\caption{The translational and rotational errors estimated by \acs{MVO} for the camera (a), the block tower (b), and the swinging block (c) for a 500-frame section of the \kjdatasegment{occlusion\_2\_unconstrained} segment from the \acs{OMD} using the pose-velocity-acceleration estimator.
			Each object motion is compared to ground-truth trajectory data and errors are reported in an arbitrary geocentric frame with the z-axis up and arbitrary x- and y-axes.
			Grey regions show when the swinging block was occluded by the tower, or when the tower was stationary and effectively part of the background.
			Dashed lines represent the extrapolation error and the solid lines represent the error in the direct or interpolated estimates.
		}%
		\label{fig:results:occlusion:wnoj}
	\end{figure}

	\begin{figure}[t]
		\begin{subfigure}{0.48\textwidth}
			\includegraphics[width=\textwidth]{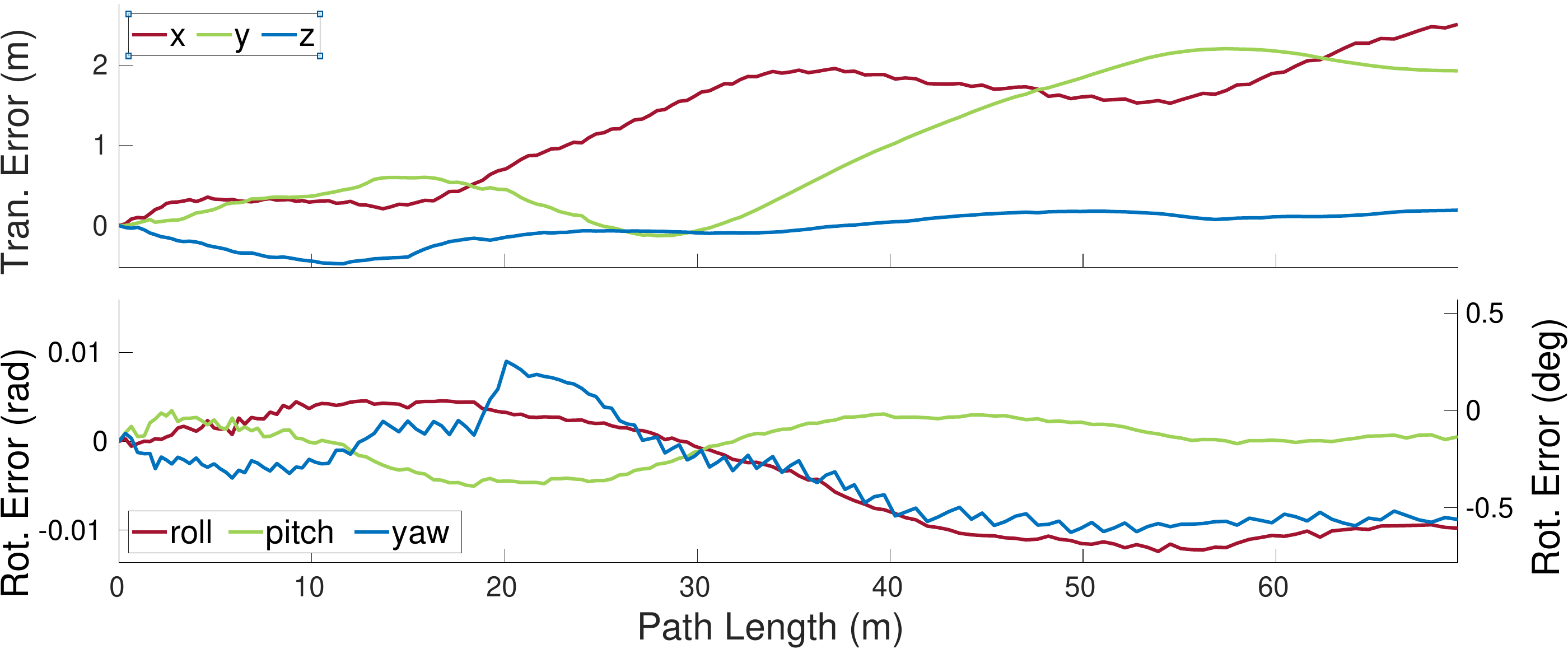}
			\caption{Pose-Only Estimator}
			\label{fig:results:kitti:discrete}
		\end{subfigure}\\
		\begin{subfigure}{0.48\textwidth}
			\includegraphics[width=\textwidth]{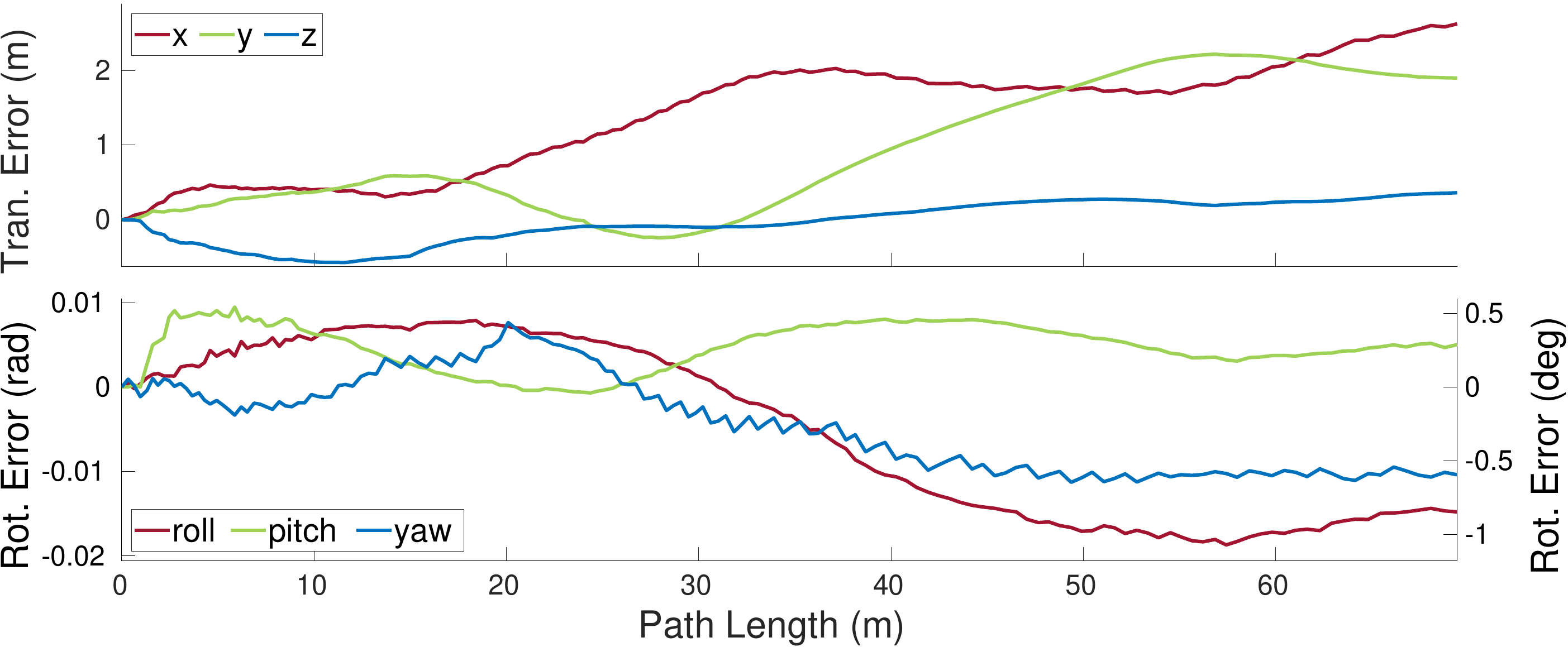}
			\caption{Pose-Velocity Estimator}
			\label{fig:results:kitti:wnoa}
		\end{subfigure}\\
		\begin{subfigure}{0.48\textwidth}
			\includegraphics[width=\textwidth]{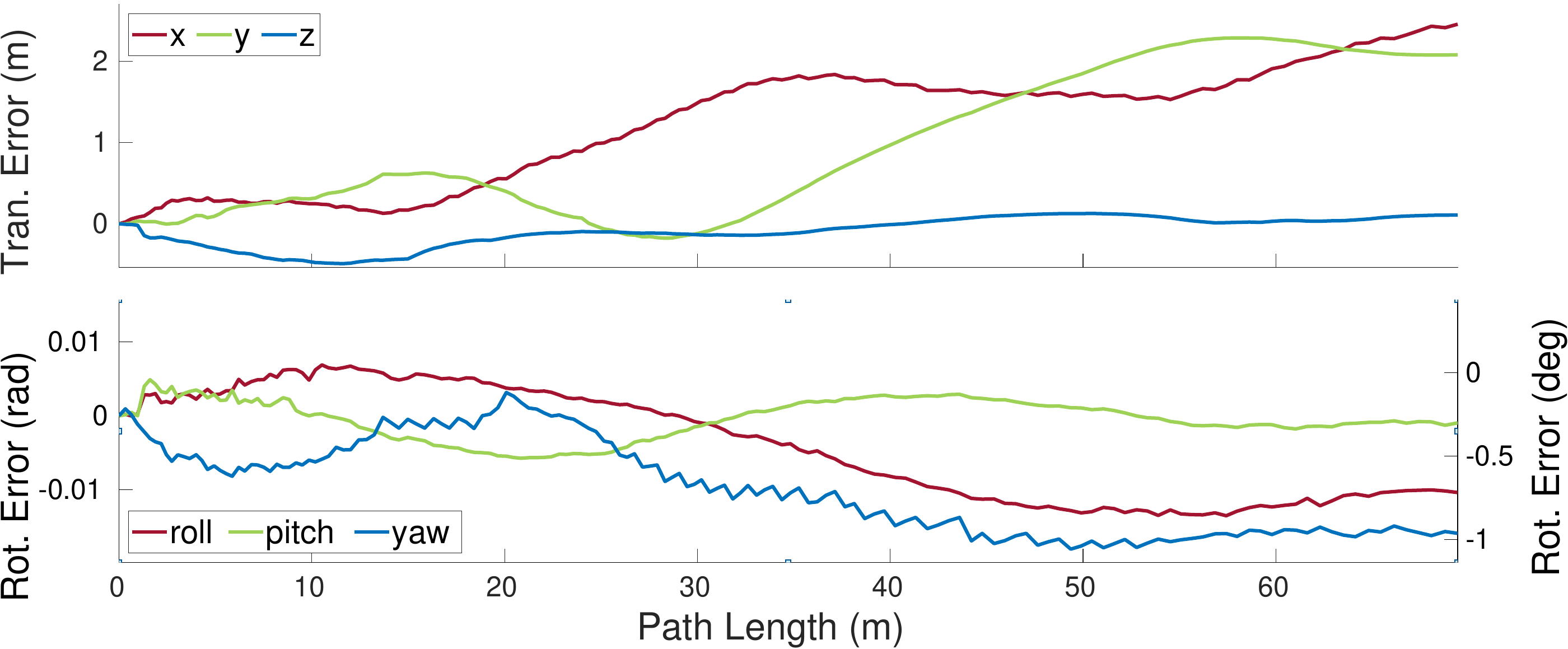}
			\caption{Pose-Velocity-Acceleration Estimator}
			\label{fig:results:kitti:wnoj}
		\end{subfigure}
		\caption{The translational and rotational errors estimated by \acs{MVO} for the camera for the full 154-frame \kjdatasegment{Drive 0005} data segment from the KITTI Vision Benchmark Suite using each \acs{MVO} estimator.
			There is no ground-truth trajectory data for the van or the cyclist.
		}%
		\label{fig:results:kitti}
	\end{figure}
	
	\begin{figure}[t]
		\begin{subfigure}{0.48\textwidth}
			\includegraphics[width=\textwidth]{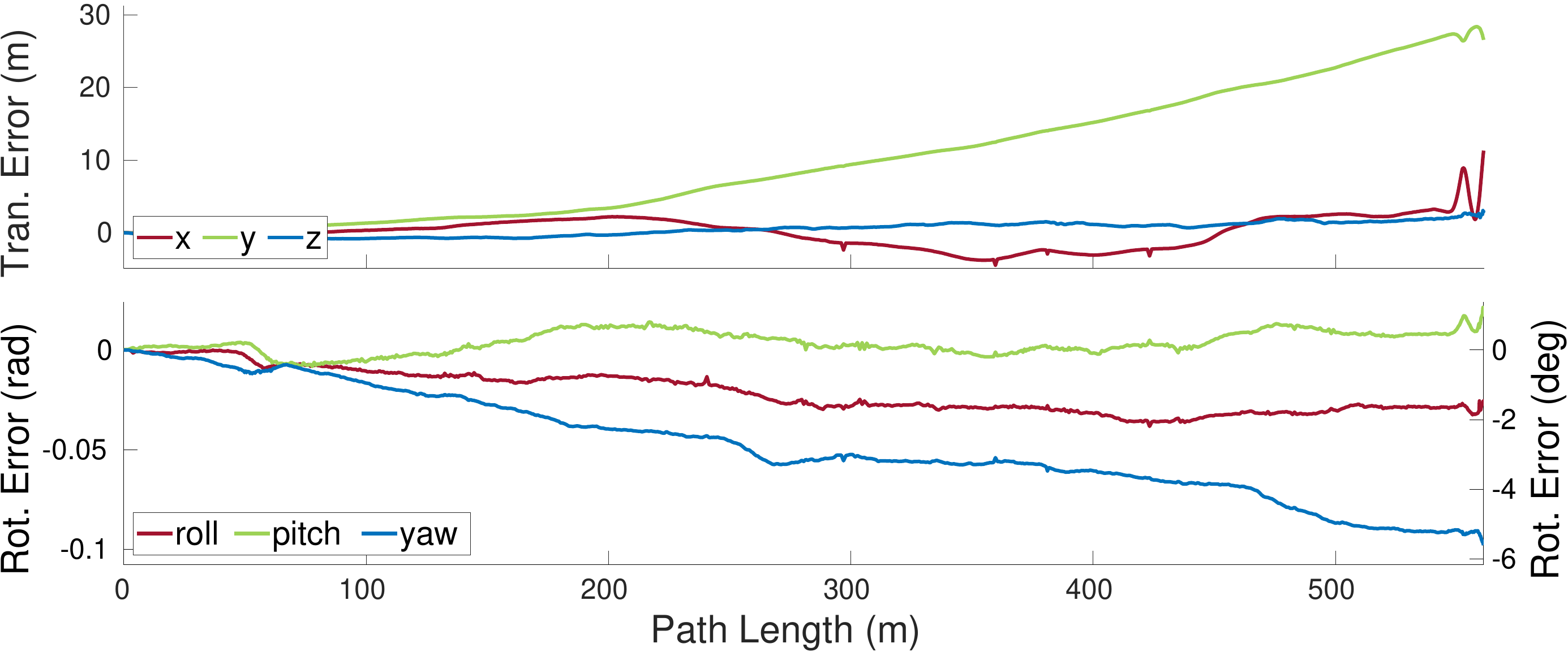}
			\caption{Pose-Only Estimator}
			\label{fig:results:kitti_od:discrete}
		\end{subfigure}\\
		\begin{subfigure}{0.48\textwidth}
			\includegraphics[width=\textwidth]{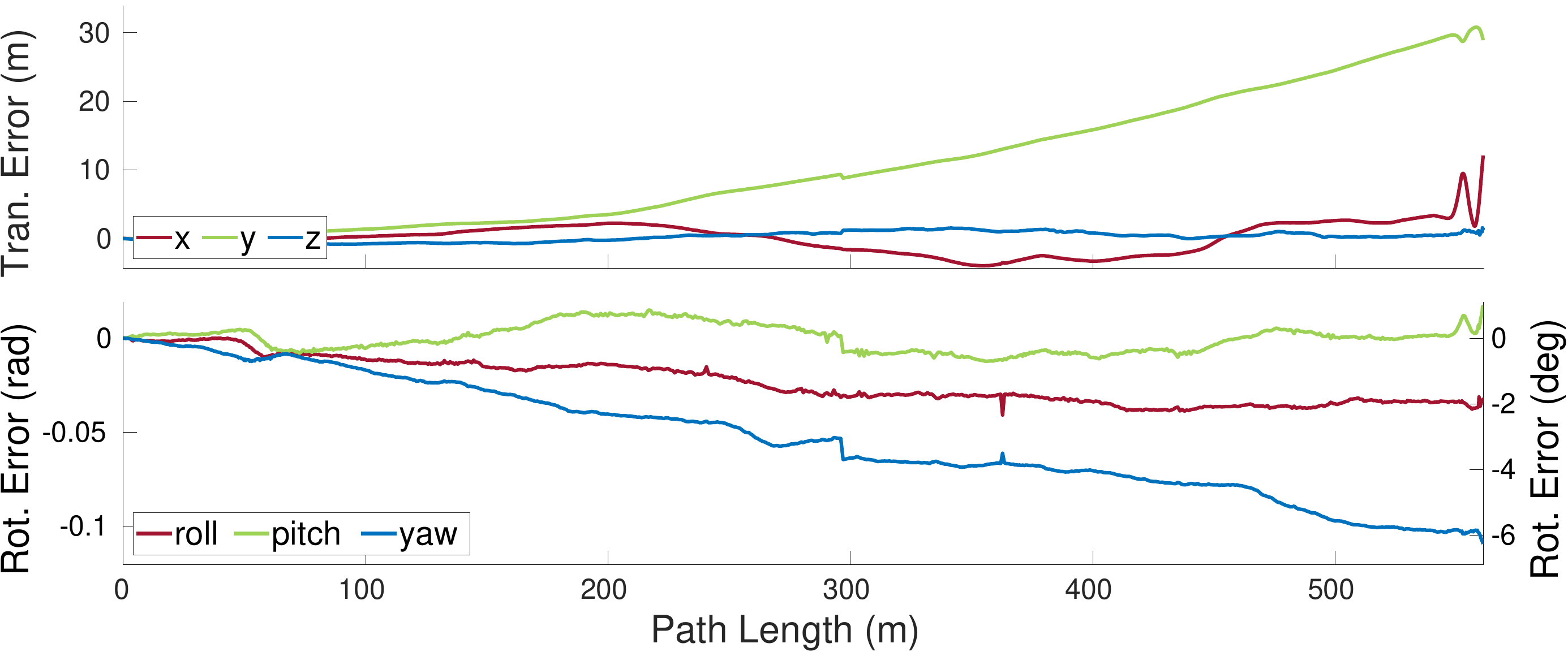}
			\caption{Pose-Velocity Estimator}
			\label{fig:results:kitti_od:wnoa}
		\end{subfigure}\\
		\begin{subfigure}{0.48\textwidth}
			\includegraphics[width=\textwidth]{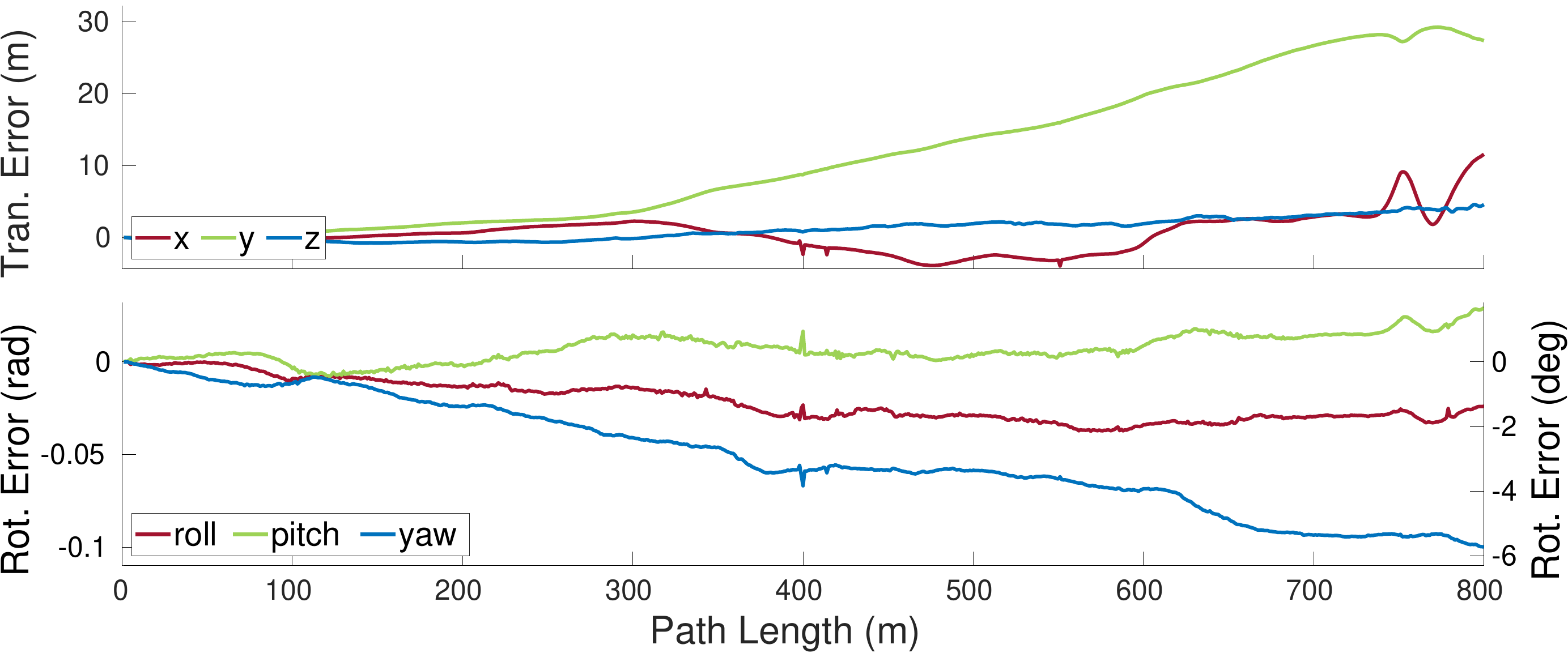}
			\caption{Pose-Velocity-Acceleration Estimator}
			\label{fig:results:kitti_od:wnoj}
		\end{subfigure}
		\caption{The translational and rotational errors estimated by \acs{MVO} for the camera for the full 800-frame \kjdatasegment{Odometry 03} data segment from the KITTI Vision Benchmark Suite using each \acs{MVO} estimator.
			There is no ground-truth trajectory data for the other car.
		}%
		\label{fig:results:kitti_od}
	\end{figure}
	
	%%%%% REFERENCES
	\bibliographystyle{SageH}
	\bibliography{references.bib}
	
\end{document}